\documentclass[10pt,twocolumn,letterpaper]{article}
\usepackage{cvpr}  
\usepackage{graphicx}
\usepackage{amsmath}
\usepackage{amssymb}
\usepackage{booktabs}
\usepackage{times}
\usepackage{epsfig}
\usepackage{graphicx}
\usepackage{amsmath}
\usepackage{amssymb}
\usepackage{graphicx}
\usepackage{comment}
\usepackage{amsmath,amssymb}  
\usepackage{varwidth}
\usepackage{multirow}
\usepackage{lipsum,graphicx}
\usepackage{booktabs}
\usepackage{varwidth}
\usepackage{fancyhdr,graphicx,amsmath,amssymb}
\usepackage[ruled,vlined]{algorithm2e} 
\usepackage{graphicx}
\usepackage{lipsum}     
\usepackage{graphicx}   
\usepackage{dblfloatfix}    
\usepackage{stackengine}
\usepackage{amsmath}
\usepackage{bm}
\usepackage{color}
\usepackage{empheq}
\usepackage{microtype} 
\usepackage{xcolor, soul}
\sethlcolor{lightgray}
\usepackage{rotating}
\usepackage{enumitem}
	
\usepackage{color, colortbl}

\usepackage[first=0,last=9]{lcg}

\usepackage[breaklinks=true,colorlinks,bookmarks=false,pagebackref]{hyperref}    

\usepackage{cleveref}
\crefname{section}{sec.}{secs.}
\Crefname{section}{Sec.}{Secs.}
\crefname{table}{tab.}{tabs.}
\Crefname{table}{Tab.}{Tabs.}
\crefname{figure}{fig.}{figs.}
\Crefname{figure}{Fig.}{Figs.}
\crefname{equation}{eq.}{eqs.}
\Crefname{equation}{Eq.}{Eqs.}

\def\cvprPaperID{2919} 
\def\confName{CVPR}
\def\confYear{2022}

\begin{document}

\title{Matching Feature Sets for Few-Shot Image Classification} 
\author{Arman Afrasiyabi$^{\star \bullet}$, 
Hugo Larochelle$^{\diamond \dag \bullet}$, 
Jean-Fran\c{c}ois Lalonde$^\star$, 
Christian Gagn\'e$^{\star \dag \bullet}$ 
\\
$^\star$Universit\'e Laval, $^\diamond$Google Brain, $^\dag$Canada CIFAR AI Chair, $^\bullet$Mila   \\ 
\small{\texttt{\url{https://lvsn.github.io/SetFeat/}}} 
} 
\newcommand\myparagraph[1]{\vspace{0.25em}\noindent\textbf{#1}\quad}
\maketitle

\begin{abstract}
In image classification, it is common practice to train deep networks to extract a single feature vector per input image. Few-shot classification methods also mostly follow this trend. In this work, we depart from this established direction and instead propose to extract \emph{sets} of feature vectors for each image. We argue that a set-based representation intrinsically builds a richer representation of images from the base classes, which can subsequently better transfer to the few-shot classes. To do so, we propose to adapt existing feature extractors to instead produce \emph{sets} of feature vectors from images. Our approach, dubbed SetFeat, embeds shallow self-attention mechanisms inside existing encoder architectures. The attention modules are lightweight, and as such our method results in encoders that have approximately the same number of parameters as their original versions. During training and inference, a set-to-set matching metric is used to perform image classification. The effectiveness of our proposed architecture and metrics is demonstrated via thorough experiments on standard few-shot datasets---namely miniImageNet, tieredImageNet, and CUB---in both the 1- and 5-shot scenarios. In all cases but one, our method outperforms the state-of-the-art.
\end{abstract}


\section{Introduction}
\label{sec:intro}

The task of few-shot image classification is to transfer knowledge gained on a set of ``base'' categories, assumed to be available in large quantities, to another set of ``novel'' classes of which we are given only very few examples. To solve this problem, a popular strategy is to employ a deep feature extractor which learns to convert an input image into a feature vector that is both discriminative and transferable to the novel classes. In this context, the common practice 
is to train a model to extract, for a given input, a \emph{single} feature vector from which classification decisions are made. 

In this paper, we depart from this established strategy by proposing instead to represent images as \emph{sets} of feature vectors. With this, we aim at learning a richer feature space that is both more discriminative and easier to transfer to the novel domain, by allowing the network to focus on different characteristics of the image and at different scales. The intuition motivating that approach is that decomposing the representation into independent components should allow the capture of several distinctive aspects of images that can then be used efficiently to represent images of novel classes. 

To do so, we take inspiration from Feature Pyramid Networks~\cite{lin2017feature} which proposes to concatenate multi-scale feature maps from convolutional backbones. In contrast, however, we do not just poll the features themselves but rather embed shallow self-attention modules (called ``mappers'') at various scales in the network. This adapted network therefore learns to represent an image via a set of attention-based latent representations. The network is first pre-trained by injecting the signal of a classification loss at every mapper. Then, it is fine-tuned in a meta-training stage, which performs classification by computing the distance between a query (test) and a set of support (training) samples in a manner similar to Prototypical Networks~\cite{snell2017prototypical}. Here, the main difference is that the distance between samples is computed using set-to-set metrics rather than traditional distance functions. To this end, we propose and experiment with three set-to-set metrics. 

This paper presents the following contributions. First, it presents the idea of reasoning on \emph{sets} and a set-based inference of feature vectors extracted from images. It shows that set representation yields improved performance on few-shot image classification without increasing the total number of network parameters. Second, it presents a straightforward way to adapt \emph{existing} backbones to make them extract \emph{sets} of features rather than single ones, and processing them in order to achieve decisions. To do so, it proposed to embed simple self-attention modules in between convolutional blocks, with examples of adapted three popular backbones, namely Conv4-64, Conv4-256, and ResNet12. 
It also proposes set-to-set metrics for evaluation of differences between query and support set. Third, it presents extensive experiments on three few-shot datasets, namely miniImageNet, tieredImageNet and CUB. In almost all cases, our method outperforms the state-of-the-art. Notably, our method gains 1.83\%, 1.42\%, and 1.83\% accuracy in 1-shot over the baselines in miniImageNet, tieredImageNet and CUB, respectively. 

\section{Related work}
\label{sec:related-work} 
Our work falls within the domain of inductive few-shot image classification~\cite{vinyals2016matching, ravi2016optimization, snell2017prototypical, finn2017model}, unlike transductive methods~\cite{dhillon2019baseline, NEURIPS2019_01894d6f, boudiaf2020transductive, zhang2021prototype} which exploit the structural information of the entire novel set.
%
The following covers the most relevant works under few-shot learning research area, and beyond.  

\paragraph{Training framework}
Two main training frameworks have been explored so far, namely meta learning or standard transfer learning. On one hand, meta learning~\cite{vinyals2016matching,ravi2016optimization,snell2017prototypical,finn2017model}, also named episodic training, repeatedly samples small subsets of base classes to train the network, thereby simulating few-shot ``episodes'' during training. For example, some methods (e.g.~\cite{finn2017model,kim2018bayesian,antoniou2018train}) aim at training a model to classify the novel classes with a small number of gradient updates. On the other hand, standard transfer learning methods~\cite{gidaris2018dynamic,Qi_2018_CVPR,chen2019closer,Tian_2020_ECCV_good,Afrasiyabi_2020_ECCV} usually rely on a generic batch training with a metric-based (such as margin-based) criteria. Recently, several works~\cite{Ye_2020_CVPR, xu2021learning, Zhang_2020_CVPR} have shown that combining both standard transfer learning, following by a second meta-training stage can offer good performance. We employ a similar two-stage training procedure in this paper. 
%
%
\paragraph{Metrics}
Metric-based approaches~\cite{vinyals2016matching,snell2017prototypical,bertinetto2018metalearning,garcia2017few,kim2019variational,Li_2019_CVPR,lifchitz2019dense,oreshkin2018tadam,sung2018learning,tseng2020cross,wertheimer2019few,Zhang_2020_CVPR,zhang2019variational, rizve2021exploring} aim at improving how the distance is calculated for better performance at training and inference. In this aspect, our work is related to ProtoNet~\cite{snell2017prototypical} as it also seeks to reduce the distance between a query and the centroid of a set of support examples of the corresponding class, while differing by proposing the use of \emph{set-to-set} distance metrics for computing distance over several feature vectors. Other highly related works include FEAT~\cite{Ye_2020_CVPR}, CTX~\cite{doersch2020crosstransformers}, TapNet~\cite{yoon2019tapnet} and ConstellationNet~\cite{xu2020attentional}, which apply attention embedding adaptation functions on the episodes before computing the distance between query and the prototypes of the support set. Unlike them, our method extracts a set of feature vectors given a query and support set, over which a set-to-set metric is applied for computing the distances.

\paragraph{Extra data}
Relying on extra data~\cite{Chen_2019_CVPR,Chu_2019_CVPR,gao2018low,gidaris2019boosting,gidaris2019generating,hariharan2017low,Liu_2019_ICCV,mehrotra2017generative,ren2018meta,schwartz2018delta,wang2018low,wang2016learning,zhang2017mixup,Zhang_2019_CVPR, zhang2020iept} is another strategy for building a well-generalized model. The augmented data can be in the form of hallucination with a data generator function~\cite{hariharan2017low, wang2018low}, using unlabelled data under semi-supervised~\cite{ren2018meta, Yu_2020_CVPR} or self-supervised~\cite{su2019does, gidaris2019boosting} frameworks, or aligning the novel classes to the base data~\cite{Afrasiyabi_2020_ECCV}. In contrast, our approach does not require any additional data beyond the base classes. 

\paragraph{Vision transformers}
Our method employs shallow attention mappers that are inspired by the multi-head attention mechanism proposed in~\cite{vaswani2017attention} and adapted to images by Dosovitskiy~\etal~\cite{dosovitskiy2020image}. In contrast to these works, our feature mappers are independent, are shallow (thus lightweight), are not unified by FC-layers, and can extend to convolutions as in~\cite{wu2021cvt, graham2021levit}. We also employ several independent mappers at different depths/scales in the network. 





\paragraph{Feature sets}
Feature sets have long been investigated in computer vision~\cite{lazebnik2006beyond,quelhas2005modeling,hong2009sigma}. 
In the more recent deep learning literature, our method bears resemblance to FPN~\cite{lin2017feature} which extracts multi-scale features for object detection, and deep sets~\cite{zaheer2017deep} which proposed permutation-invariant networks that operate on \emph{input} sets. In contrast, our work computes feature sets to tackle few-shot image classification.
%
%
From the few-shot perspective, our work is related to the transductive approach of \cite{NEURIPS2019_01894d6f}, which employs the unlabeled query set to augment the support set, and \cite{Zhang_2020_CVPR} which uses Earth Mover's Distance over the representations of multi-cropped images with a generic data augmentation. Here, we focus on the extraction of feature sets for each support example in an inductive setting.


\section{Preliminaries}
\label{sec:approach_overview} 

In $N$-way $K$-shot (where $K$ is small) image classification, we aim to predict the class of a given query example $\mathbf{x}_q$ using a support set $\mathcal{S}$ containing $K$ examples for each of the $N$ different classes considered.
Let $\mathcal{S}^n\in\mathcal{S}$ and $\mathcal{S}^n=\{(\mathbf{x}^n_i, y_i=n)\}_{i=1}^{K}$ be a set of example-label pairs, all pairs of that set $\mathcal{S}^n$ belonging to class $n$. In addition, let $f(\mathbf{x}|\theta^f)$ be a convolutional feature extractor composed of $B$ blocks parameterized by $\theta^{f} = \{\theta^f_b\}_{b=1}^B$, where $\theta^f_b$ are the parameters of the $b$-th block. Here, a ``block'' broadly refers to a group of convolutional layers (with or without skip links), typically followed by a downscaling operation reducing the features spatial dimensions (\eg, pooling). The features after a given block $b$ can be obtained from $\mathbf{z}_b \equiv f(\mathbf{x}|\{\theta^f_i\}_{i=1}^b)$.  

In this work, we introduce a \emph{set-feature} extractor, dubbed ``SetFeat'', which extracts a set of $M$ feature vectors from images, rather than a single vector as it is typically done in the literature~\cite{vinyals2016matching, snell2017prototypical, finn2017model}. Formally, SetFeat produces the set $\mathcal{H} = \{ \mathbf{h}_m \}_{m=1}^{M}$ of $M$ feature vectors $\mathbf{h}_m$ through shallow self-attention mappers, and employs set-to-set matching metrics to establish the similarity between images in set-feature space. The following section presents our approach in detail. 

 
\section{Set-based few-shot image classification}
\label{sec:set_based_fs} 

\begin{figure} 
    \centering
    \footnotesize
    \setlength{\tabcolsep}{1pt}
    \begin{tabular}{ccc} 
    \includegraphics[width=0.52\linewidth]{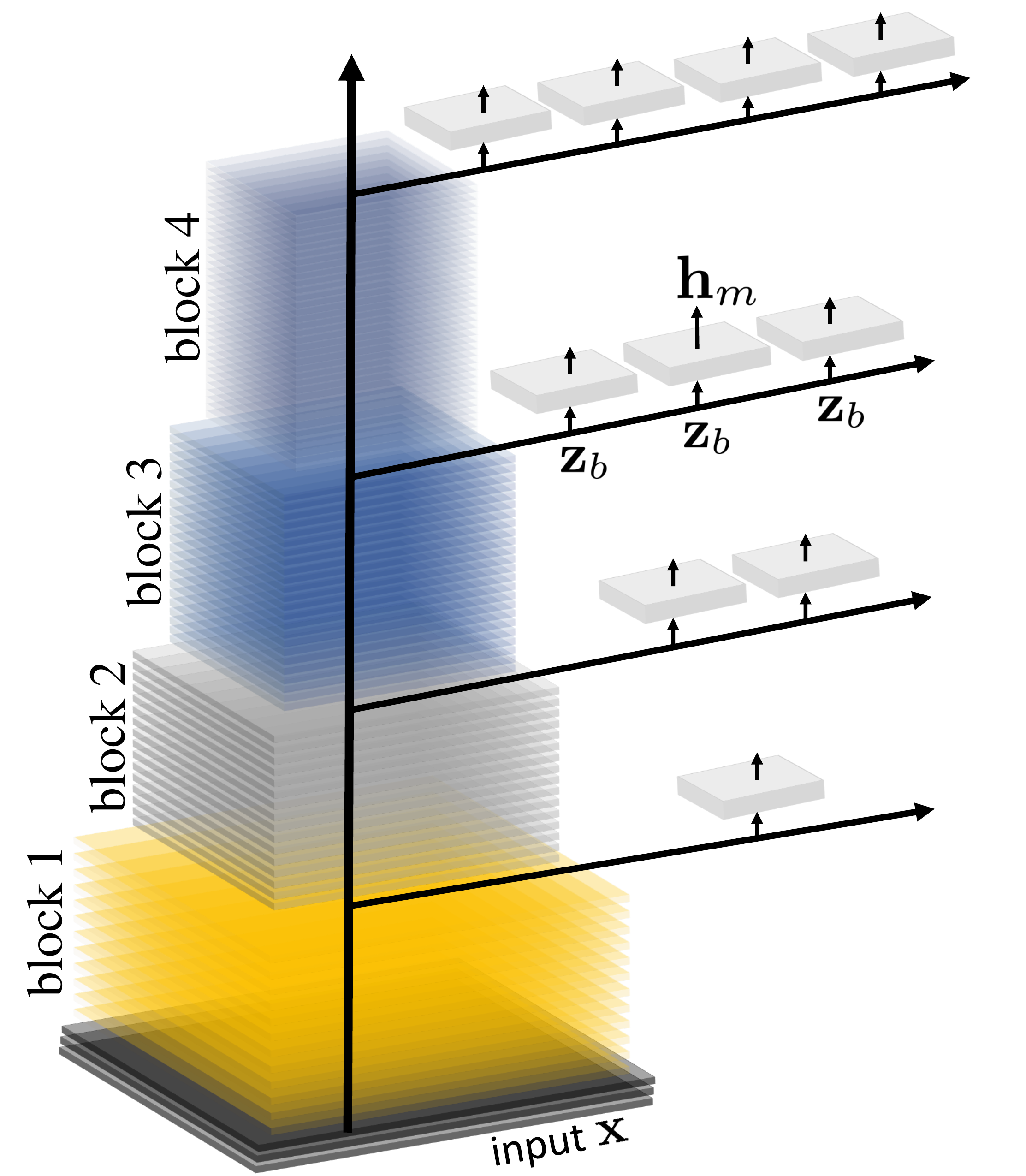} & 
    \includegraphics[width=0.45\linewidth]{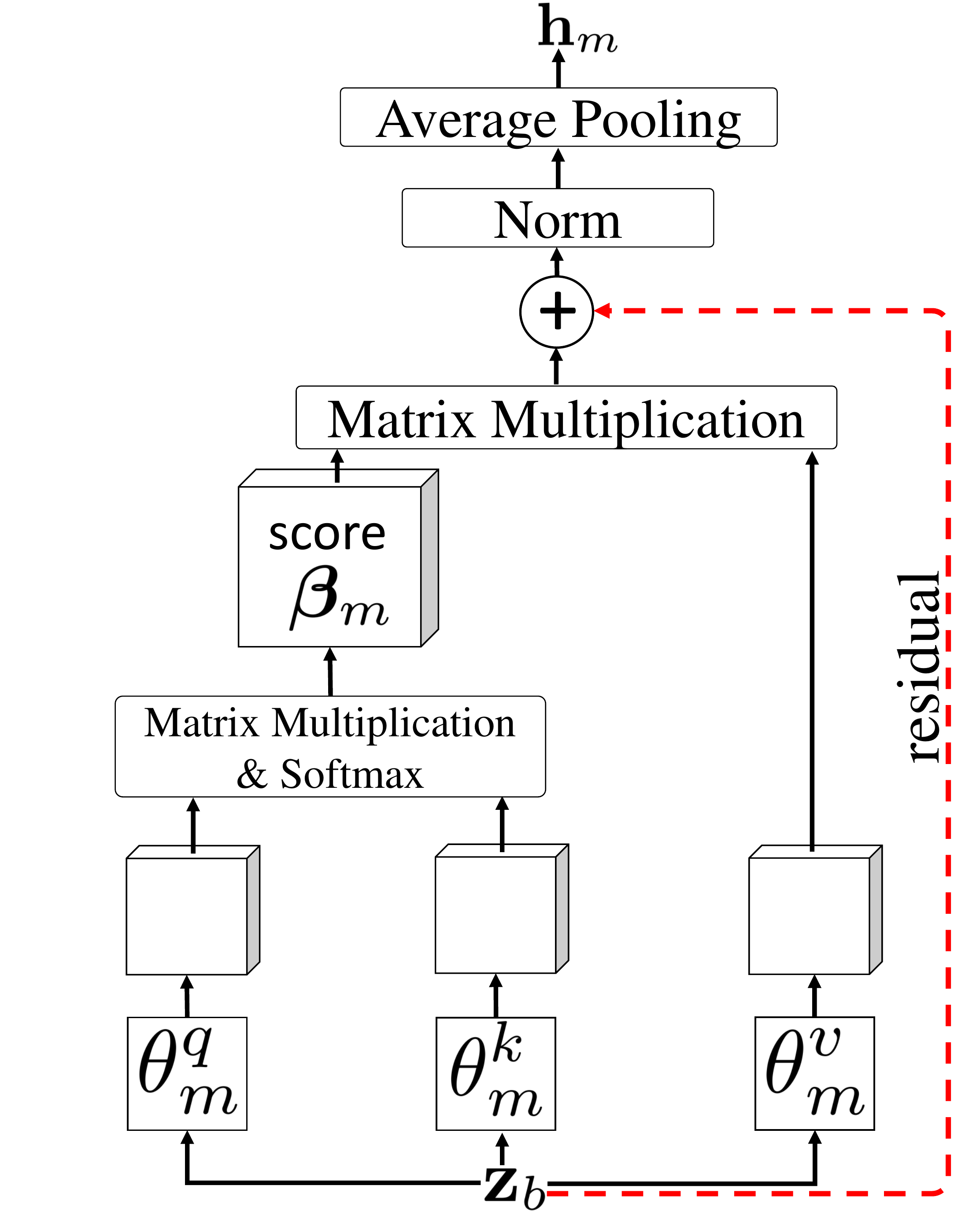} &   \\ 
    (a) Set-feature extractor (SetFeat)   & 
    (b) Mapper $g(\cdot|\theta^g_m)$   \\       
    \end{tabular}
    \vspace{0.5em}
    \caption{The schematic overview of the proposed set-feature extractor (SetFeat) and detail of a single attention-based mapper: (a) given an input $\mathbf{x}$, SetFeat first extracts (convolutional) feature vectors $\mathbf{z}_b$ at each of its blocks, while at each block attention-based mappers (illustrated as small rectangles) convert $\mathbf{z}_b$ into a different embedding $\mathbf{h}_m$; (b) a single mapper $m$ at block $b$ extracts embedding $\mathbf{h}_m$ using an attention mechanism containing query $\theta^q_m$ and key $\theta^k_m$ to build attention scores $\bm{\beta}_m$, with self-attention inferred using value $\theta^v_m$ and score $\bm{\beta}_m$. This work focuses on backbones made of $B=4$ blocks, consistent with popular few-shot image classification backbones such as Conv4~\cite{vinyals2016matching} and ResNet~\cite{he2016deep}.  
    } 
    \label{fig:overall_fig}
\end{figure}

In this section, we first discuss our proposed set-feature extractor SetFeat, then dive into the details of our proposed set-to-set metrics. Finally, our proposed inference and training procedures are presented.

\subsection{Set-feature extractor}
\label{sec:setfeat} 

The overall architecture of SetFeat is illustrated in \cref{fig:overall_fig}. As mentioned in \cref{sec:approach_overview}, its goal is to map an input image $\mathbf{x}$ to a feature set $\mathcal{H}$. 
To this end, and inspired by \cite{vaswani2017attention,dosovitskiy2020image}, we embed segregated self-attention mappers $g(\cdot)$ throughout the network, as shown in \cref{fig:overall_fig}a. We reiterate (\cf \cref{sec:related-work}), however, that our mappers are different from multi-head attention-based models~\cite{vaswani2017attention, dosovitskiy2020image} for two main reasons. First, each mapper in our approach is composed of a single attention head, thus we do not rely on fully connected layers to concatenate multi-head outputs. Our feature mappers are therefore separate from each other and each extract their own set of features. Second, our feature mappers are shallow (unit depth), with the learning mechanisms relying on the convolutional layers of the backbone. 


%
%


The detail of the $m$-th feature mapper $g(\mathbf{z}_{b_m}|\theta^g_m)$, where $b_m$ represents the block preceding the mapper, is illustrated in \cref{fig:overall_fig}b. 
The learned representation $\mathbf{z}_{b_m} \in \mathbb{R}^{P\times D^p}$ is separated into $P$ non-overlapping patches of $D^p$ dimensions.
In this work, we use patches of size $1\times1$, each patch is therefore a 1-D vector of $D^p$ elements.
%
%
Following Vaswani~\etal~\cite{vaswani2017attention}, an attention map is first computed using two parameterized elements $q(\mathbf{z}_{b_m}|\theta^q_m)$ and $k(\mathbf{z}_{b_m}|\theta^k_m)$:
\begin{equation} 
\label{eq:}  
     \bm{\beta}_m = \text{Softmax}\left(q(\mathbf{z}_{b_m}|\theta^q_m) k(\mathbf{z}_{b_m}|\theta^k_m)^\top/{\sqrt{d_k}}\right) \, , 
\end{equation}  
where $\bm{\beta}_m \in \mathbb{R}^{P\times P}$ is the attention score over the patches of $\mathbf{z}_{b_m}$, and $\sqrt{d_k}$ is the scaling factor.  
Then, we compute the dot-product attention over the patches of $\bm{\beta}_m$ using $v(\mathbf{z}_{b_m}|\theta^{v}_m)$ in the following form:
\begin{equation} 
      \mathbf{a}_m = \bm{\beta}_m \, v(\mathbf{z}_{b_m}|\theta^{v}_m) \, ,
\end{equation} 
where $\mathbf{a}_m\in\mathbb{R}^{P\times D^a}$ consists of $P$ patches of $D^a$ dimensions and $D^a$ is the dimension of $z_b$. If the backbone feature extractor is ResNet~\cite{he2016deep} (see \cref{sec:backbones}), we add a residual to the computed attention ($\mathbf{a}_m + \mathbf{z}_{b_m}$). In this case, if the dimensions mismatch ($D^a \neq D^p$), we use $1\times1$ convolution of unit stride and kernel size similar to downsampling. Finally, the feature vector $\mathbf{h}_m$ is computed by taking the mean of over the patches (over the $P$ dimension).

\subsection{Set-to-set matching metrics}
\label{sec:set2set}

Having covered how SetFeat extracts a feature set for each input instance to process, we now proceed to how it leverages this set for image classification. In this context, we need to compare the feature set of the query with the feature sets corresponding to each instance of the support set of each class, to infer the class of the query.
More specifically, in order to proceed with a distance-based approach as we do with prototypical networks, we need a set-to-set metric allowing the measure of distance over sets. We now present three distinct set-to-set metrics $d_\mathrm{set}(\mathbf{x}_q, \mathcal{S}^n)$, which measure the distance between multiple feature sets, where $\mathbf{x}_q$ is the query, and $\mathcal{S}^n$ is a support set for class $n$ (\cf \cref{sec:approach_overview}). We employ the shorthand $\mathbf{h}_m(\mathbf{x}) \equiv g_m(\mathbf{z}_{b_m}|\theta^g_m)$ to refer to a feature extracted by mapper $m$. In addition, we also define $\bar{\mathbf{h}}_m(\mathcal{S}) \equiv \frac{1}{|\mathcal{S}|}\sum_{\mathbf{x} \in \mathcal{S}}  \mathbf{h}_m(\mathbf{x})$ as the centroid of features extracted by mapper $m$ on the support set $\mathcal{S}$. 
The following set metrics are built upon a generic distance function $d(\cdot, \cdot)$. In practice, we employ the negative cosine similarity function, \ie, $d(\cdot, \cdot) = -\cos(\cdot, \cdot)$.

\begin{figure}[t] 
    \centering
    \footnotesize
    \setlength{\tabcolsep}{1pt}
    \begin{tabular}{cccc} 
    \rotatebox{90}{\footnotesize{FEAT~\cite{Ye_2020_CVPR}} \quad \ \  \ \ \ \ \ \ \  \footnotesize{MN~\cite{vinyals2016matching}} \quad  \ \ \ \  \ \footnotesize{PN~\cite{snell2017prototypical}} }  
    \includegraphics[width=0.45\linewidth, angle=0]{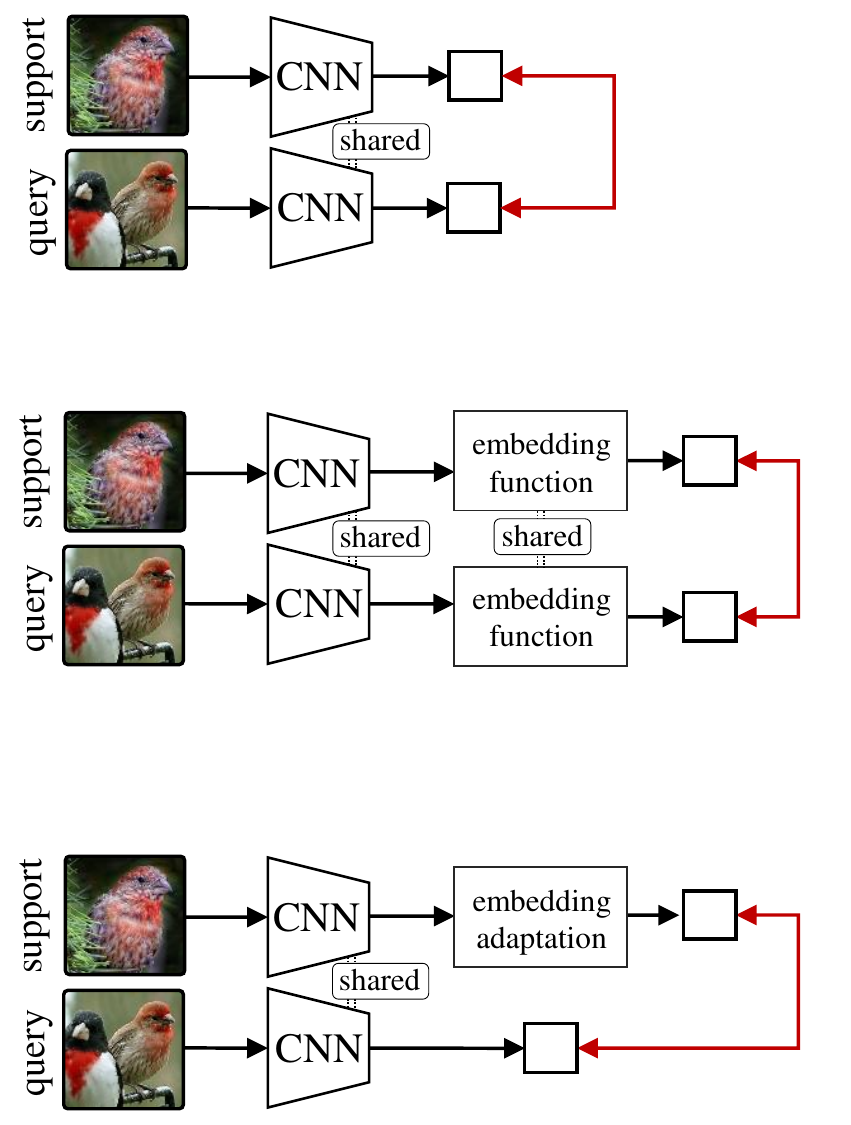} &   
    \includegraphics[width=0.51\linewidth, angle=0]{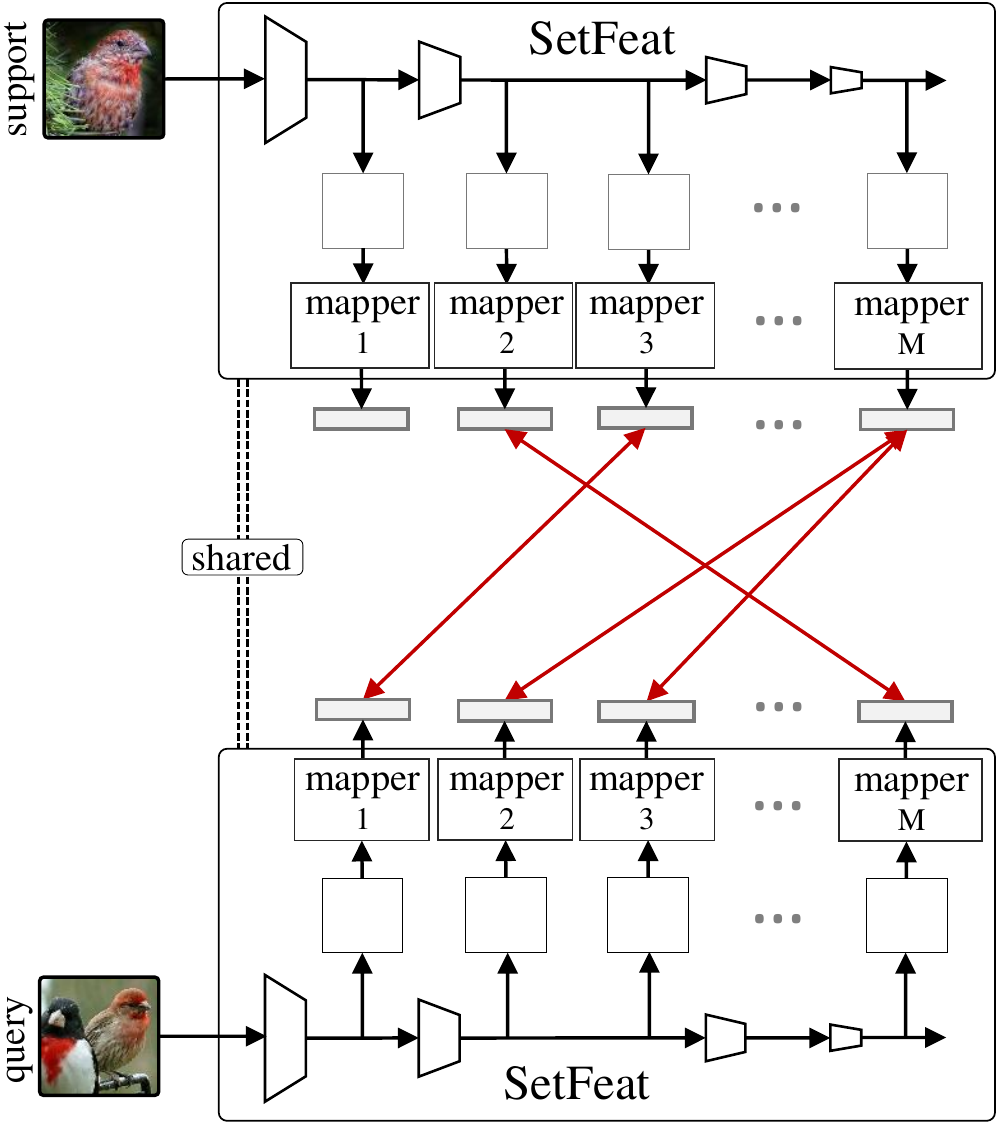}\\ 
    (a) Three baseline methods   & \quad   (b) Set matching with sum-min \\       
    \end{tabular}
    \vspace{0.5em}
    \caption[]{Illustration of 1-shot image classification using (a) three existing methods and (b) our approach with the sum-min metric. (a) Given a query and support, existing methods either directly match the query to support (ProtoNet (PN)~\cite{snell2017prototypical}), apply a single embedding function over both support and query (MatchingNetwork (MN)~\cite{vinyals2016matching}), or perform embedding adaptation on the support before matching it with the query (FEAT~\cite{Ye_2020_CVPR}). (b) Our SetFeat method extracts sets of features for both of the support and query, which are then processed by the self-attention mappers. The set metric is then computed over the embeddings. }
    \label{fig:set2set_cartoon}
\end{figure}

\paragraph{Match-sum} aggregates the distance between matching mappers for the query and supports: 
\begin{equation}
    d_\mathrm{ms}(\mathbf{x}_q, \mathcal{S}^n) = \sum_{i=1}^M
    d \left( \mathbf{h}_i(\mathbf{x}_q), \bar{\mathbf{h}}_i(\mathcal{S}^n)\right).
\end{equation} 
We use this metric as a baseline, as it parallels a common strategy of building representations simply by concatenating several feature vectors and invoking a standard metric on the flattened feature space.

\paragraph{Min-min} uses the minimum distance across all possible pairs of elements from the query and support set centroids: 
\begin{equation}
    d_\mathrm{mm}(\mathbf{x}_q, \mathcal{S}^n) = 
    \min_{i=1}^M \min_{j=1}^M
    d \left( \mathbf{h}_i(\mathbf{x}_q), \bar{\mathbf{h}}_j(\mathcal{S}^n)\right).
    \label{eq:min-min}
\end{equation} 
Such a metric leverages directly the set structure of features. 

\paragraph{Sum-min} departs from the min-min metric by aggregating with a sum the minimum distances between the mappers computed on query and support set centroids: 
\begin{equation}
    d_\mathrm{sm}(\mathbf{x}_q, \mathcal{S}^n) = \sum_{i=1}^M \min_{j=1}^M
    d \left( \mathbf{h}_i(\mathbf{x}_q), \bar{\mathbf{h}}_j(\mathcal{S}^n)\right) \,.
    \label{eq:sum-min}
\end{equation} 
A schematic illustration of the sum-min metric is shown in \cref{fig:set2set_cartoon}, which also illustrates its difference with respect to three baseline few-shot models.  Our method is different from FEAT~\cite{Ye_2020_CVPR} (and MN~\cite{vinyals2016matching}) in two main ways. First, we define sets over features extracted from each example while FEAT/MN do so over the support set directly. In an extreme 1-shot case, the FEAT ``set'' degenerates to a single element (1 support). Beneficial for few-shot, our work always keeps sets of many elements, regardless of the support set cardinality.
Second, our method employs the parameterized mappers for set feature extraction. Here, we adjust the backbone (unlike FEAT and MN) so that adding mappers results in the same total number of parameters. Third, our method employs non-parametric set-to-set metrics, used for inference. 

\subsection{Inference} 


%
Given one of our metrics $d_\mathrm{set} \in \{d_\mathrm{ms}, d_\mathrm{mm}, d_\mathrm{sm}\}$ defined in the previous section, we follow the approach of Prototypical Networks~\cite{snell2017prototypical} with SetFeat and model the probability of a query example $\mathbf{x}_q$ belonging to class $y=n$, where $n \in \{1, \ldots, C\}$ ($N$-way), using a softmax function: 
\begin{equation} 
    \label{eq:inference}
    p(y = n| \mathbf{x}_{q}, \mathcal{S}) = \frac{\exp (-d_\mathrm{set}(\mathbf{x}_{q},\mathcal{S}^n))}{ \sum_{  \mathcal{S}^i\in\mathcal{S}}   \exp (-d_\mathrm{set}(\mathbf{x}_{q},\mathcal{S}^i))} \,,
\end{equation} 
with $\mathcal{S}$ as the (few-shot) support set.

\begin{algorithm}[t]
    \SetAlgoLined
    {\small
        \KwData{ Network parameterized by $\theta=\{\theta^f,\theta^g\}$ made of a backbone of $B$ convolution blocks ($\theta^f=\{\theta^{f}_{b}\}_{b=1}^B$) and $M$ mappers ($\theta^g=\{\theta^g_m\}_{m=1}^M$); episodic train dataset $\mathcal{X}_{\text{train}}$ containing episodes of support set $\mathcal{S}$ and a query example $\mathbf{x}_q$; validation dataset $\mathcal{X}_{\text{valid}}$; maximum number of epochs $t^\text{max}$; 0-1 loss function $\ell_{0-1}$ used to measure the validation accuracy.}
        \KwResult{Best model defined as $\theta_\text{best}=\{\theta^f_\text{best},\theta^g_\text{best}\}$}
        $E^\text{best}_\text{valid} \leftarrow \infty$\\ 
        \For{$t=1,\ldots,t^\text{\emph{max}}$} {  
            \For{$(\mathbf{x}_q,\mathcal{S})\in  \mathcal{X}_\text{\emph{train}} $}{
                $\ell^t \leftarrow - \log p(y_q|\mathbf{x}_q,\mathcal{S})$ \quad using eq.~\ref{eq:inference}\\
                Update network $\theta$ with backpropagation of loss $\ell^t$\\
            }
                $\hat{y}_q \leftarrow {\arg\min}_{\mathcal{S}^n\in\mathcal{S}} d_\text{set}(\mathbf{x}_q,\mathcal{S}^n),~~\forall (\mathbf{x}_q,\mathcal{S}) \in \mathcal{X}_\text{valid}$\\
                $E_\text{valid} \leftarrow \frac{1}{|\mathcal{X}_\text{valid}|} \sum_{(\mathbf{x}_q,\mathcal{S}) \in \mathcal{X}_\text{valid}} \ell_{0-1}(\hat{y}_q, y_q)$\\
            \If{$E_\text{\emph{valid}} < E^\text{\emph{best}}_\text{\emph{valid}}$}{
                $E^\text{best}_\text{valid} \leftarrow E_\text{valid}$\\
                $\theta_\text{best} \leftarrow \theta$
            } 
        }

    } 
         
    
    \caption{SetFeat meta-training and validation.}
    \label{algo:metatraining}
\end{algorithm}

\subsection{Training procedure} 

We follow recent literature~\cite{Ye_2020_CVPR, xu2021learning, Zhang_2020_CVPR} and leverage a two-stage procedure to train SetFeat using one of our proposed set-to-set metrics. The first stage performs standard pre-training where a random batch $\mathcal{X}_\text{batch}$ of instances $\mathbf{x}$ from base classes are drawn from the training set. Here, we append fully-connected (FC) layers $\mathbf{o}_m$ to convert each of the mapper features $\mathbf{h}_m$ into logits in order to achieve classification over the $C$ classes. From that, cross-entropy loss is used to train each mapper independently:
\begin{equation}
\ell_{\text{pre}} = -\sum_{ \mathbf{x}_i\in\mathcal{X}_\text{batch}} \sum_{m=1}^{M} \log \frac{\exp(o_{m,y_i}(\mathbf{h}_{m,i}))}{\sum_{c=1}^C \exp(o_{m,c}(\mathbf{h}_{m,i}))} \,,
\label{eq:pretrainloss}
\end{equation}
where $o_{m,c}$ is the FC layer output of mapper $m$ for class $c$, $\mathbf{h}_{m,i}$ is the feature set of mapper $m$ for instance $\mathbf{x}_i$, and $y_i$ is the target output corresponding to instance $\mathbf{x}_i$.

The second stage discards the FC layers that were added in the first stage, and employs episodic training~\cite{vinyals2016matching, snell2017prototypical} which simulates the few-shot scenario on the base training dataset.  This stage is presented in \cref{algo:metatraining}. Specifically, we randomly sample $N$-way $K$-shot and $Q$-queries, then we compute the probability scores for each query using eq.~\ref{eq:inference}. Finally, we update the parameters of the network after computing the cross-entropy loss.

\section{Evaluation}
\label{sec:evaluation}

This section first covers the details of our experiments with SetFeat, which are based on conventional backbones employed in the few-shot image classification literature. This is followed by description of the datasets and implementation details are described next. Finally, we present the evaluations of SetFeat with our set-matching metrics using four backbones with three datasets.

\subsection{Backbones}
\label{sec:backbones}

We adopt the following three popular backbones, each composed of four blocks: (a) Conv4-64~\cite{vinyals2016matching}, which consists of 4 convolution layers with 64/64/64/64 filters for a total of 0.113M parameters, (b) Conv4-512~\cite{vinyals2016matching}: 96/128/256/512 for 1.591M parameters, and (c) ResNet12~\cite{he2016deep, oreshkin2018tadam}: 64/160/320/640 for 12.424M parameters. In all experiments below, we embed a total of 10 self-attention mappers throughout each backbone by following this per-block pattern: 1 mapper after block 1, then 2, 3 and 4 mappers for the three subsequent blocks. We experiment with other choices of mapper configurations in \cref{sec:ablation-mappers}.

Since our attention-based feature mappers require additional parameters, we correspondingly reduce the number of kernels in the backbone feature extractors to ensure that the performance gains are not simply due to the over-parameterization. Specifically, our SetFeat4-512, the counterpart of Conv4-512, uses a reduced set of 96/128/160/200 convolution kernels for a total of 1.583M parameters (compared to 1.591M for Conv4-512). SetFeat12, counterpart of ResNet12, consists of 128/150/180/512 kernels for 12.349M parameters (comp. 12.424M for ResNet12). For Conv4-64, reducing the amount of parameters collapses the training (as noted in\cite{fei2020melr, Ye_2020_CVPR, wu2019parn}) since it contains very few parameters already. Our SetFeat4-64 therefore has more parameters (0.238M vs 0.113M for Conv4-64), but in \cref{SetFeat4_overparam_ablation} we artificially augment the number of parameters for Conv4-64 and show our approach still outperforms it.

Convolutional attention~\cite{wu2021cvt} is used in SetFeat4-512 and SetFeat12. Particularly, we used single depth convolution and batch normalization to parameterize key, query and value in each mapper. The output dimension of the feature mappers is set to the number of channels in the last layer of the feature extractor --- having all mappers producing feature vectors of the same dimension is a necessary condition for our proposed metrics. 
For SetFeat4, FC-layers are used to compute the attention in order to limit the number of additional parameters as much as possible. The supplementary material includes the details of our implementation.   

\begin{table}[t]
\renewcommand{\tabcolsep}{2pt}
\centering 
\caption{Evaluation on miniImageNet in 5-way. Bold/blue is best/second, and $\pm$ is the 95\% confidence intervals in 600 episodes.} 
\begin{tabular}{llccccccc}
    \bottomrule  
        & \textbf{Method}   
        & \textbf{\footnotesize Backbone} 
        & \textbf{1-shot}  
        & \textbf{5-shot}    
        \\ 
       \midrule
            & ProtoNet~\cite{snell2017prototypical}   
            & \multirow{11}{*}{\rotatebox{90}{------------ Conv4-64 ------------}}  
            & 49.42\scriptsize{$\pm$0.78}      & 68.20\scriptsize{$\pm$0.66}  
            \\
                     
            & MAML~\cite{finn2018probabilistic}       
            & 
            & 48.07\scriptsize{$\pm$1.75}      & 63.15\scriptsize{$\pm$0.91}  
            \\
                     
            & RelationNet~\cite{sung2018learning}    
            & 
            & 50.44\scriptsize{$\pm$0.82}      & 65.32\scriptsize{$\pm$0.70}    
            \\  
                    
            & Baseline++~\cite{chen2019closer}    
            & 
            & 48.24\scriptsize{$\pm$0.75} & 66.43\scriptsize{$\pm$0.63}      
            \\
                  
            & IMP~\cite{allen2019infinite}
            & 
            & 49.60\scriptsize{$\pm$0.80}    & 68.10\scriptsize{$\pm$0.80}  
            \\ 
            & MemoryNet~\cite{Cai_2018_CVPR}
            & 
            & 53.37\scriptsize{$\pm$0.48}    & 66.97\scriptsize{$\pm$0.35}  
            \\ 
            & Neg-Margin~\cite{Bin_2020_ECCV_margin_matter}       
                & 
                & 52.84\scriptsize{$\pm$0.76}      & 70.41\scriptsize{$\pm$0.66} \\
           
            & MixtFSL~\cite{afrasiyabi2021MixtFSL}  
                & 
                & 52.82\scriptsize{$\pm$0.63}      & 70.67\scriptsize{$\pm$0.57}  \\

            & FEAT~\cite{Ye_2020_CVPR} 
                & 
                & 55.15\scriptsize{$\pm$0.20}      & 71.61\scriptsize{$\pm$0.16 }   \\
                
            & MELR~\cite{fei2020melr}   
                & 
                & {55.35}\scriptsize{$\pm$0.43}      & {72.27}\scriptsize{$\pm$0.35}  \\
            
            & BOIL~\cite{oh2021boil} 
                & \multirow{5}{*}{\rotatebox{90}{SF4-64\hspace*{1.5em}}}
                & 49.61\scriptsize{$\pm$0.16}      & 66.45\scriptsize{$\pm$0.37}    
                \\*[0.5em]

            \multirow{3}{*}{\rotatebox{90}{-- Ours --}} 
                
            &  Match-sum  
                &      
                &  55.74\scriptsize{$\pm$0.65}       &72.18\scriptsize{$\pm$0.70}  \\ 
               
            &  Min-min 
                &       
                & {\color{red}56.22}\scriptsize{$\pm$0.89}        &{\color{red}72.70}\scriptsize{$\pm$0.65}  \\ 
               
            &   Sum-min 
                &     
                & \textbf{57.18}\scriptsize{$\pm$0.89}        &\textbf{73.67}\scriptsize{$\pm$0.71}  
               \\*[0.2em]

        

      \midrule
            & ProtoNet$^\dag$~\cite{snell2017prototypical}   
            & \multirow{6}{*}{\rotatebox{90}{--- Conv4-512 ---}}   
            & 53.52\scriptsize{$\pm$0.43}      & 73.34\scriptsize{$\pm$0.36}  
            \\    
            
            & MAML~\cite{finn2017model}   
            & 
            & 49.33\scriptsize{$\pm$0.60}      & 65.17\scriptsize{$\pm$0.49}  
            \\    
 
            & Relation Net~\cite{sung2018learning}   
            & 
            & 50.86\scriptsize{$\pm$0.57}      & 67.32\scriptsize{$\pm$0.44}  
            \\     
            
            & PN+rot~\cite{gidaris2018dynamic}   
            & 
            & 56.02\scriptsize{$\pm$0.46}      & 74.00\scriptsize{$\pm$0.35}  
            \\      
            
            & CC+rot~\cite{gidaris2019boosting}   
            & 
            & 56.27\scriptsize{$\pm$0.43}      & 74.30\scriptsize{$\pm$0.33}  
            \\
            
             & MELR~\cite{fei2020melr}   
            & \multirow{5}{*}{\rotatebox{90}{\hspace*{-0.5em}SF4-512\hspace*{1.5em}}}
            & 57.54\scriptsize{$\pm$0.44}      & {\color{red}74.37}\scriptsize{$\pm$0.34}  
            \\*[0.8em]  
            
            \multirow{3}{*}{\rotatebox{90}{-- Ours --}}  
            & Match-sum  
                &  
                & 56.50\scriptsize{$\pm$0.85}      & 72.69\scriptsize{$\pm$0.68}   \\  
                
            &  Min-min  
                &  
                & {\color{red}58.57}\scriptsize{$\pm$0.87}      &  73.46\scriptsize{$\pm$0.68}   \\

            &  Sum-min  
                & 
                & \textbf{59.10}\scriptsize{$\pm$0.87}      & \textbf{74.97}\scriptsize{$\pm$0.66}   
                \\*[0.2em]

        \midrule 
                    
                    
                    
                & AdaResNet~\cite{munkhdalai2018rapid}  
                    & \multirow{12}{*}{\rotatebox{90}{------------ ResNet12 ------------}}  
                    & 56.88\scriptsize{$\pm$0.62}            & 71.94\scriptsize{$\pm$0.57}  
                    \\
                    
                & TADAM~\cite{oreshkin2018tadam}     
                    & 
                    & 58.50\scriptsize{$\pm$0.30}            & 76.70\scriptsize{$\pm$0.30}  
                    \\
                    
                & MetaOptNet~\cite{lee2019meta}      
                    & 
                    & 62.64\scriptsize{$\pm$0.61}               & 78.63\scriptsize{$\pm$0.46} 
                    \\
                     
                    
                  
                & Neg-Margin~\cite{Bin_2020_ECCV_margin_matter}      
                    & 
                    & 63.85\scriptsize{$\pm$0.76}      & 81.57\scriptsize{$\pm$0.56}      
                     \\

                & MixtFSL~\cite{afrasiyabi2021MixtFSL}    
                    & 
                    & 63.98\scriptsize{$\pm$0.79}       & 82.04\scriptsize{$\pm$0.49} \\
                    
                & Meta-Baseline~\cite{chen2021meta}  
                    & 
                    &  63.17\scriptsize{$\pm$0.23}      & 79.26\scriptsize{$\pm$0.17}  \\

                & Distill~\cite{Tian_2020_ECCV_good}     
                 & 
                   & 64.82\scriptsize{$\pm$0.60}       & 82.14\scriptsize{$\pm$0.43}  \\     
                   
                   
                & DeepEMD~\cite{Zhang_2020_CVPR} 
                 & 
                   & 65.91\scriptsize{$\pm$0.82}       & 82.41\scriptsize{$\pm$0.56}   \\
         
                & DMF~\cite{xu2021learning}  
                    & 
                    & 67.76\scriptsize{$\pm$0.46}       & {\color{red}82.71}\scriptsize{$\pm$0.31}   \\

                & MELR~\cite{fei2020melr} 
                    & 
                    & 67.40\scriptsize{$\pm$0.43}       & \textbf{83.40}\scriptsize{$\pm$0.28} \\
                
                & ProtoNet$^{\S}$~\cite{snell2017prototypical}    
                    & 
                    & 62.39\scriptsize{\quad \quad \ \ \ }       & 80.53\scriptsize{\quad \quad \ \ \ }    
                \\
                & FEAT$^{\S}$~\cite{Ye_2020_CVPR} 
                    & \multirow{5}{*}{\rotatebox{90}{- SF-12 -\hspace*{1.5em}}}
                   & 66.78\scriptsize{\quad \quad \ \ \ }                 & 82.05\scriptsize{\quad \quad \ \ \ }  
   
            \\*[0.5em]  
            \multirow{3}{*}{\rotatebox{90}{-- Ours --}}  
            &  Match-sum 
                &    
                & 67.41\scriptsize{$\pm$0.64}        & 81.79\scriptsize{$\pm$0.55}  \\ 
                
            &  Min-min 
                &  
                & {\color{red}67.88}\scriptsize{$\pm$0.55}      & 82.07\scriptsize{$\pm$0.61}  \\

            &  Sum-min  
                & 
                & \textbf{68.32}\scriptsize{$\pm$0.62}      & {\color{red}82.71}\scriptsize{$\pm$0.46} 
               \\*[0.2em]

            \bottomrule    
    \end{tabular}  \\
    {\footnotesize $^\S$confidence interval not provided  \quad $^\dag$ taken from~\cite{fei2020melr}}
    
    {\footnotesize Mappers dimension: SF4-64$\in\mathbb{R}^{64}$, SF4-512$\in\mathbb{R}^{200}$, SF12$\in\mathbb{R}^{512}$}
    \label{tab:miniImageNet} 
\end{table}

\subsection{Datasets and implementation details}

We conduct experiments on  miniImageNet~\cite{vinyals2016matching} (100/50/50 train/validation/test classes), 
tieredImageNet~\cite{ren2018meta} (351/97/160) for object recognition, and CUB~\cite{wah2011caltech} (100/50/50) for fine-grained classification.  
To pretrain SetFeat4, we used Adam~\cite{oreshkin2018tadam} with a learning rate (lr) of 0.001 and weight decay of $5\times 10^{-4}$.  Batch size is fixed to 64. For SetFeat12, we used Nesterov momentum with an initial lr of 0.1, momentum of 0.9 and weight decay of $5\times 10^{-4}$. We follow~\cite{xu2021learning, Zhang_2020_CVPR, Ye_2020_CVPR} for normalization and data augmentation. In the meta-training stage, SGD is used for all architectures. Validation sets are used to tune the schedule of the optimizer.

\subsection{Quantitative and comparative evaluations}
\label{sec:quant-evaluation}

\paragraph{miniImageNet} Table~\ref{tab:miniImageNet} presents evaluations of SetFeat with our set-to-set metrics on the miniImageNet dataset. First, we observe that our sum-min metric outperforms both the other proposed metrics and the state-of-the-art except in the 5-shot with SetFeat12. In particular, SetFeat4-64 (sum-min) results in an accuracy gain of 1.83\% and 1.4\% over MELR\cite{fei2020melr} in 1- and 5-shot, respectively. 


\begin{table}[t]
\renewcommand{\tabcolsep}{2pt}
\centering
\caption{TieredImageNet  evaluation. Bold/red is best/second best, and $\pm$ indicates the 95\% conf. intervals over 600 episodes of 5-way.} 
\begin{tabular}{llcccc}  
        \toprule  
        & \textbf{Method}   
        & \textbf{Backbone} 
        & \textbf{1-shot}  
        & \textbf{5-shot}    
        \\  
        \midrule     
                & OptNet~\cite{lee2019meta}      
                    & \multirow{13}{*}{\rotatebox{90}{---------------- ResNet12 ---------------}}    
                    & 65.99\scriptsize{$\pm$0.72}               & 81.56\scriptsize{$\pm$0.53} 
                    \\
                & MTL~\cite{sun2019meta}       
                    & 
                    & 65.62\scriptsize{$\pm$1.80}           & 80.61\scriptsize{$\pm$0.90} 
                    \\  
                & DNS~\cite{Simon_2020_CVPR}  
                    & 
                    & 66.22\scriptsize{$\pm$0.75}           & 82.79\scriptsize{$\pm$0.48}   
                    \\  
                & Simple~\cite{Tian_2020_ECCV_good}  
                    & 
                    &  69.74\scriptsize{$\pm$0.72}           & 84.41\scriptsize{$\pm$0.55} 
                    \\ 
                    
                & TapNet~\cite{yoon2019tapnet}    
                    & 
                    & 63.08\scriptsize{$\pm$0.15}            & 80.26\scriptsize{$\pm$0.12}  
                    \\ 
                     
                & ProtoNet$^{\dag}$~\cite{snell2017prototypical}       
                    & 
                    & 68.23\scriptsize{$\pm$ 0.23}      & 84.03\scriptsize{$\pm$0.16}  
                \\
                & FEAT~\cite{Ye_2020_CVPR}  
                    & 
                   & 70.80\scriptsize{$\pm$0.23}       & 84.79\scriptsize{$\pm$0.16} 
                \\
 
                & MixtFSL~\cite{afrasiyabi2021MixtFSL}
                    & 
                    & 70.97\scriptsize{$\pm$1.03}       & 86.16\scriptsize{$\pm$0.67}      
                    \\ 
                       
                & Distill~\cite{Tian_2020_ECCV_good}         
                    & 
                    & 71.52\scriptsize{$\pm$0.69}       & 86.03\scriptsize{$\pm$0.49} 
                    \\     
                & DeepEMD~\cite{Zhang_2020_CVPR} 
                    & 
                    & 71.16\scriptsize{$\pm$0.87}       & 86.03\scriptsize{$\pm$0.58} 
                   \\
         
                & DMF~\cite{xu2021learning}   
                    & 
                    & 71.89\scriptsize{$\pm$0.52}       & 85.96\scriptsize{$\pm$0.35} 
                    \\

                & MELR~\cite{fei2020melr}   
                    & 
                    & {72.14}\scriptsize{$\pm$0.51}       & {87.01}\scriptsize{$\pm$0.35} \ \\

                & Distill~\cite{rizve2021exploring}   
                    & 
                    & {\color{red}72.21}\scriptsize{$\pm$0.90}       & {\color{red} 87.08}\scriptsize{$\pm$0.58} 
                    
                \\*[0.5em]
                
                \multirow{3}{*}{\rotatebox{90}{ {-- Ours --}} }  
                 & Match-sum  
                    &   \multirow{3}{*}{\rotatebox{90}{- SF12 -\hspace*{0.25em}}} 
                    & 71.22\scriptsize{$\pm$0.86}      & 85.43\scriptsize{$\pm$0.55}   \\ 
                    
                &  Min-min
                        & 
                        & 71.75\scriptsize{$\pm$0.90}      & 86.40\scriptsize{$\pm$0.56}    \\   
                &  Sum-min 
                        & 
                        & \textbf{73.63}\scriptsize{$\pm$0.88}      & \textbf{87.59}\scriptsize{$\pm$0.57}    \\*[0.2em]

            \bottomrule       
    \end{tabular}  \\
    {\footnotesize $^{\dag}$taken from \cite{lee2019meta}}; {\footnotesize Mappers dimension: SF12 $\in\mathbb{R}^{512}$}
    \label{tab:tieredImageNet}
\end{table}

\paragraph{tieredImageNet} Table~\ref{tab:tieredImageNet} presents the tieredImageNet evaluation of SetFeat12 with our proposed metrics.  
Our sum-min metric results in 1.42\% and 0.51~\% improvement over the baseline Distill~\cite{rizve2021exploring} in 1- and 5-shot. Please note that baselines such as Distill~\cite{rizve2021exploring}, MELR~\cite{fei2020melr}, and FEAT~\cite{Ye_2020_CVPR} contain more parameters than the original ResNet12 and SetFeat12.   
 
\begin{table}[t]
\renewcommand{\tabcolsep}{2pt}
\centering
\caption{
Fine-grained evaluation using CUB in 5-way. 
$\pm$ is the 95\% confidence intervals on 600 episodes( $^\ddag$taken from \cite{tang2020long}). 
}. 
\begin{tabular}{rlcccc}  
        \toprule   
        &  \textbf{Method} & \textbf{Backbone} &\textbf{1-shot} &\textbf{5-shot}  
        \\ 
        \midrule          
             &  MatchingNet\cite{vinyals2016matching}  
                & \multirow{6}{*}{\rotatebox{90}{--- Conv4-{64} ---}}  
                & 61.16\scriptsize{$\pm$0.89}        &72.86\scriptsize{$\pm$0.70}  \\ 
                
            &  ProtoNet\cite{snell2017prototypical}  
                & 
                & 64.42\scriptsize{$\pm$0.48}        & 81.82\scriptsize{$\pm$0.35}   \\  
                
            &  MAML\cite{finn2017model}  
                & 
                & 55.92\scriptsize{$\pm$0.95}        & 72.09\scriptsize{$\pm$0.76}   \\  
                
            &  RelationNet\cite{sung2018learning}  
                & 
                & 62.45\scriptsize{$\pm$0.98}        & 76.11\scriptsize{$\pm$0.69}   \\  

            &  FEAT\cite{Ye_2020_CVPR}  
                & 
                & 68.87\scriptsize{$\pm$0.22}        & 82.90\scriptsize{$\pm$0.15}   \\  
                
             &  MELR\cite{fei2020melr}  
                & \multirow{5}{*}{\rotatebox{90}{SF4-64\hspace*{1.5em}}} 
                & {\color{red}70.26}\scriptsize{$\pm$0.50}        &  {\color{red}85.01}\scriptsize{$\pm$0.32}   \\*[0.5em]

            \multirow{3}{*}{\rotatebox{90}{-- Ours  --}} 
            &  Match-sum  
                & 
                & 67.35\scriptsize{$\pm$0.93}        & 83.82\scriptsize{$\pm$0.61}   \\ 
            &    Min-min 
                &  
                & 70.15\scriptsize{$\pm$0.93}        & 84.94\scriptsize{$\pm$0.64} \\ 

            &  Sum-min 
               & 
                & \textbf{72.09}\scriptsize{$\pm$0.92}        & \textbf{87.05}\scriptsize{$\pm$0.58}    
                \\*[0.2em]

            \bottomrule
                & Robust-20~\cite{dvornik2019diversity}   
                &  \multirow{7}{*}{ \rotatebox{90}{ ----- ResNet18 ------ }}
                & 58.67\scriptsize{$\pm$0.7}     & 75.62\scriptsize{$\pm$0.5}  
                    \\      
                      
                & RelationNet$^\ddag$~\cite{sung2018learning}      
                    & 
                    & 67.59\scriptsize{$\pm$1.0}             & 82.75\scriptsize{$\pm$0.6} 
                     \\
                     
                & MAML$^\ddag$~\cite{finn2017model}    
                    & 
                    & 68.42\scriptsize{$\pm$1.0}           & 83.47\scriptsize{$\pm$0.6} 
                    \\

                & ProtoNet$_{\ddag}$~\cite{snell2017prototypical}    
                    & 
                    & 71.88\scriptsize{$\pm$0.9}             & 86.64\scriptsize{$\pm$0.5} 
                    \\
                    
                & Baseline++~\cite{chen2019closer} 
                    & 
                    & 67.02\scriptsize{$\pm$0.9}      &  83.58\scriptsize{$\pm$0.5}
                    \\  
                    
                 & MixtFSL~\cite{afrasiyabi2021MixtFSL}  
                    & 
                    & {\color{red}73.94}\scriptsize{$\pm$1.1}      & 86.01\scriptsize{$\pm$0.5} 
                    \\ 
                & Neg-Margin~\cite{Bin_2020_ECCV_margin_matter}   
                    & \multirow{3}{*}{\rotatebox{90}{- SF12$^*$-\hspace*{1.75em}}}
                    & 72.66\scriptsize{$\pm$0.9}           & {\color{red}89.40}\scriptsize{$\pm$0.4}  
                    \\*[0.5em]

            \multirow{3}{*}{\rotatebox{90}{-- Ours  --}}
            & Match-sum 
                & 
                & 77.95\scriptsize{$\pm$ 0.83}      & {88.93}\scriptsize{$\pm$ 0.49}   
                \ \\ 
                
            &  Min-min 
                & 
                & 78.51\scriptsize{$\pm$0.82}        & 89.73\scriptsize{$\pm$0.47}   
                \\ 
                   
            &  Sum-min 
                & 
                & \textbf{79.60}\scriptsize{$\pm$0.80}      & \textbf{90.48}\scriptsize{$\pm$ 0.44}  
               \\*[0.2em]

                   

                
                   

        \bottomrule       
    \end{tabular}  \\
    {\footnotesize Mappers dimension: SF4-64 $\in\mathbb{R}^{64}$, and  SF12$^* \in\mathbb{R}^{480}$}
    \label{tab:cub}
\end{table} 
\paragraph{CUB} Table~\ref{tab:cub} illustrates the fine-grained classification evaluation of our approach, compared to Conv4-64 and ResNet18. We observe that SetFeat4-64 (min-min) again surpasses all baselines by providing gains of 1.83\% and 2.04\% over MELR~\cite{fei2020melr} in 1- and 5-shot respectively. When comparing with ResNet18, we further reduce the number of convolution kernels to 128/150/196/480 (dubbed SetFeat12$^*$) to better match the number of parameters (11.466M for SetFeat12$^*$ vs 11.511M for ResNet18). Our approach again defines a new state-of-the-art performance in this scenario.

\section{Ablation}
\label{sec:ablation}

In this section, we further analyze SetFeat to explore alternative design decisions and gain a better understanding as to why our set-based model achieves better accuracy.



\subsection{Mapper configurations}
\label{sec:mapperconfiguration}

We now experiment with different ways of embedding ten mappers throughout the backbone levels. We compare: 1) putting all mappers on the last layer (0-0-0-10); 2) a single mapper per block (1-1-1-1); 3) distributing mappers more equally (2-2-3-3); and 4) employing a progressive growth strategy (1-2-3-4) (this last one being used in the main evaluation in \cref{sec:evaluation}). Table~\ref{tab:mappers_config} compares these four strategies on both SetFeat4-64 and SetFeat4-512 on the validation set of miniImageNet. We observe that placing mappers throughout the network yields better results than putting them all at the end. The two other options perform similarly. We also observe that (2-2-3-3) only beats (1-2-3-4) using shallower network SetFeat4-64 in 5-shot. Otherwise, progressive growth either reaches or surpasses the other combinations. 
Note, that going from 0-0-0-10 to 1-2-3-4 or 2-2-3-3 improves performance while using the same number of mappers, which confirms that multi-scale indeed helps. Additionally, removing our set-based representation by concatenating all mappers outputs and treating the result as a single (multi-scale) feature vector (``concat'' in tab.~\ref{tab:mappers_config}) completely cancels any performance gain. Therefore, we conclude that it is our sets of multi-scale features that explains the performance improvement.
\begin{table}[t]
\renewcommand{\tabcolsep}{2pt}
\centering
\caption{
Ablation of different  mapper-level combinations using miniImageNet. The results are validation accuracy with min-sum. 
} 
\begin{tabular}{rlcccccccc}  
        \toprule   
        & &  & & \multicolumn {2}{c}{\textbf{SetFeat4-64}} & & & \multicolumn {2}{c}{\textbf{SetFeat4-512}}  \\
        & \textbf{Mappers} & &  &{1-shot} &{5-shot}  & &   &{1-shot}  &{5-shot}  \\ 
        \midrule    
            & ProtoNet$^*$   & &       & 53.51   & 71.57  &  &  & --  & --    \\ 
            & 0-0-0-1        & &       & 53.55   & 71.51  &  &  & --  & --    \\ 
            & 1-2-3-4 (concat)       & &           & 53.56   & 71.82   &  & & --  & --  \\
            & 1-1-1-1             & &          & 51.11   & 69.41    &  &    & 53.57  & 71.60     \\ 
            & 0-0-0-10            & &          & 52.90   & 69.49    &  &    & 55.36  & 71.59     \\ 
            & 2-2-3-3             & &          & 54.73   & 71.98    &  &    & 56.29  & 74.74      \\ 
            & 1-2-3-4             & &          & 54.71   & 71.35    &  &    & 58.74  & 75.30     \\   
        \bottomrule       
    \end{tabular}  \label{tab:mappers_config} \ \\
    \small{$^*$ with Conv4-512} 
\end{table}





             

\subsection{Over-parameterization of SetFeat4-64}
\label{SetFeat4_overparam_ablation} 
\Cref{sec:backbones} mentioned that the number of kernels in backbone feature extractors was reduced in such a way that adding our proposed attention-based mappers did not significantly change the total number of parameters in the network---but unfortunately doing so for Conv4-64 resulted in poor generalization as each of its four blocks is only composed of a single layer with 64 kernels.
Here, we instead \emph{augment} Conv4-64 and add parameters with three FC layers (of 512, 160, 64 dimensions) after the convolutional blocks. This reaches 0.239M parameters, which matches the 0.238M parameters of SetFeat4-64. Results are presented in table~\ref{tab:AugmConv4-64}. Although the augmented Conv4-64 improves over the baseline Conv4-64, the improvements are significantly below those obtained by SetFeat4-64, showing that the additional parameters alone do not explain the performance gap.

\begin{table}[t]
\renewcommand{\tabcolsep}{2pt}
\centering
\caption{
Ablation of our SetFeat with miniImageNet and CUB on 600 episodes with augmented Conv4-64 and SetFeat4-64 in 5-way. 
}
\label{tab:AugmConv4-64}
\begin{tabular}{rlccccc}  
        \toprule   
        &  & \multicolumn {2}{c}{\textbf{miniImageNet}}    
        & \multicolumn {2}{c}{\textbf{CUB}} \\
        & \textbf{Method} &{1-shot} &{5-shot}  &{1-shot}  &{5-shot}  \\ 
        \midrule    
            & ProtoNet~\cite{snell2017prototypical}
              &49.42   &68.20  & 68.23   & 84.03    \\ 
            & ProtoNet$^*$~\cite{snell2017prototypical}
            &49.98   &69.53  & 69.11   & 85.27   \\ 
            & Sum-min (ours) 
             &57.18   &73.67  &73.50   &87.61 
            \\ 
        \bottomrule       
    \end{tabular}  \label{tab:AugConv4} \ \\
    {$^*$ our implementation with augmented Conv4-64} 
\end{table}
\subsection{Probing the activation of mappers}
\label{sec:ablation-mappers}

\begin{figure} 
    \centering
    \footnotesize
    \includegraphics[width=1\linewidth]{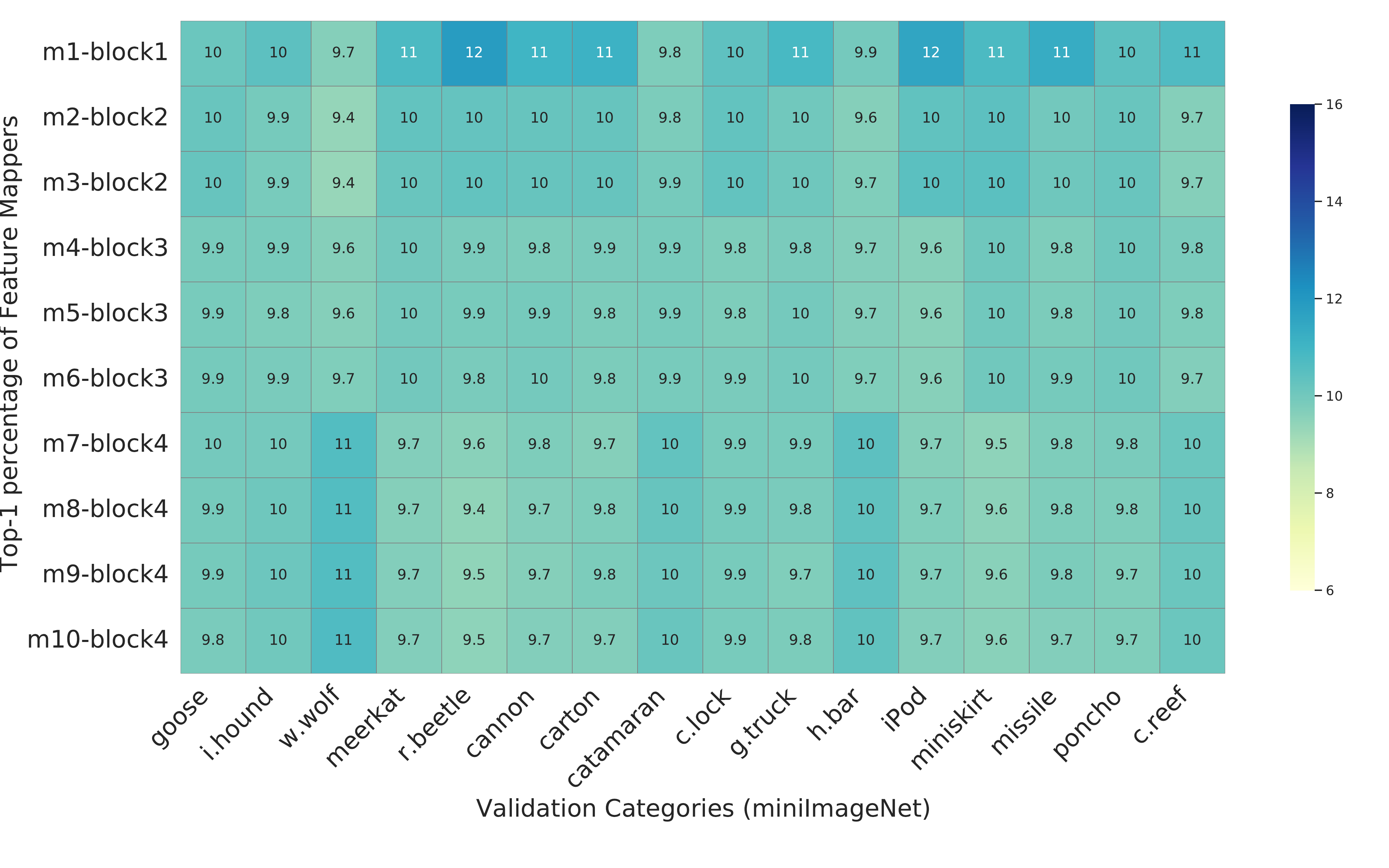}    
    \caption{The percentage time each of the mappers ($y$-axis) is selected for each of the 16 validation categories ($x$-axis) of the miniImageNet dataset. The result is obtained by SetFeat12 and averaged over 600 episodes of 5-way 1-shot. While the earlier mappers are more often active, all mappers are consistently useful. }
    \label{fig:mapper_frequency}
\end{figure}

Let us now investigate whether all mappers are actually useful by analyzing the behavior under the sum-min metric (\cref{sec:set2set}). For this,  fig.~\cref{fig:mapper_frequency} illustrates the percentage of time where a specific mapper ($y$-axis) provides the minimum prototype-query distance for each validation class ($x$-axis) in the miniImageNet dataset. This illustrates that low-level mappers are often active like the high-level ones, but all mappers are consistently being used across all validation classes, thereby validating that our proposed set-based representation is effective and working as expected.

In addition, \cref{fig:tsne_mappers} shows t-SNE~\cite{van2008visualizing} visualizations of 640 embedded examples from miniImageNet, CUB, and tieredImageNet datasets using our set-feature extractor. Note how the distributions of mapper embeddings are generally disjoint and do not collapse to overlapping points, which shows intuitively that mappers extract different features. 

\begin{figure} 
    \centering
    \footnotesize
    \setlength{\tabcolsep}{1pt}
    \begin{tabular}{cccc} 
    \includegraphics[width=0.33\linewidth]{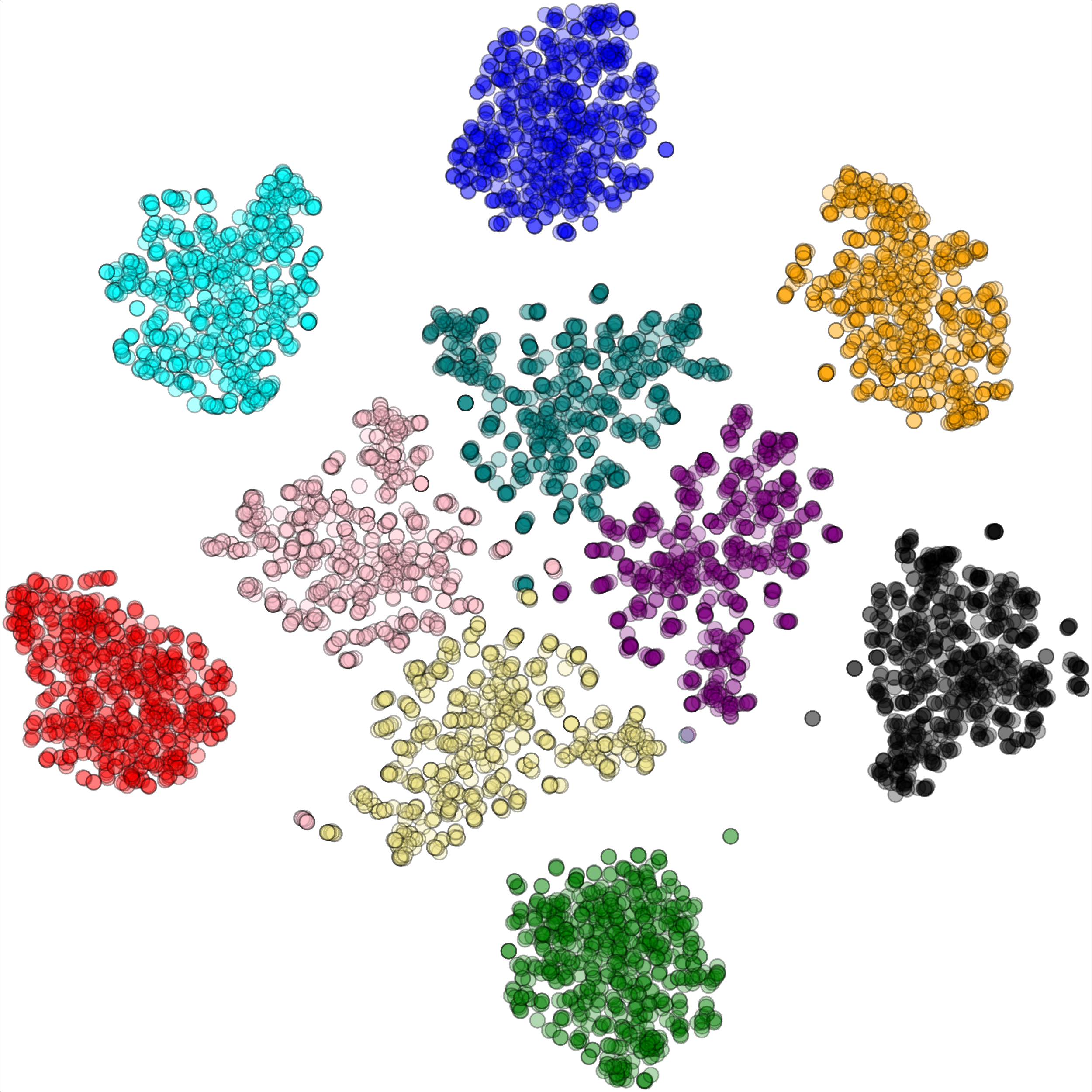} & 
    \includegraphics[width=0.33\linewidth]{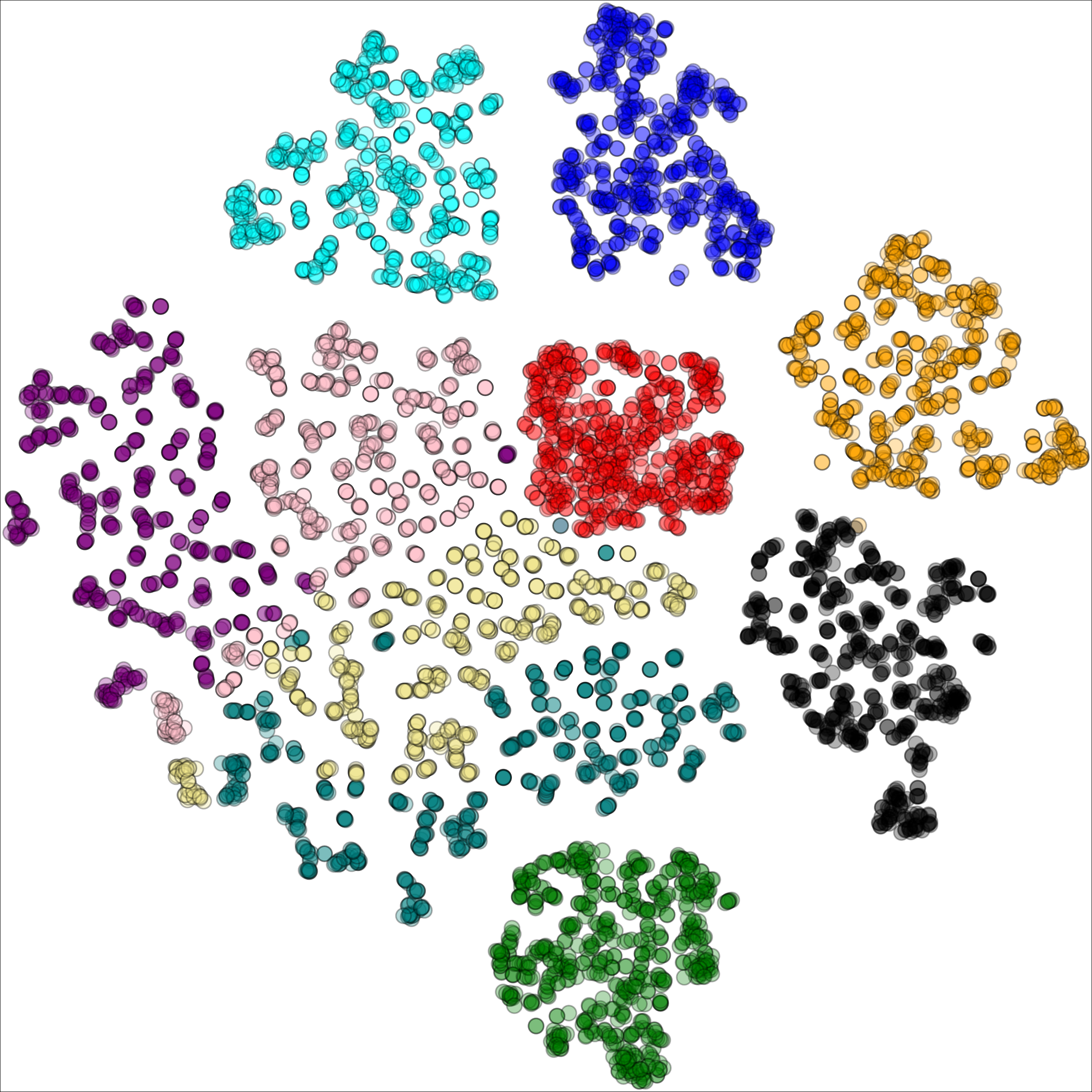} &   
    \includegraphics[width=0.33\linewidth]{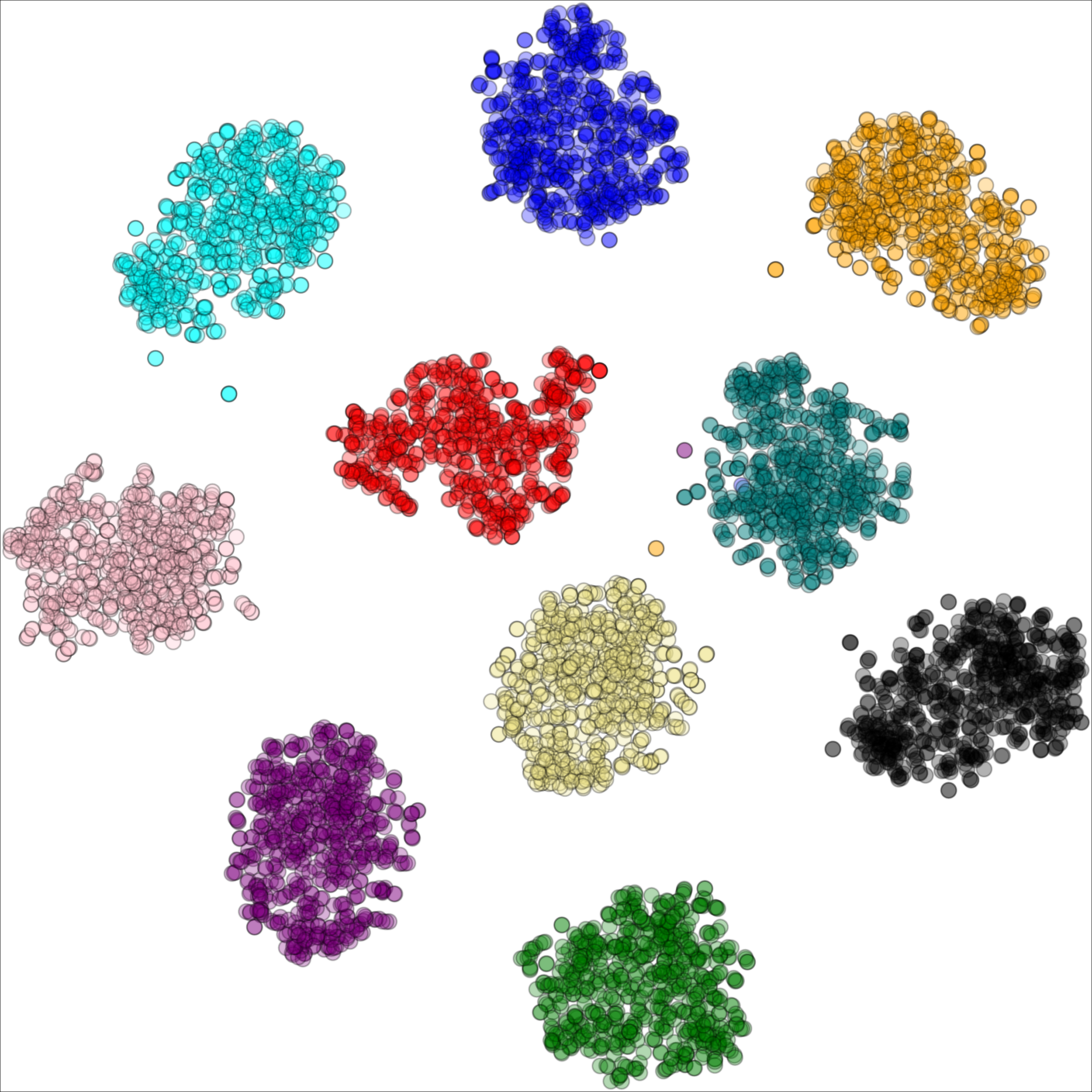} &   \\
    (a) miniImageNet  & 
    (b) CUB  & (c) tieredImageNet   \\       
    \end{tabular} 
    \caption{Visualizing mappers with t-SNE~\cite{van2008visualizing} on  640 randomly-sampled from validation set for (a) miniImageNet with SetFeat12, (b) CUB with SetFeat12$^*$ (\cref{sec:quant-evaluation}) and (c) tieredImageNet with SetFeat12. Points are color-coded according to the mapper.}  
    \label{fig:tsne_mappers}
\end{figure}

\subsection{Top-$m$ analysis}

The min-min and sum-min metrics (\cref{eq:min-min,eq:sum-min} respectively) are two ends of the spectrum: min-min takes the minimum distance across all mappers, while sum-min computes the sum over all the mappers. Here, we sort the mappers according to distance and sum the top-$m$ as an ablation shown in table \ref{tab:topK}. In general, we observe that the classification results progressively improve as we move towards sum-min, which uses all of the mappers.

\subsection{Visualizing mappers saliency}
\label{sec:viz_mapper}
We now visualize in fig.~\cref{fig:saliency_maps} the impact of learning a set of features by visualizing the saliency map of each mapper, and by comparing them with the saliency maps of the single-feature approach of Chen~\etal~\cite{chen2019closer}. We compute the smoothed saliency maps~\cite{simonyan2014deep} by single back-propagation through a classification layer. It can be seen that our approach devotes attention to many more parts of the images than when a single feature vector is learned. For example, note how a single dog is highlighted (fourth row of  fig.~\cref{fig:saliency_maps}), whereas our mappers jointly fire on all three. Please consult the supplementary materials for more examples.

\begin{table}[t]
\renewcommand{\tabcolsep}{5pt}
\centering
\caption{
Ablation of top-$m$ mapper in the min-sum metric using SetFeat4 and SetFeat12$^*$ on CUB. The results are validation set.
} 
\begin{tabular}{rlccccc}  
        \toprule   
        &  & \multicolumn {2}{c}{\textbf{SetFeat4}}    
        & \multicolumn {2}{c}{\textbf{SetFeat12$^*$}} \\
        & \textbf{Method} &{1-shot} &{5-shot}  &{1-shot}  &{5-shot}  \\ 
        \midrule    
            & top-1 (min-min)    &70.15   &84.94       &78.51    & 89.73    \\ 
            & top-2              &70.84   &85.30        &77.92    & 89.87   \\ 
            & top-4              &70.34    &85.95        &78.37    & 89.78   \\ 
            & top-8              &71.47    &86.88        &79.56    & 90.03   \\   
            & top-10 (sum-min)   & \textbf{72.09}    & \textbf{87.05}       &\textbf{79.60}    & \textbf{90.48}  \\ 
        \bottomrule       
    \end{tabular}  \label{tab:topK}  
\end{table}


\begin{figure} 
    \centering
    \footnotesize
    \setlength{\tabcolsep}{1pt}
    \begin{tabular}{cccccccccc}

     \includegraphics[width=0.135\linewidth, angle=0]{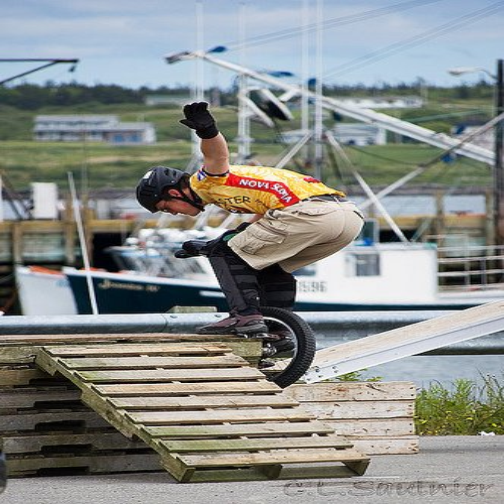} & 
    \includegraphics[width=0.135\linewidth, angle=0]{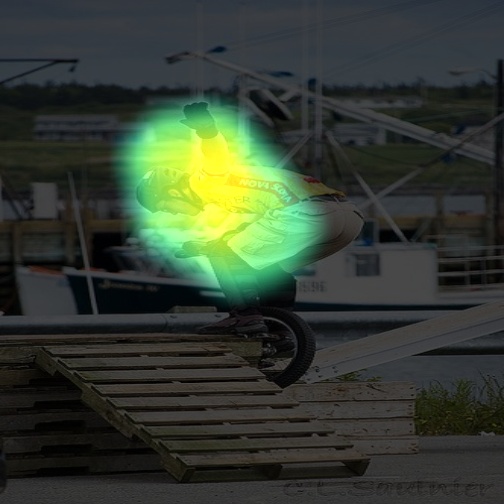} & 
    \includegraphics[width=0.135\linewidth, angle=0]{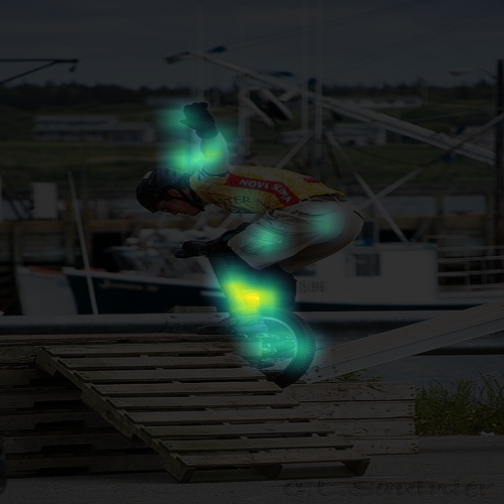} &
    \includegraphics[width=0.135\linewidth, angle=0]{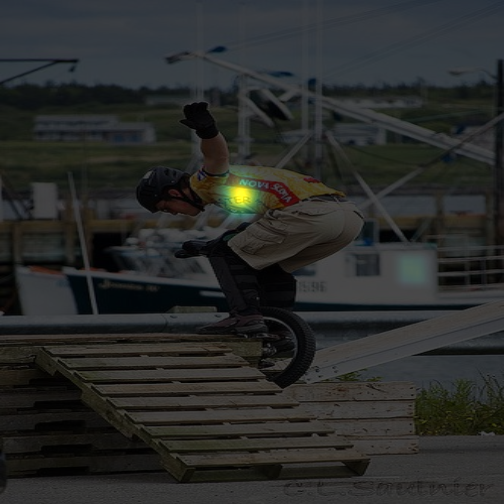} &  
    \includegraphics[width=0.135\linewidth, angle=0]{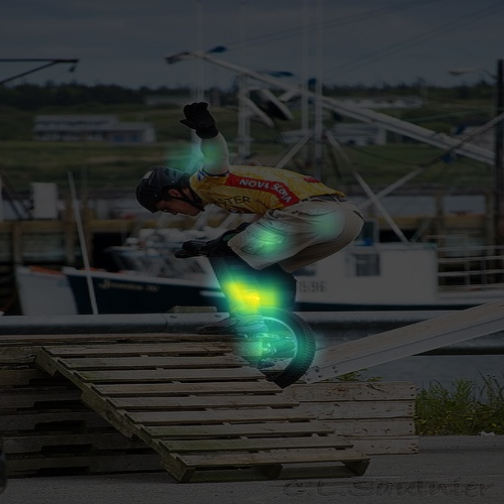} & 
    \includegraphics[width=0.135\linewidth, angle=0]{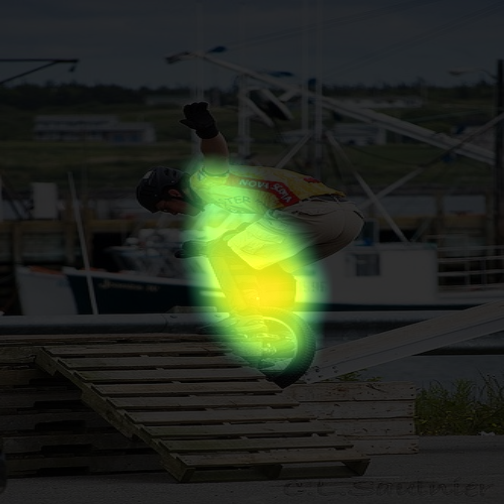} & 
    \includegraphics[width=0.135\linewidth, angle=0]{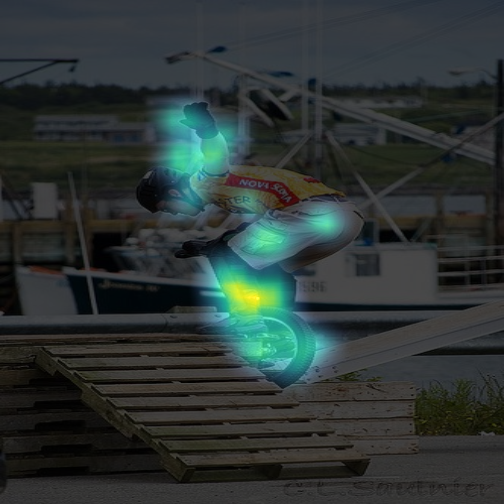} 
    \\
    
     \includegraphics[width=0.135\linewidth, angle=0]{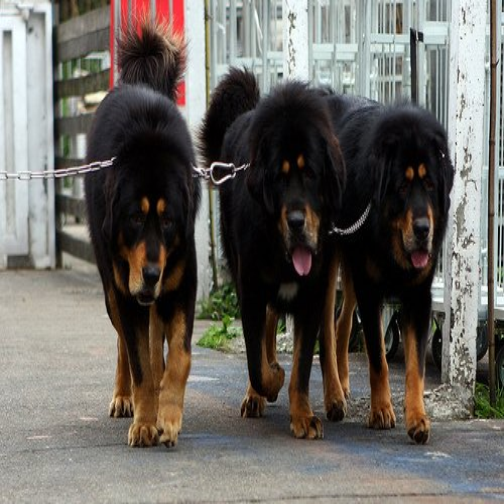} & 
    \includegraphics[width=0.135\linewidth, angle=0]{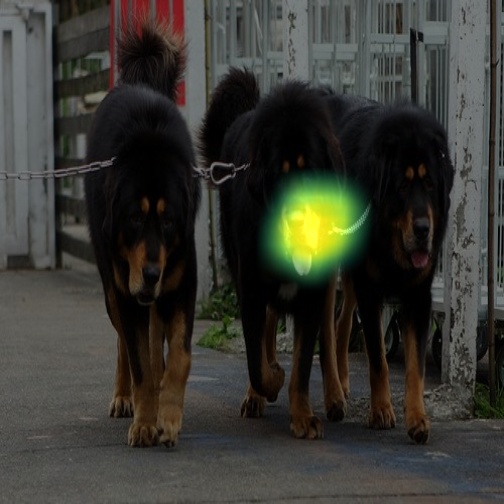} & 
    \includegraphics[width=0.135\linewidth, angle=0]{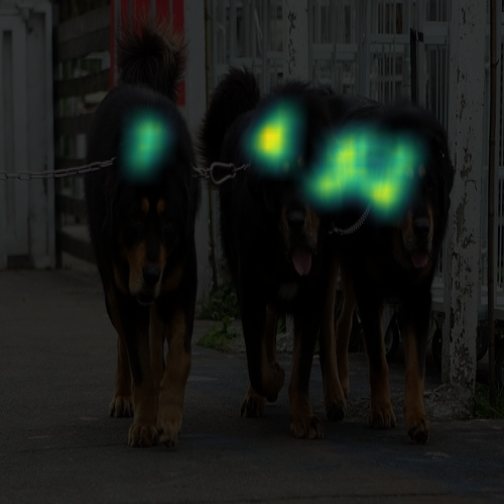} &
    \includegraphics[width=0.135\linewidth, angle=0]{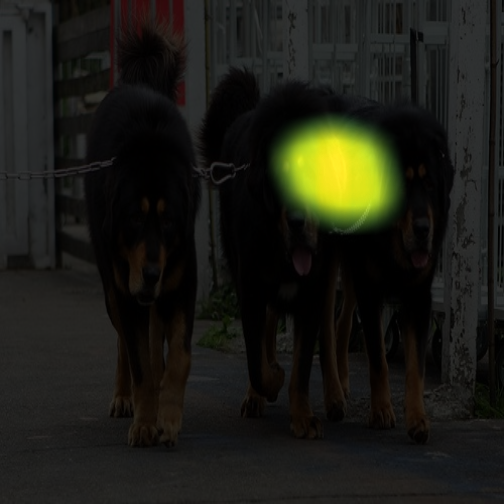} &  
    \includegraphics[width=0.135\linewidth, angle=0]{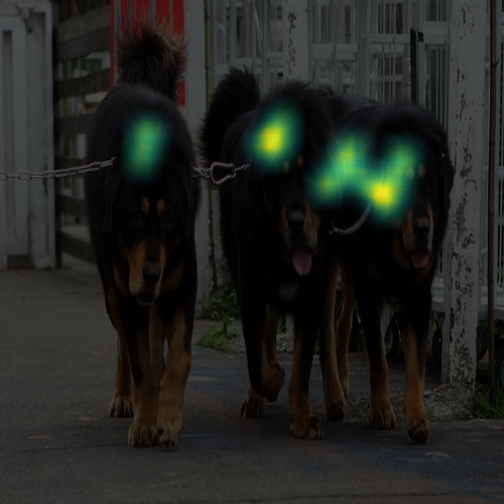} & 
    \includegraphics[width=0.135\linewidth, angle=0]{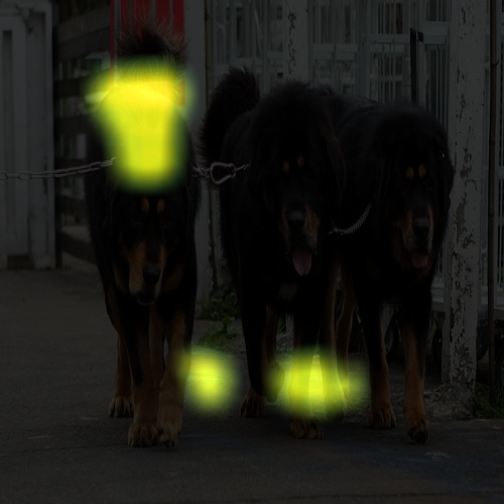} & 
    \includegraphics[width=0.135\linewidth, angle=0]{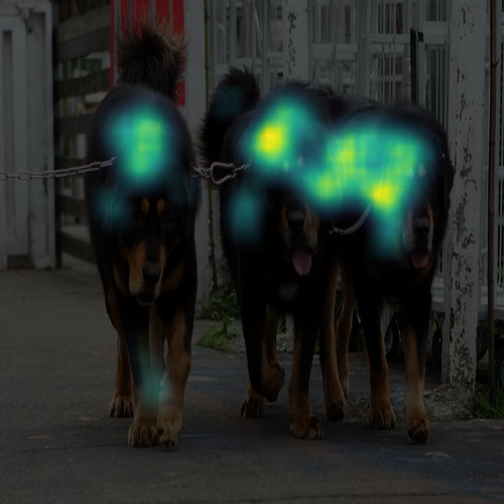} 
    \\
    
     \includegraphics[width=0.135\linewidth, angle=0]{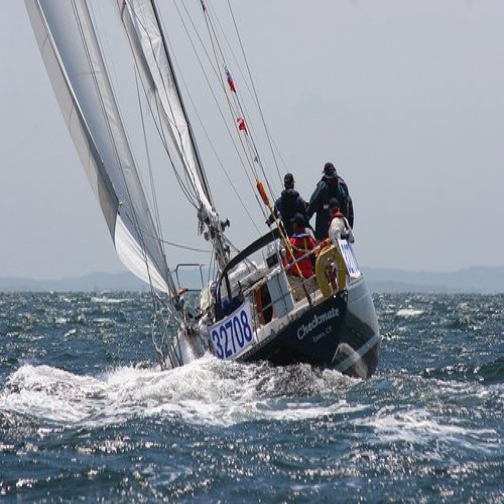} & 
    \includegraphics[width=0.135\linewidth, angle=0]{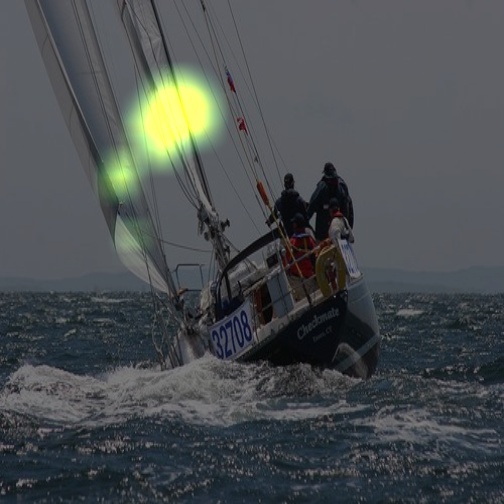} & 
    \includegraphics[width=0.135\linewidth, angle=0]{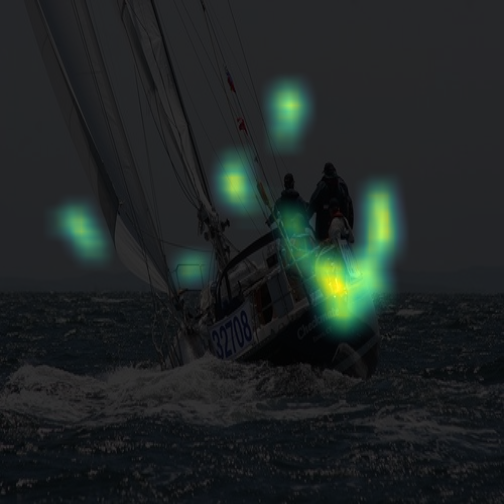} &
    \includegraphics[width=0.135\linewidth, angle=0]{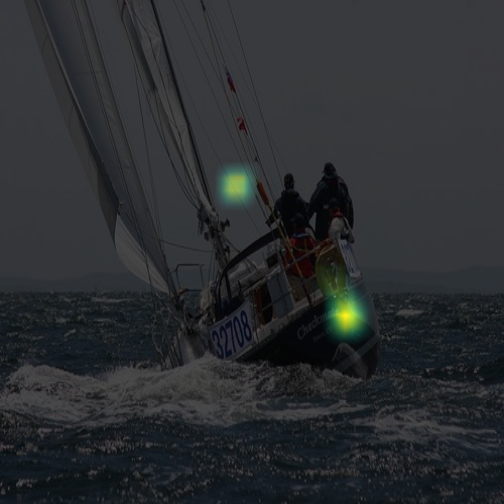} &  
    \includegraphics[width=0.135\linewidth, angle=0]{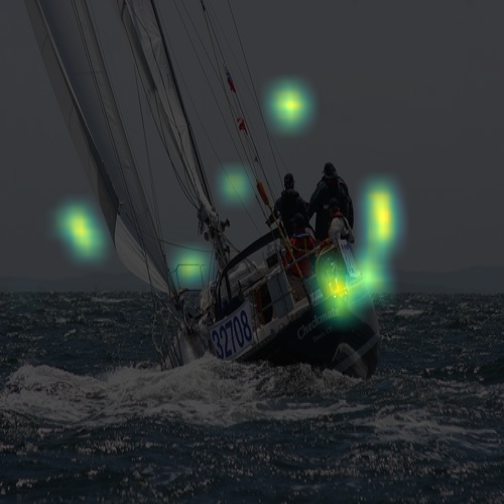} & 
    \includegraphics[width=0.135\linewidth, angle=0]{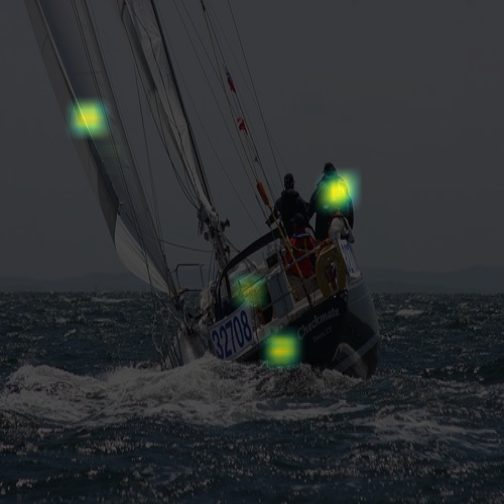} & 
    \includegraphics[width=0.135\linewidth, angle=0]{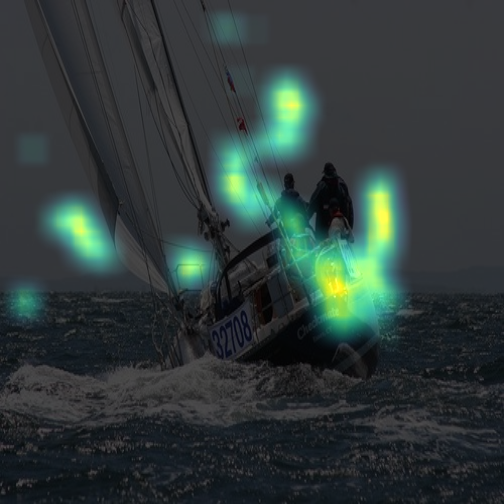} 
    \\
    \\

     \includegraphics[width=0.135\linewidth, angle=0]{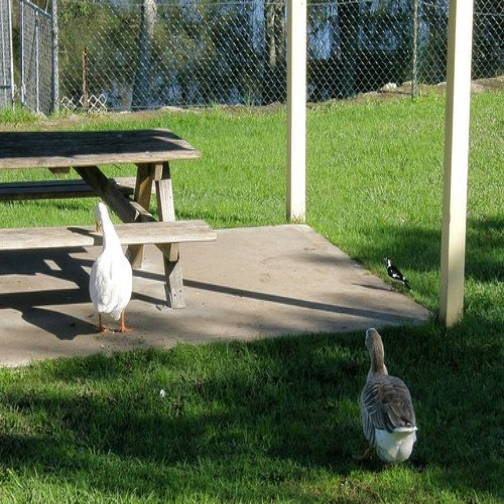} & 
    \includegraphics[width=0.135\linewidth, angle=0]{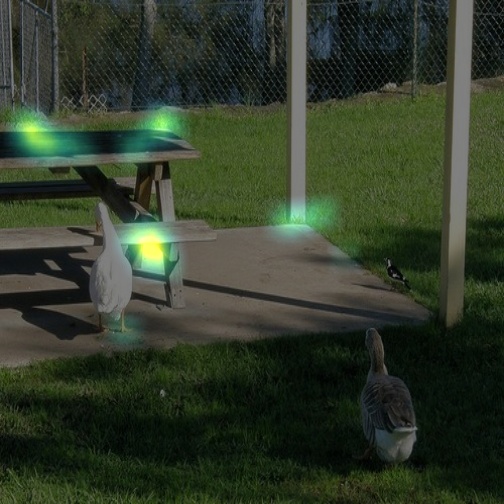} & 
    \includegraphics[width=0.135\linewidth, angle=0]{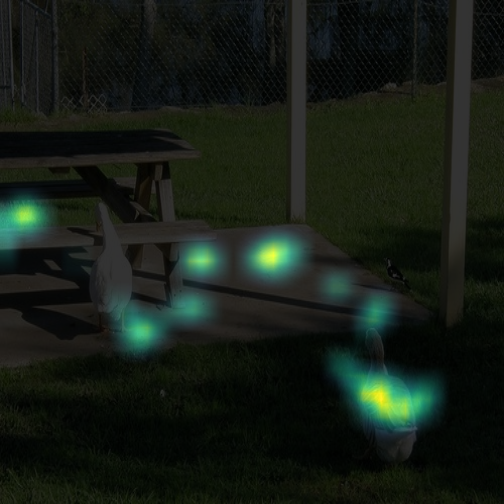} &
    \includegraphics[width=0.135\linewidth, angle=0]{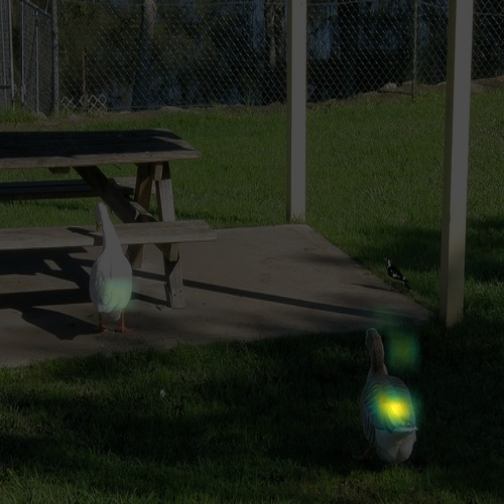} &  
    \includegraphics[width=0.135\linewidth, angle=0]{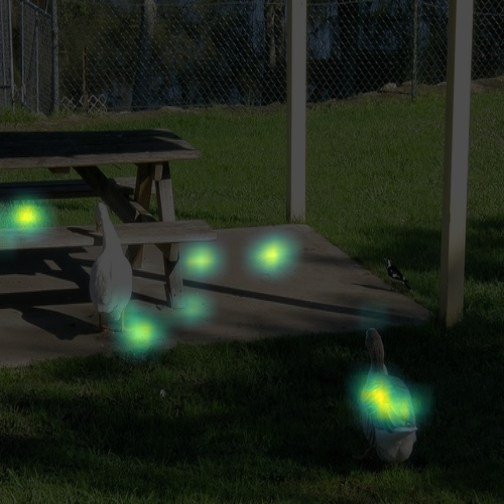} & 
    \includegraphics[width=0.135\linewidth, angle=0]{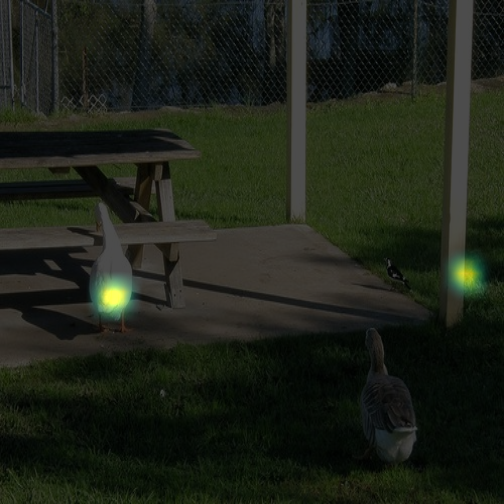} & 
    \includegraphics[width=0.135\linewidth, angle=0]{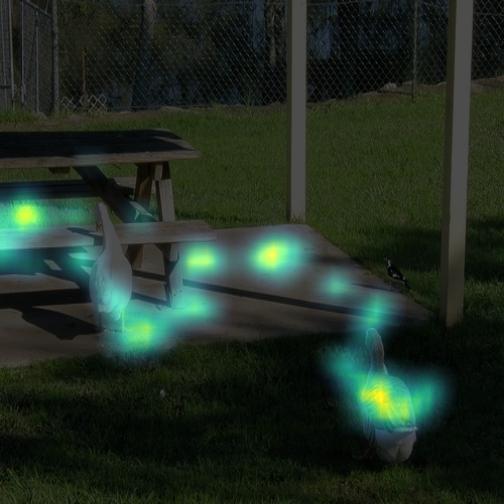} 
    \\
    
     \includegraphics[width=0.135\linewidth, angle=0]{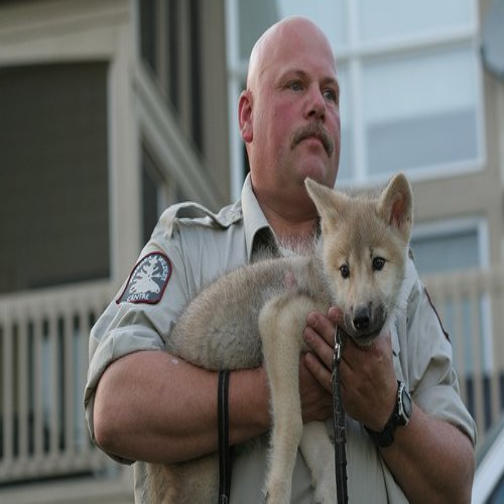} & 
    \includegraphics[width=0.135\linewidth, angle=0]{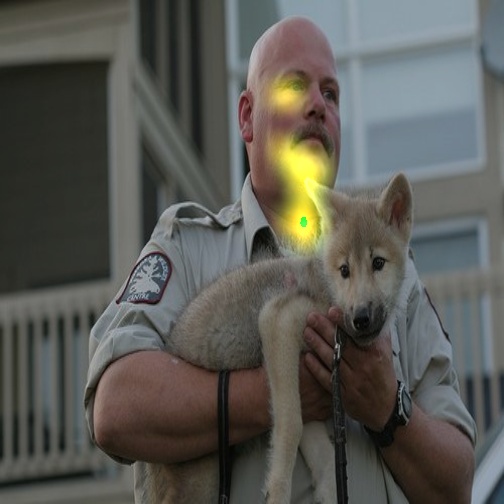} & 
    \includegraphics[width=0.135\linewidth, angle=0]{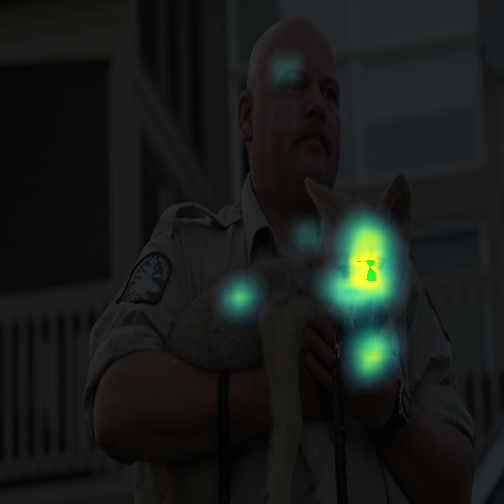} &
    \includegraphics[width=0.135\linewidth, angle=0]{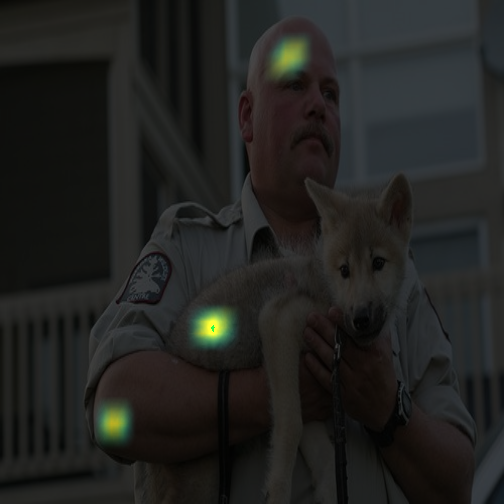} &  
    \includegraphics[width=0.135\linewidth, angle=0]{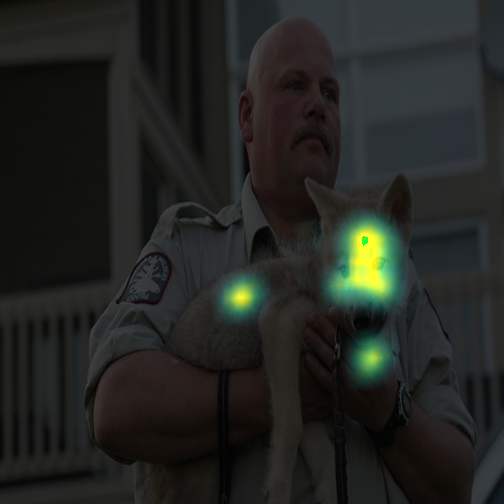} & 
    \includegraphics[width=0.135\linewidth, angle=0]{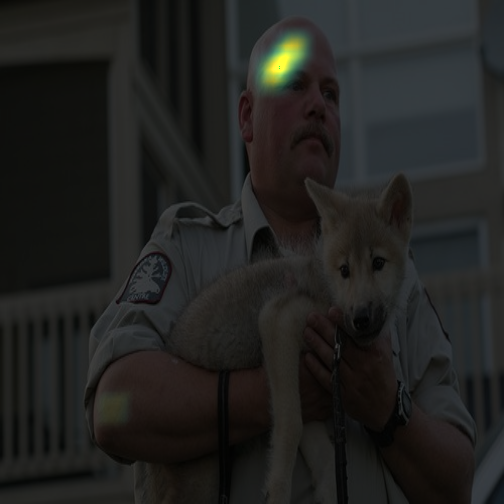} & 
    \includegraphics[width=0.135\linewidth, angle=0]{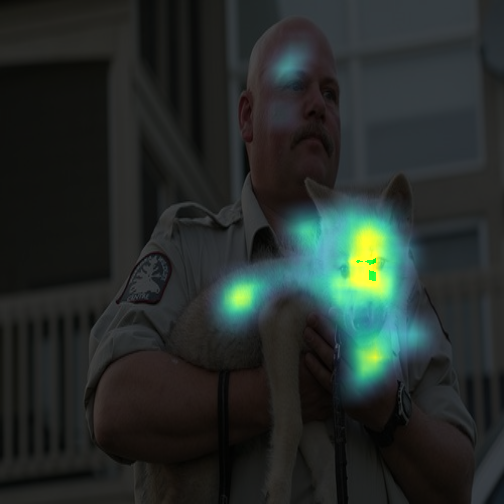} 
    \\
    
    \includegraphics[width=0.135\linewidth, angle=0]{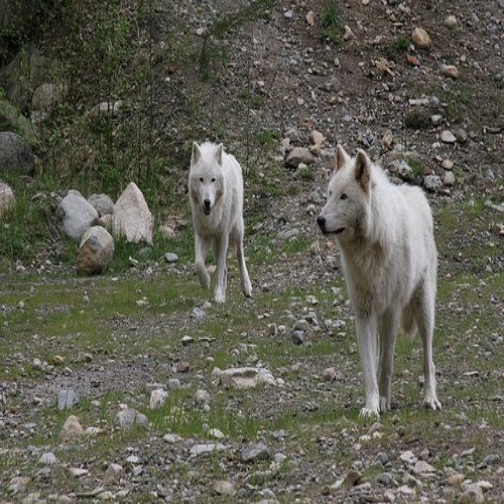} & 
    \includegraphics[width=0.135\linewidth, angle=0]{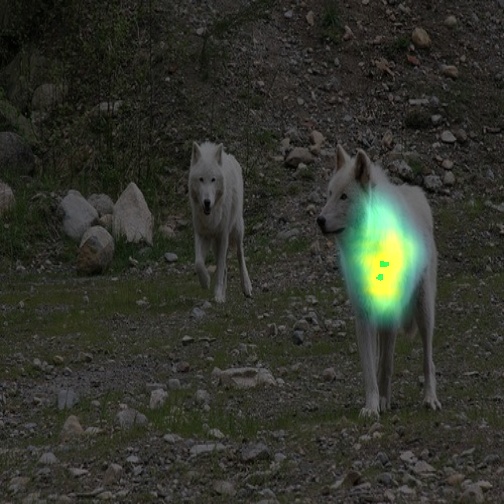} & 
    \includegraphics[width=0.135\linewidth, angle=0]{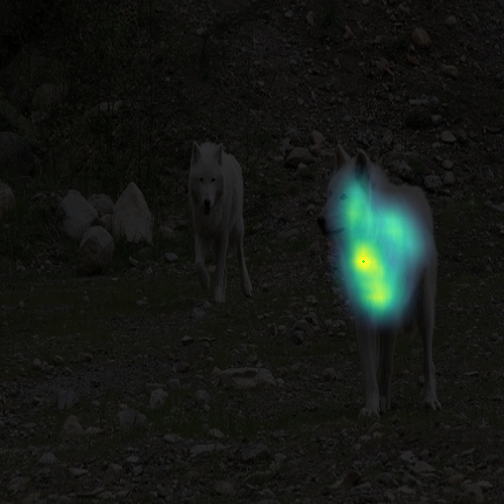} &
    \includegraphics[width=0.135\linewidth, angle=0]{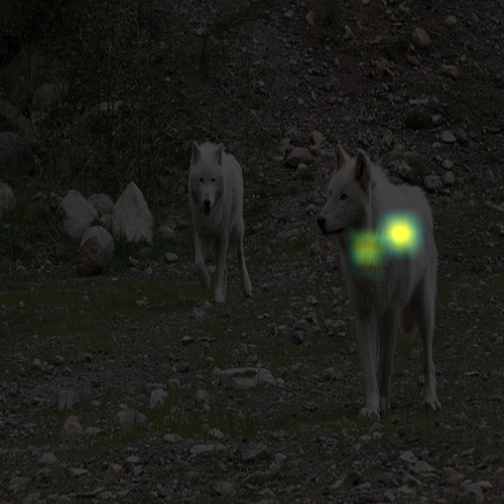} &  
    \includegraphics[width=0.135\linewidth, angle=0]{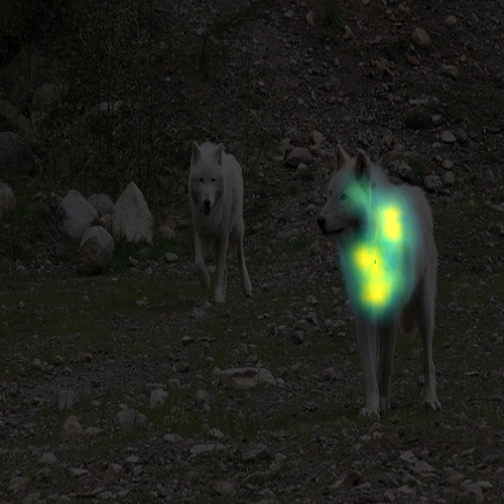} & 
    \includegraphics[width=0.135\linewidth, angle=0]{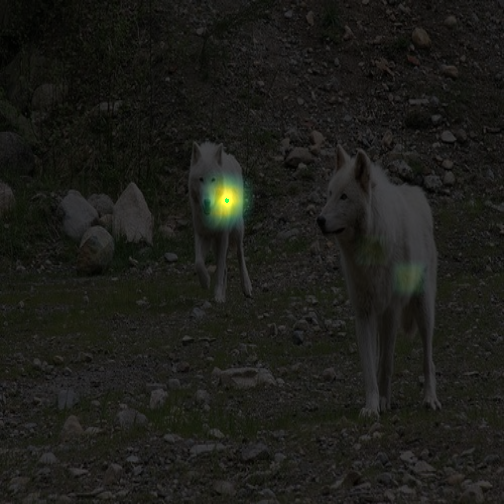} & 
    \includegraphics[width=0.135\linewidth, angle=0]{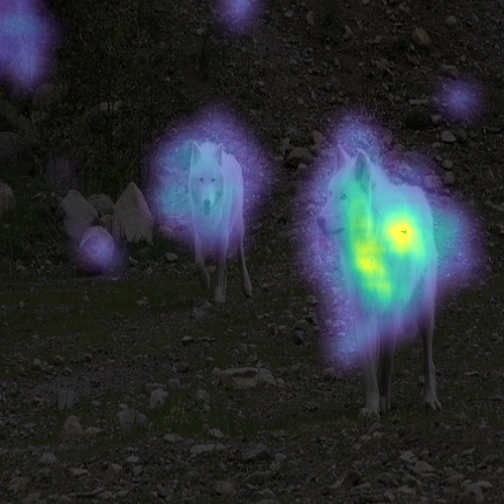} 
    \\   
    (a) input & 
    (b) baseline &
    \multicolumn{5}{c}{(c) ours}
    \end{tabular} 
    \caption{Comparision of gradient saliency maps. 
    From left, we look at the (a) input original image, (b) baseline~\cite{chen2019closer}, and (c) subset of five feature vectors extracted by SetFeat12.  
    The figure presents three examples of the training data in the first rows and four examples from the valid. set of miniImageNet in the last four rows.  
    }
    \label{fig:saliency_maps} 
\end{figure}

\section{Discussion}

This paper proposes to extract and match \emph{sets} of feature vectors for few-shot image classification. This contrasts with the use of a monolithic single-vector representation, which is a popular strategy in that context. To produce these sets, we embed shallow attention-based mappers at different stages of conventional convolutional backbones. 
These mappers aim at extracting distinct sets of features with random initialization, capturing different properties of the images seen. 
Here,  the non-linearity in the sum-min and min-min creates diversity: the inner minimum distance causes a non-linearity that forces the selection of a given mapper. Match-sum, our worst metric, only benefits from random initialization.
We then rely on set-to-set matching metrics for inferring the class of a given query from the support set examples, following the usual approach for inference with prototypical networks. 
Experiments with four different adaptations of two main backbones demonstrate the effectiveness of our approach by achieving state-of-the-art results in miniImageNet, tieredImageNet, and CUB datasets. For fair comparison, the parameters of all the adapted backbones are reduced according to the number of parameters added by the mappers.

\paragraph{Limitations} 

Even though a comparison with different mapper configurations has been provided in \cref{sec:mapperconfiguration}, we have evaluated our method using a fixed set of $M=10$ mappers. Using more mappers ($M>10$) has been considered, but was eventually dismissed since increasing the number of mappers would require reducing the number of filters, which in turn could cause underfitting due to under-parameterization. As future work, we see great potential on analyzing the effect of increasing the number of mappers, possibly with larger backbones.
Another topic requiring further investigations would be to vary the weighting of each mapper through more flexible set-to-set matching metrics. Although the min-sum and min-min metric non-linearly match the feature sets (through the $\min$ operation), investigating the weighted sum-min would be an interesting future work. Here, adapting Deep Set~\cite{zaheer2017deep} before computing the min-sum metric would be a potential direction to investigate the weighted set-to-set mapping.    
Finally, we are particularly enthusiastic regarding the adaptation of our approach to self-supervised, since the set of features provide more choices for the comparison of different variations of single images.

\paragraph{Acknowledgements} 
This project was supported by NSERC grant RGPIN-2020-04799, Mitacs, Prompt-Quebec, and Compute Canada. 
We thank A. Schwerdtfeger, A. Tupper, C. Shui, I. Hedhli,
for proofreading the manuscript.



{\small
\bibliographystyle{ieee_fullname}
\bibliography{egbib.bib}
}

\onecolumn
\section*{Supplementary Material}
    
    
\section*{Dataset and backbone specifications}
\label{sec:data_description}
Table~\cref{tab:datasets} present the detailed specification of dataset, and  and Table~\cref{tab:backboneparameters} specifies the number of overall parameters in our set feature SetFeat extractor compared to the popular backbones used in the few-shot image classification literature.

\begin{table}[hbt!]
\centering
\caption{Specifications of miniImageNet, tieredImageNet and CUB.} 
    \begin{tabular}{rccccc}   
       \toprule
        & Dataset    &Number of examples     &Source  &Splits(train/val/test)   & Split Reference      \\ 
        \midrule
        & MiniImageNet   
        & 60,000
        & ImageNet$^\dag$~\cite{russakovsky2015imagenet}
        & 64/16/20
        & Vinyals~\etal~\cite{vinyals2016matching} 
        \\  
        
        & TieredImageNet   
        & 779,165
        & ImageNet$^\dag$~\cite{russakovsky2015imagenet}
        & 351/97/160
        & Ren~\etal~\cite{ren2018meta} 
        \\  
        
        & CUB 
        & 11,788
        & CUB-200-2011$^*$~\cite{wah2011caltech}
        & 100/50/50
        & Chen~\etal~\cite{chen2019closer} 
        \\    
        \bottomrule
    \end{tabular}\label{tab:datasets}
    \\ 
    $^\dag$ \scriptsize\url{https://www.image-net.org/} \qquad
    $^*$ \scriptsize\url{http://www.vision.caltech.edu/visipedia/CUB-200-2011.html}

\end{table}

\begin{table}[hbt!]
\centering
\caption{Number of parameters for various backbones, compared with our SetFeat implementations (in blue). Blocks column illustrates the number of parameters in all the convolution layers. Mappers column shows the number of parameters in 10 employed mappers in SetFeat.}      
    \begin{tabular}{rcccc}   
      \toprule
        & Backbone       & Blocks  & Mappers       & Total \\ 
        \midrule
        & Conv4-64                     &0.113 M               &--              & 0.113 M   \\  
         & {\color{blue}SetFeat4-64}  & 0.113 M               & 0.124 M              & 0.238 M  \\[0.5em]
          & Conv4-512                  &1.591 M               &--              & 1.591 M  \\  
         & {\color{blue}SetFeat4-512}  & 0.587 M              & 0.996 M             & 1.583 M  \\[0.5em]
         
         &ResNet18                     & 11.511 M              &--              & 11.511 M  \\  
         &{\color{blue}SetFeat12$^{*}$}& 6.977 M               & 4.489 M              & 11.466 M  \\[0.5em]
         
         &ResNet12                     & 12.424 M               & --             & 12.424 M  \\  
         &{\color{blue}SetFeat12}      & 7.447 M               & 4.902 M             & 12.349 M  \\[0.5em]  
         
        \bottomrule
    \end{tabular}
    \\ 
    \label{tab:backboneparameters}
\end{table}

    
    



\newpage
\section*{Ablation with more ways and cross-domain results from $\text{miniImageNet} \mapsto \text{CUB}$}
\label{sec:number_of_ways} 

\Cref{tab:More_way} shows 5-way, 10-way, and 20-way comparisonds of SetFeat12$^*$ and SetFeat12 with ResNet18 and ResNet12, respectively. As illustrated in cref{tab:backboneparameters} and mentioned in sec.~5.3 of the main paper, SetFeat12$^*$ (11.466M parameters) is the counterpart of ResNet18 (11.511M parameters). 

\Cref{tab:More_way} shows that SetFeat with the sum-min metric (eq.~(5) from the main paper) achieves state-of-the-art results in 5-shot for all of 5-, 10- and 20-way classification. Notably, SetFeat12$^*$ and SetFeat12 gain 6.18\% and 2.84\% over MixtFSL~\cite{afrasiyabi2021MixtFSL} in 5-way, respectively.
Additionally, last column of \cref{tab:More_way} shows cross domain adaptation, where we pre-train our model on miniImageNet and test on the CUB dataset. Here, our SetFeat12$^*$ obtains the second best and is 0.92\% below MixtFSL~\cite{afrasiyabi2021MixtFSL}.    
 
\begin{table*}[hbt!]
\centering
\caption{$N$-way 5-shot classification results on miniImageNet using ResNet and SetFeat.   $\pm$ denotes the $95\%$ confidence intervals over 600 episodes. The best results prior to this work is highlighted in red, and the best results are presented in boldface.} 
    \begin{tabular}{ rrcccccccc}  
        \toprule          
          &    &    &   &  & \textbf{miniImageNet}  &   &     &   & \textbf{miniImageNet$\xrightarrow{}$CUB}        \\  
          & Method   &Backbone   & \textbf{5-way}  &  & \textbf{10-way}  &   & \textbf{20-way}     &   & \textbf{5-way}      \\  
        \midrule

        & MatchingNet$^\ddag$~\cite{vinyals2016matching}   
        & \multirow{8}{*}{\rotatebox{90}{\ \ \ -------- ResNet18 --------}}  
        & 68.88\scriptsize{ $\pm$0.69}     &       
        & 52.27\scriptsize{ $\pm$0.46}     &        
        & 36.78\scriptsize{ $\pm$0.25}     &
        & --
        \\

    & Neg-Margin$^\ddag$~\cite{Bin_2020_ECCV_margin_matter}   
        & 
        & --    &       
        & --    &        
        & --   &
        & 67.03\scriptsize{ $\pm$0.80} 
        \\

        & ProtoNet$^\ddag$~\cite{snell2017prototypical}    &
        & 73.68\scriptsize{ $\pm$0.65}        &         
        & 59.22\scriptsize{ $\pm$0.44}        &       
        & 44.96\scriptsize{ $\pm$0.26}        &
        & 62.02\scriptsize{ $\pm$0.70} 
        \\

        & RelationNet$^\ddag$~\cite{sung2018learning}             &
        & 69.83\scriptsize{ $\pm$0.68}                               &           
        & 53.88\scriptsize{ $\pm$0.48}                                 &        
        & 39.17\scriptsize{ $\pm$0.25}         & 
        & 57.71\scriptsize{ $\pm$0.70}
        \\      
  
        & Baseline~\cite{chen2019closer}                &
        & 74.27\scriptsize{ $\pm$0.63}                   &           
        & 55.00\scriptsize{ $\pm$0.46}                  &
        & 42.03\scriptsize{ $\pm$0.25}         &
        & 65.57\scriptsize{ $\pm$0.25} 
        \\

        & Baseline++~\cite{chen2019closer}              &
        & 75.68\scriptsize{ $\pm$0.63}                &        
        & 63.40\scriptsize{ $\pm$0.44}                 &
        & 50.85\scriptsize{ $\pm$0.25}                 &
        & 64.38\scriptsize{ $\pm$0.90}
        \\

        & Pos-Margin~\cite{afrasiyabi2021MixtFSL}                     &
        & 76.62\scriptsize{ $\pm$0.58}                 &
        & 62.95\scriptsize{ $\pm$0.83}                &
        & 51.92\scriptsize{ $\pm$1.02}           &
        & 64.93\scriptsize{ $\pm$1.00}                 
        \\ 
 
        & MixtFSL~\cite{afrasiyabi2021MixtFSL}                                  &
        & {\color{red}77.76}\scriptsize{ $\pm$0.58}                                          &
        & {\color{red}64.18}\scriptsize{ $\pm$0.76}                                          &
        & {\color{red}53.15}\scriptsize{ $\pm$0.71}                                          &
        & \textbf{68.77}\scriptsize{ $\pm$0.90}
        \\[0.5em]  
        &  Sum-min (ours)                    &SetFeat12$^*$ 
            & \textbf{81.22}\scriptsize{$\pm$0.45}   &
            & \textbf{70.36}\scriptsize{$\pm$0.46}    &
            & \textbf{57.36}\scriptsize{$\pm$0.36}   &
            & {\color{red}67.85}\scriptsize{$\pm$0.70}
            \\*[0.2em]
        \midrule

        & MixtFSL~\cite{afrasiyabi2021MixtFSL}                                  &ResNet12
        & {\color{red}82.04}\scriptsize{ $\pm$0.49}                             &
        & {\color{red}68.26}\scriptsize{ $\pm$0.71}                             &
        & {\color{red}55.41}\scriptsize{ $\pm$0.71}              &
        & --
       \\*[0.2em] 
        
        &  Sum-min (ours)                    &SetFeat12$ \ $  
            & \textbf{82.71}\scriptsize{$\pm$0.46}   &
            & \textbf{71.10}\scriptsize{$\pm$0.46}    &
            & \textbf{57.97}\scriptsize{$\pm$0.36} &
            & --
            \\  
    \toprule
    \end{tabular}
    \\  
    $^\ddag$ implementation from \cite{chen2019closer}  
    \label{tab:More_way}
\end{table*}

\newpage
\section*{Visualizing mappers saliency}
\label{sec:graiden_map}

\begin{figure*} 
    \centering
    \footnotesize
    \setlength{\tabcolsep}{1pt}
\begin{tabular}{ccccccccccccc}  
    
    \includegraphics[width=0.08\linewidth, angle=0]{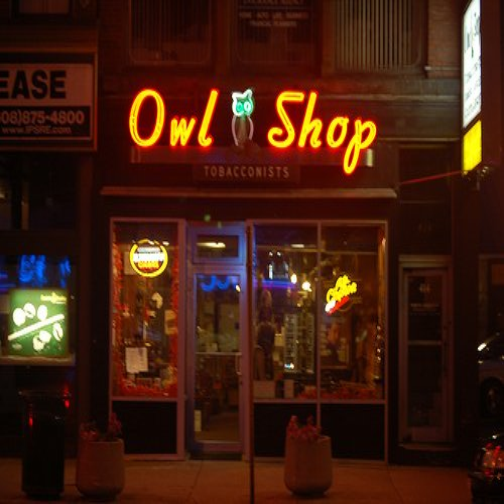} & 
    \includegraphics[width=0.08\linewidth, angle=0]{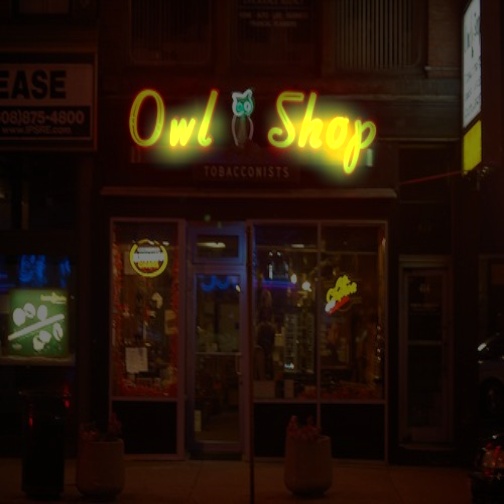} & 
    \includegraphics[width=0.08\linewidth, angle=0]{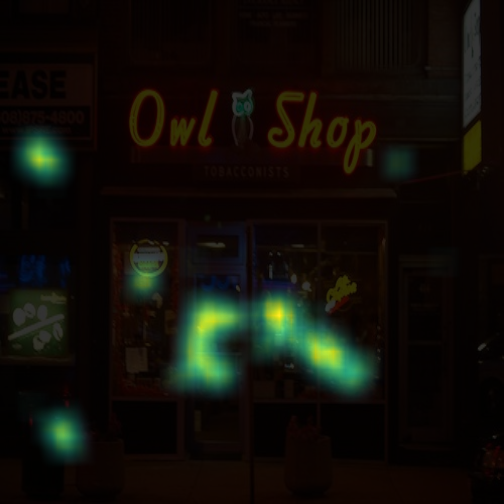} &
    \includegraphics[width=0.08\linewidth, angle=0]{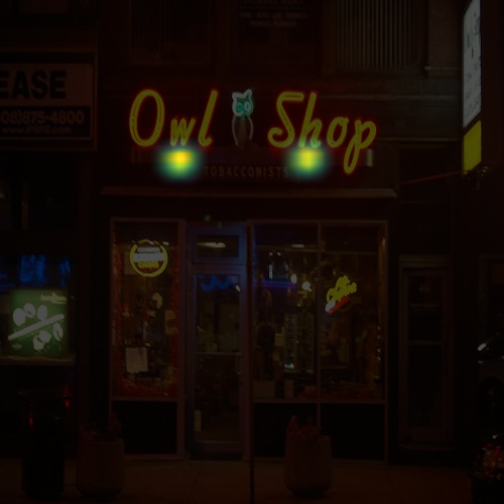} &  
    \includegraphics[width=0.08\linewidth, angle=0]{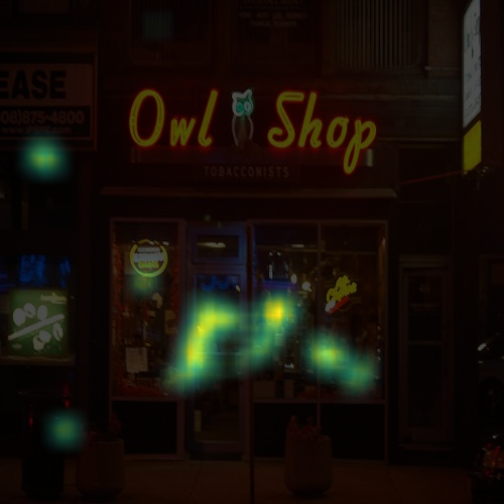} & 
    \includegraphics[width=0.08\linewidth, angle=0]{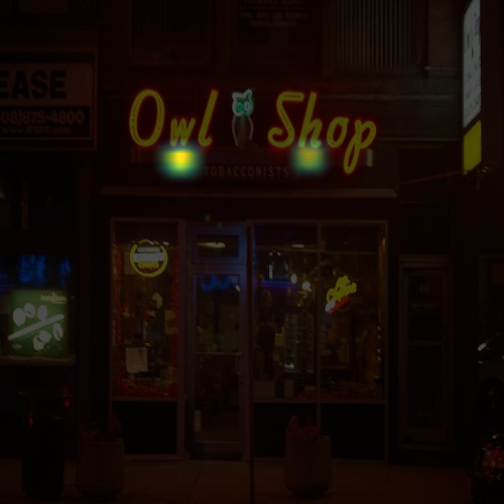} & 
    \includegraphics[width=0.08\linewidth, angle=0]{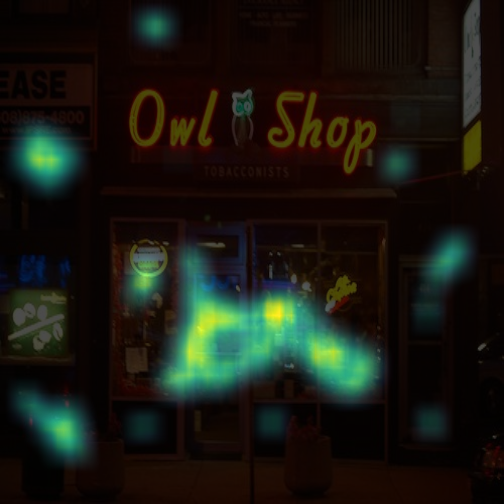} & 
    \includegraphics[width=0.08\linewidth, angle=0]{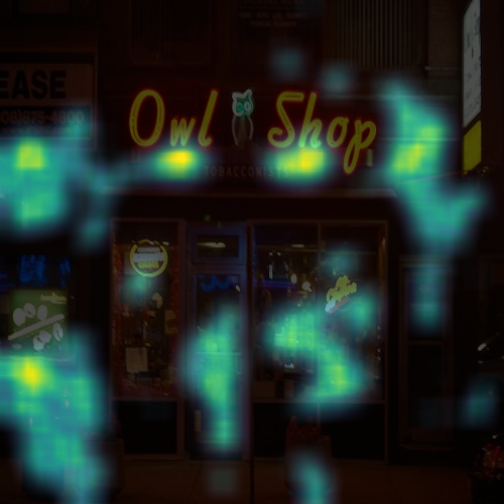} & 
    \includegraphics[width=0.08\linewidth, angle=0]{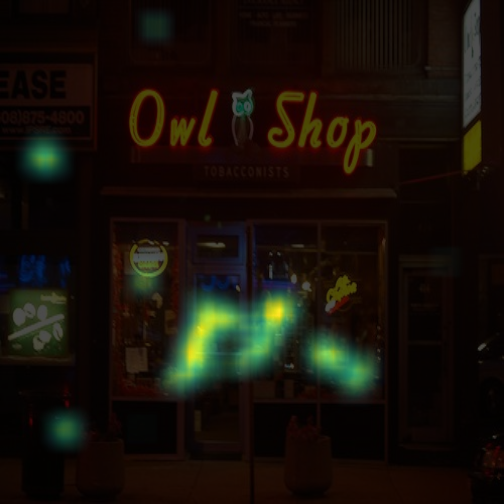} & 
    \includegraphics[width=0.08\linewidth, angle=0]{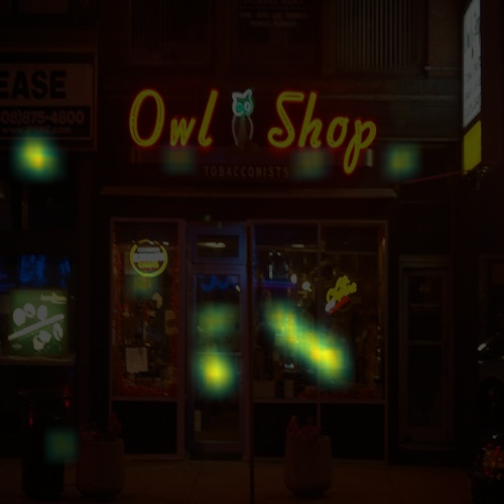} & 
    \includegraphics[width=0.08\linewidth, angle=0]{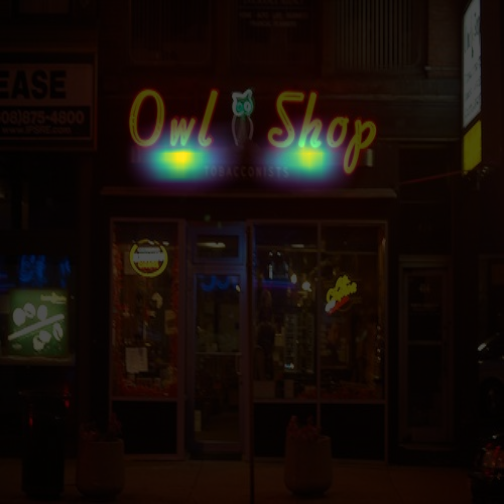} & 
    \includegraphics[width=0.08\linewidth, angle=0]{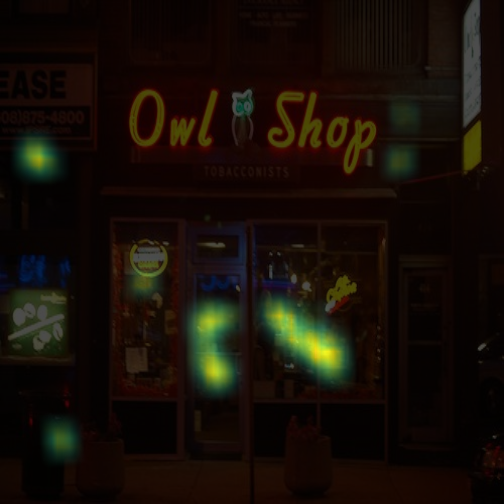} & 
    \\
    
     \includegraphics[width=0.08\linewidth, angle=0]{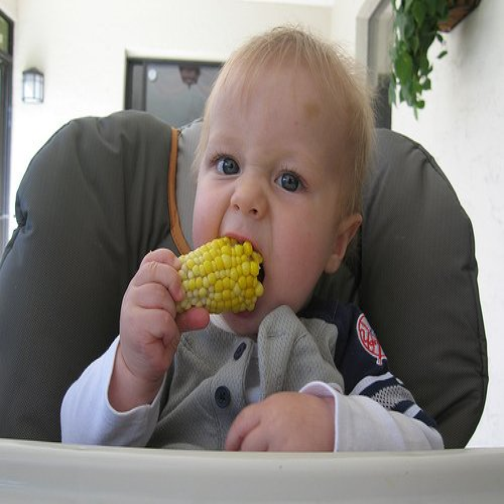} & 
    \includegraphics[width=0.08\linewidth, angle=0]{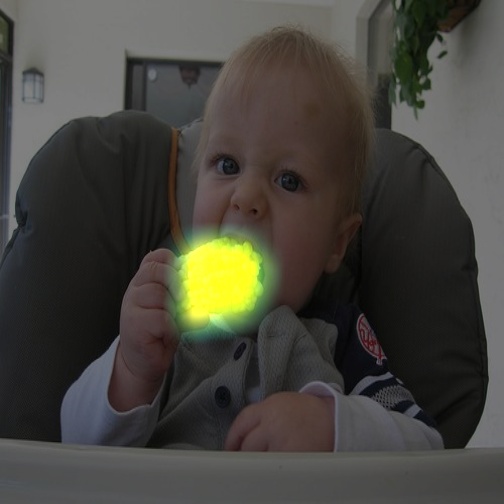} & 
    \includegraphics[width=0.08\linewidth, angle=0]{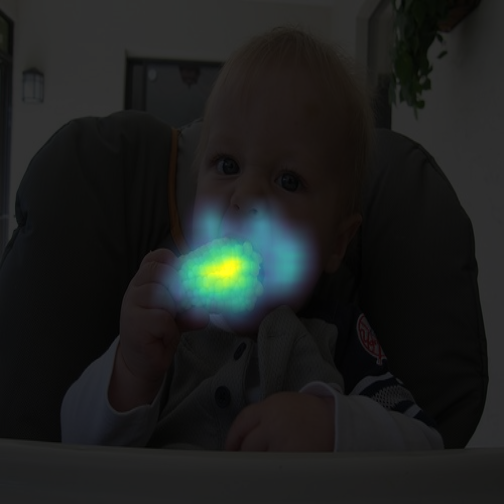} &
    \includegraphics[width=0.08\linewidth, angle=0]{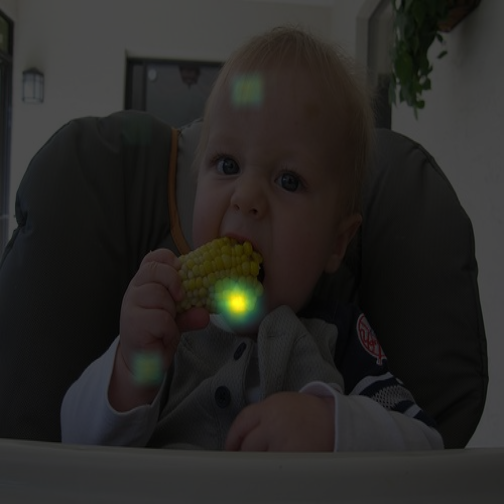} &  
    \includegraphics[width=0.08\linewidth, angle=0]{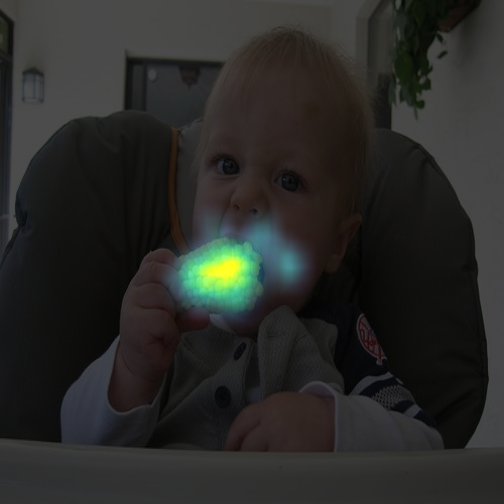} & 
    \includegraphics[width=0.08\linewidth, angle=0]{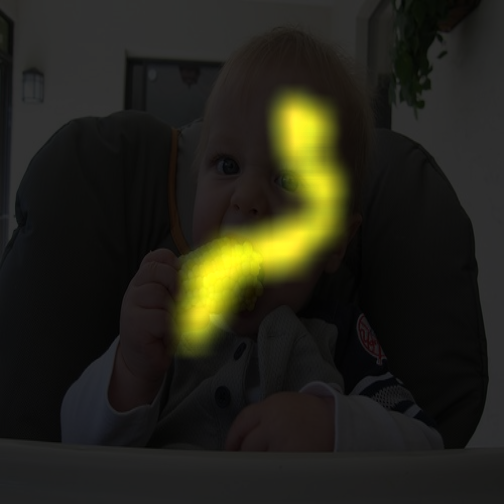} & 
    \includegraphics[width=0.08\linewidth, angle=0]{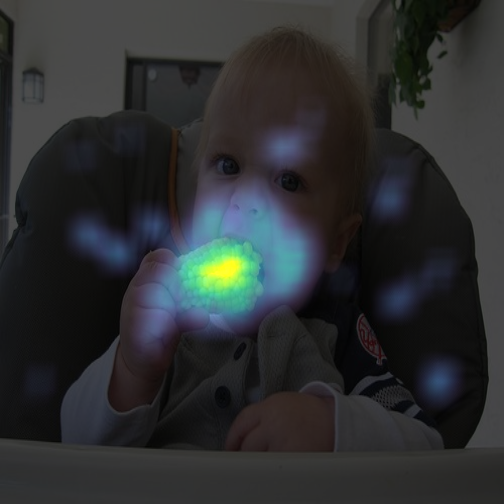} & 
    \includegraphics[width=0.08\linewidth, angle=0]{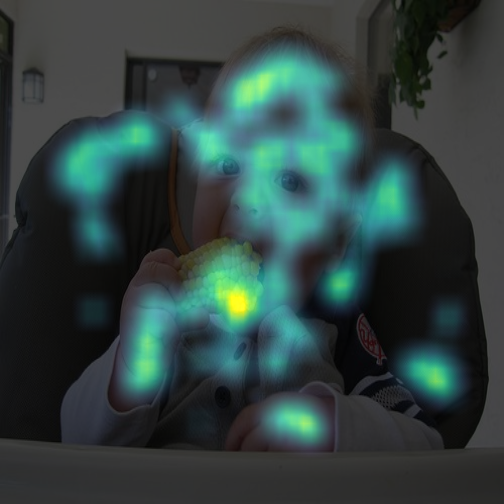} & 
    \includegraphics[width=0.08\linewidth, angle=0]{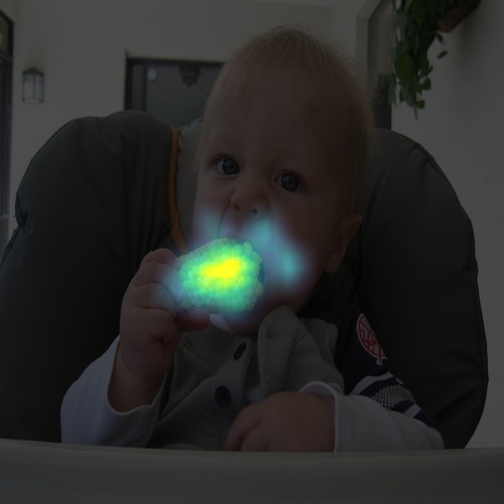} & 
    \includegraphics[width=0.08\linewidth, angle=0]{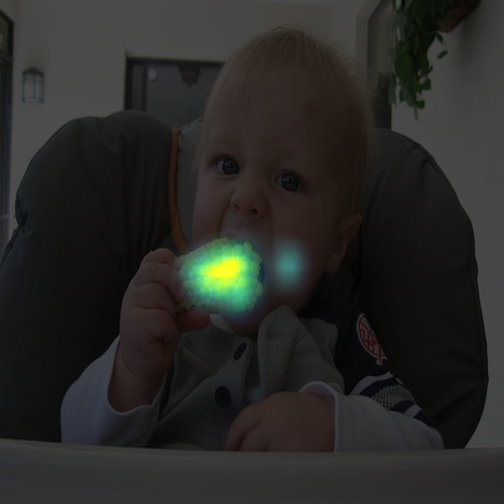} & 
    \includegraphics[width=0.08\linewidth, angle=0]{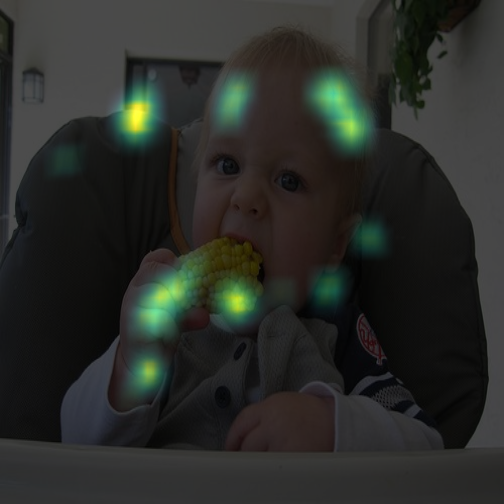} & 
    \includegraphics[width=0.08\linewidth, angle=0]{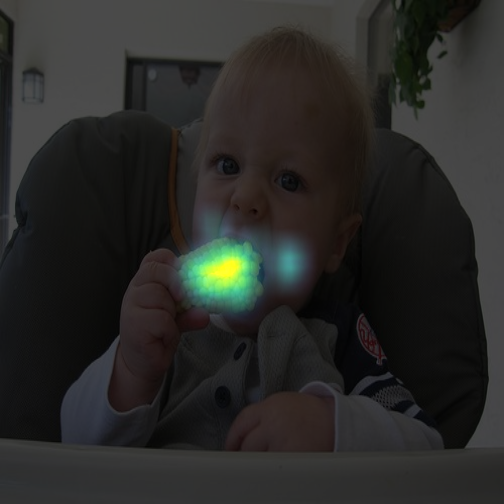} & 
    \\
    
     \includegraphics[width=0.08\linewidth, angle=0]{figures_and_plots/g_viz/train_n0450941700000097/ori_.png} & 
    \includegraphics[width=0.08\linewidth, angle=0]{figures_and_plots/g_viz/train_n0450941700000097/cross_entropy.jpg} & 
    \includegraphics[width=0.08\linewidth, angle=0]{figures_and_plots/g_viz/train_n0450941700000097/g_0_.png} &
    \includegraphics[width=0.08\linewidth, angle=0]{figures_and_plots/g_viz/train_n0450941700000097/g_1_.png} &  
    \includegraphics[width=0.08\linewidth, angle=0]{figures_and_plots/g_viz/train_n0450941700000097/g_2_.png} & 
    \includegraphics[width=0.08\linewidth, angle=0]{figures_and_plots/g_viz/train_n0450941700000097/g_3_.png} & 
    \includegraphics[width=0.08\linewidth, angle=0]{figures_and_plots/g_viz/train_n0450941700000097/g_4_.png} & 
    \includegraphics[width=0.08\linewidth, angle=0]{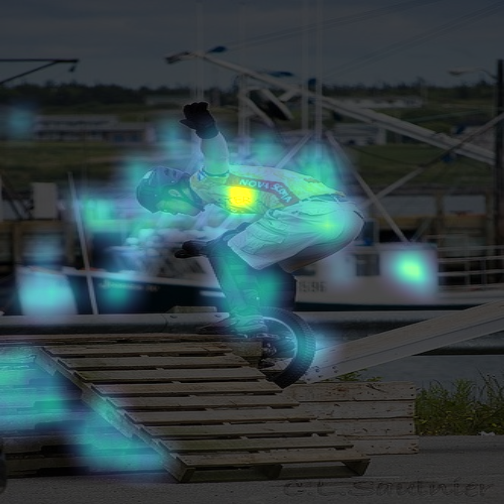} & 
    \includegraphics[width=0.08\linewidth, angle=0]{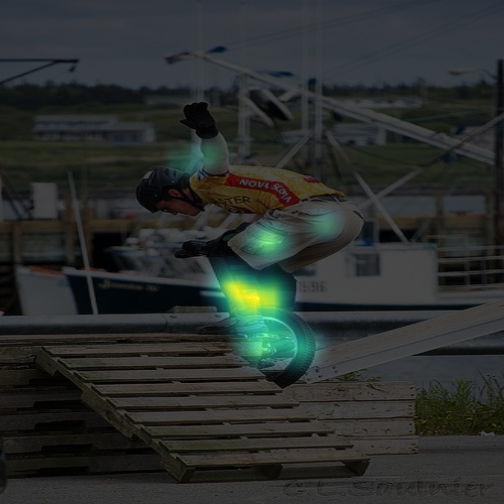} & 
    \includegraphics[width=0.08\linewidth, angle=0]{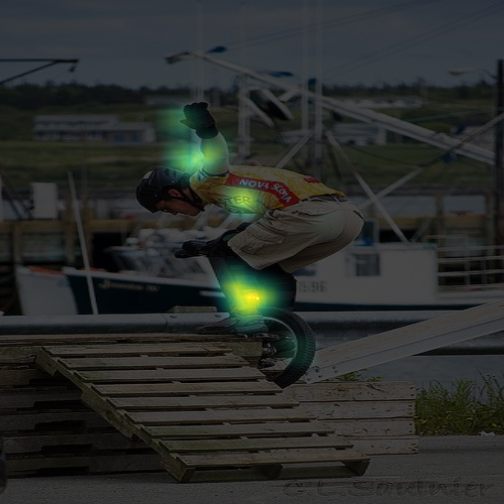} & 
    \includegraphics[width=0.08\linewidth, angle=0]{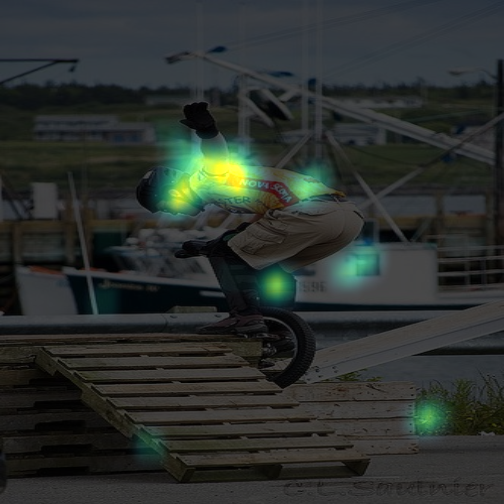} & 
    \includegraphics[width=0.08\linewidth, angle=0]{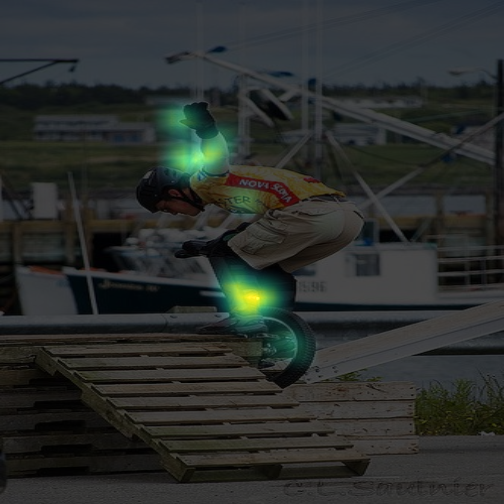} & 
    \\
    
     \includegraphics[width=0.08\linewidth, angle=0]{figures_and_plots/g_viz/train_n0210855100000112/ori_.png} & 
    \includegraphics[width=0.08\linewidth, angle=0]{figures_and_plots/g_viz/train_n0210855100000112/cross_entropy.jpg} & 
    \includegraphics[width=0.08\linewidth, angle=0]{figures_and_plots/g_viz/train_n0210855100000112/g_0_.png} &
    \includegraphics[width=0.08\linewidth, angle=0]{figures_and_plots/g_viz/train_n0210855100000112/g_1_.png} &  
    \includegraphics[width=0.08\linewidth, angle=0]{figures_and_plots/g_viz/train_n0210855100000112/g_2_.png} & 
    \includegraphics[width=0.08\linewidth, angle=0]{figures_and_plots/g_viz/train_n0210855100000112/g_3_.png} & 
    \includegraphics[width=0.08\linewidth, angle=0]{figures_and_plots/g_viz/train_n0210855100000112/g_4_.png} & 
    \includegraphics[width=0.08\linewidth, angle=0]{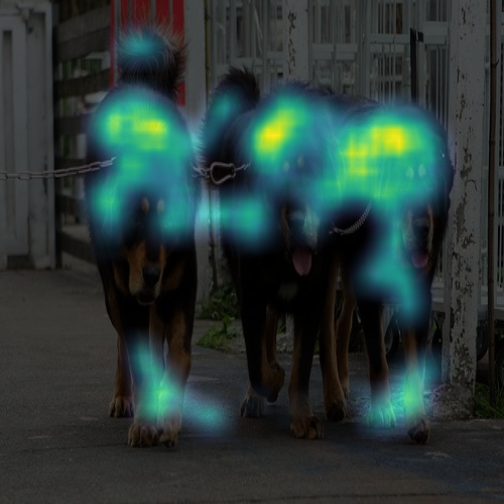} & 
    \includegraphics[width=0.08\linewidth, angle=0]{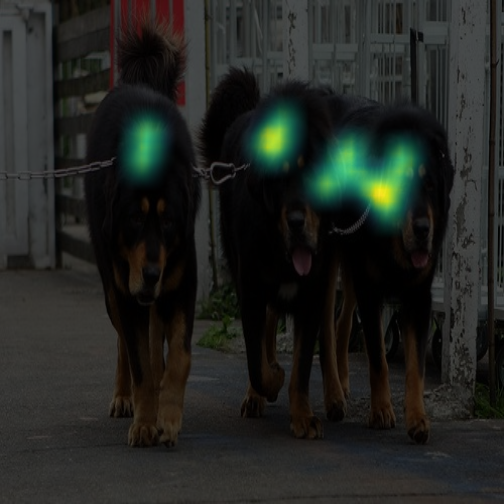} & 
    \includegraphics[width=0.08\linewidth, angle=0]{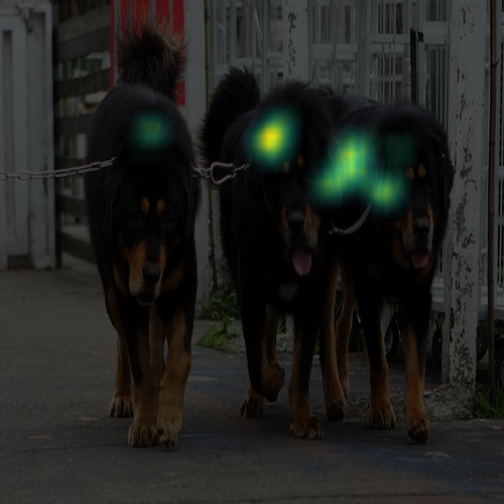} & 
    \includegraphics[width=0.08\linewidth, angle=0]{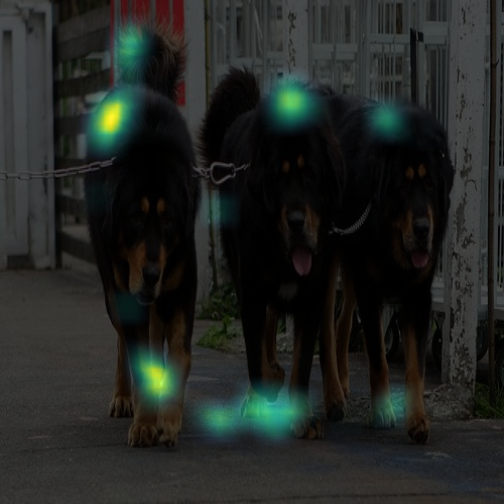} & 
    \includegraphics[width=0.08\linewidth, angle=0]{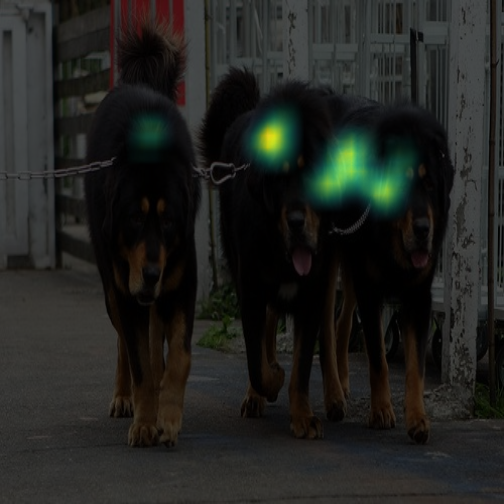} & 
    \\
    
     \includegraphics[width=0.08\linewidth, angle=0]{figures_and_plots/g_viz/train_n0461250400000071/ori_.png} & 
    \includegraphics[width=0.08\linewidth, angle=0]{figures_and_plots/g_viz/train_n0461250400000071/cross_entropy.jpg} & 
    \includegraphics[width=0.08\linewidth, angle=0]{figures_and_plots/g_viz/train_n0461250400000071/g_0_.png} &
    \includegraphics[width=0.08\linewidth, angle=0]{figures_and_plots/g_viz/train_n0461250400000071/g_1_.png} &  
    \includegraphics[width=0.08\linewidth, angle=0]{figures_and_plots/g_viz/train_n0461250400000071/g_2_.png} & 
    \includegraphics[width=0.08\linewidth, angle=0]{figures_and_plots/g_viz/train_n0461250400000071/g_3_.png} & 
    \includegraphics[width=0.08\linewidth, angle=0]{figures_and_plots/g_viz/train_n0461250400000071/g_4_.png} & 
    \includegraphics[width=0.08\linewidth, angle=0]{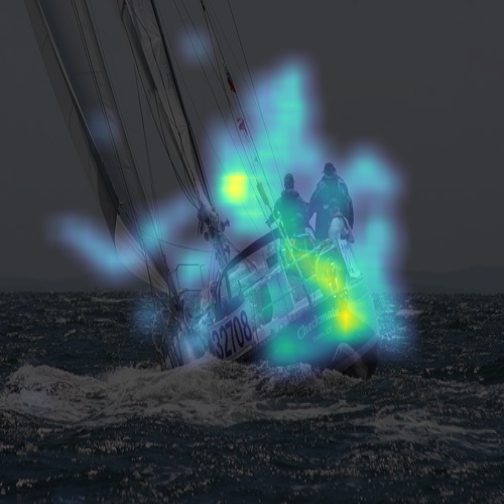} & 
    \includegraphics[width=0.08\linewidth, angle=0]{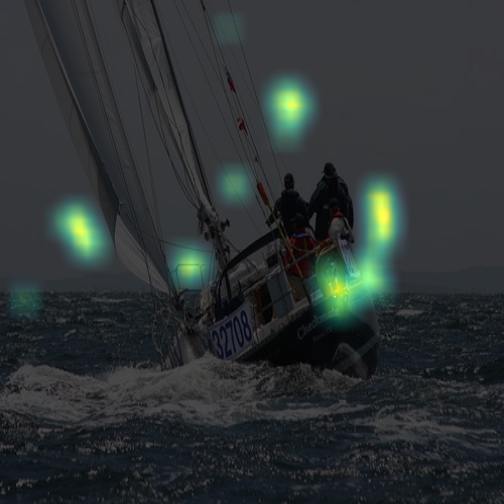} & 
    \includegraphics[width=0.08\linewidth, angle=0]{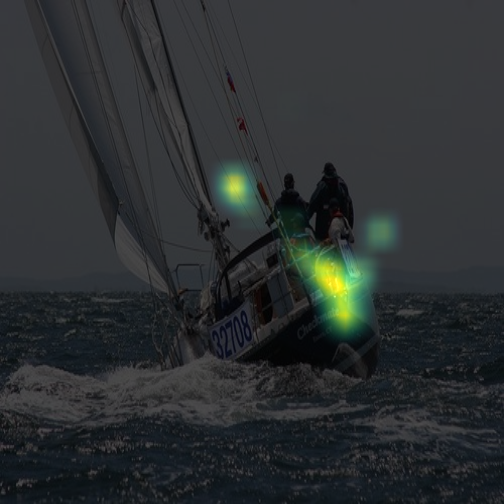} & 
    \includegraphics[width=0.08\linewidth, angle=0]{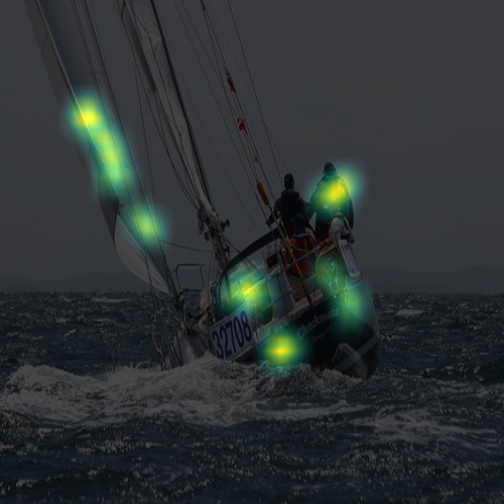} & 
    \includegraphics[width=0.08\linewidth, angle=0]{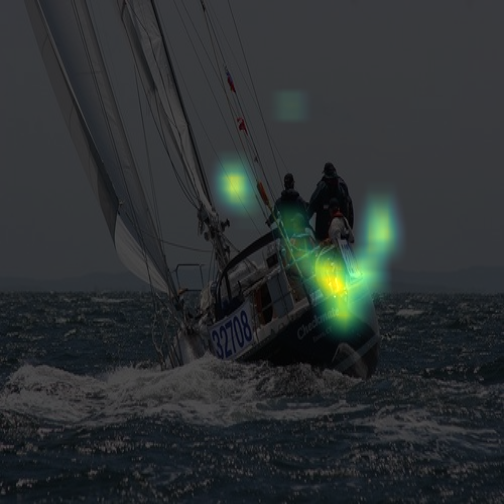} & 
    \\
    \\
    
    \includegraphics[width=0.08\linewidth, angle=0]{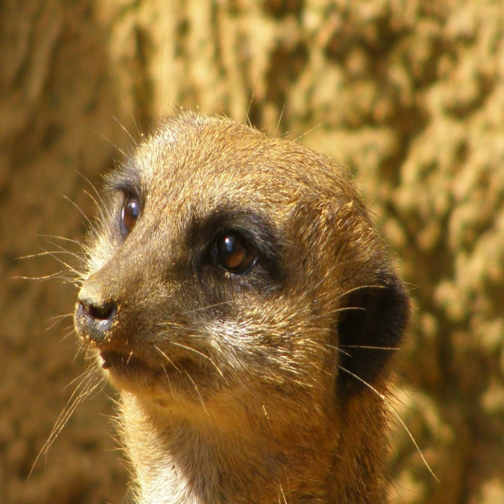} & 
    \includegraphics[width=0.08\linewidth, angle=0]{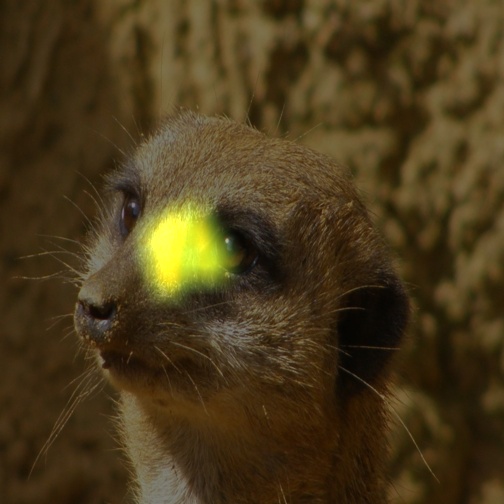} & 
    \includegraphics[width=0.08\linewidth, angle=0]{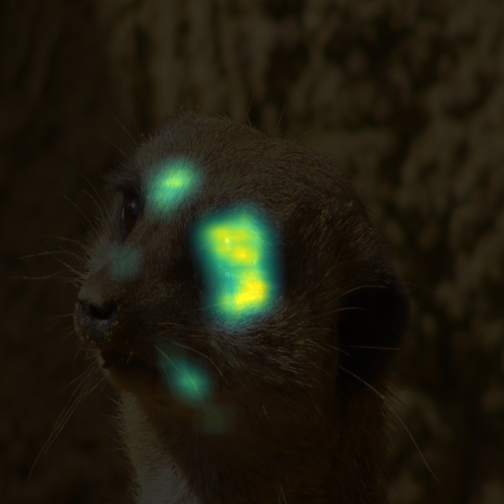} &
    \includegraphics[width=0.08\linewidth, angle=0]{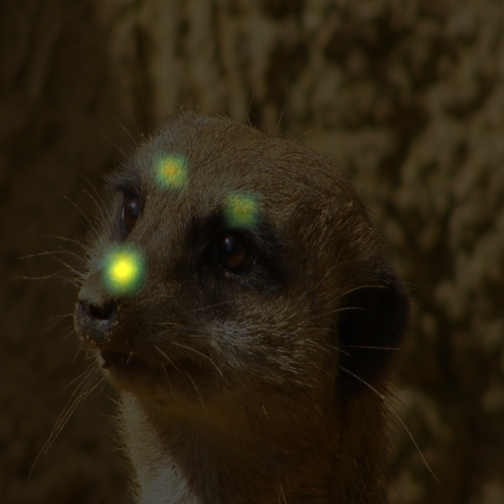} &  
    \includegraphics[width=0.08\linewidth, angle=0]{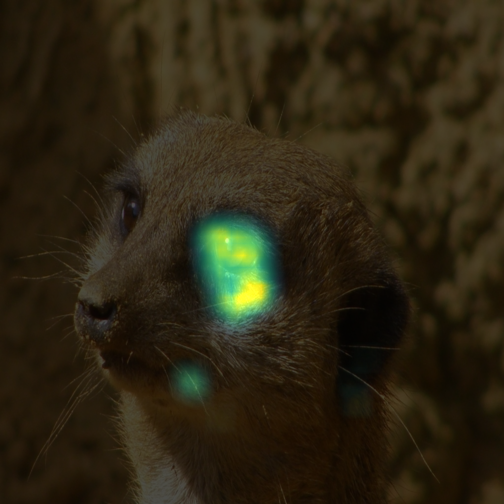} & 
    \includegraphics[width=0.08\linewidth, angle=0]{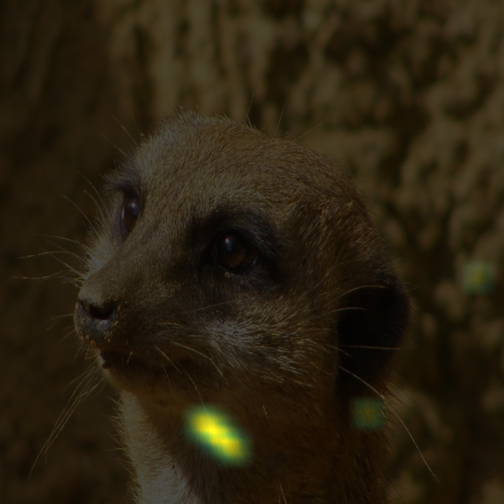} & 
    \includegraphics[width=0.08\linewidth, angle=0]{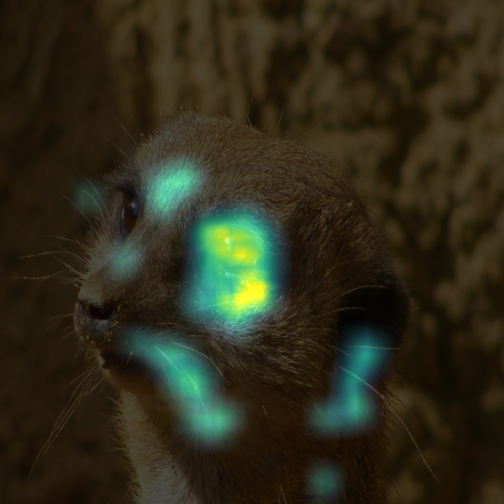} & 
    \includegraphics[width=0.08\linewidth, angle=0]{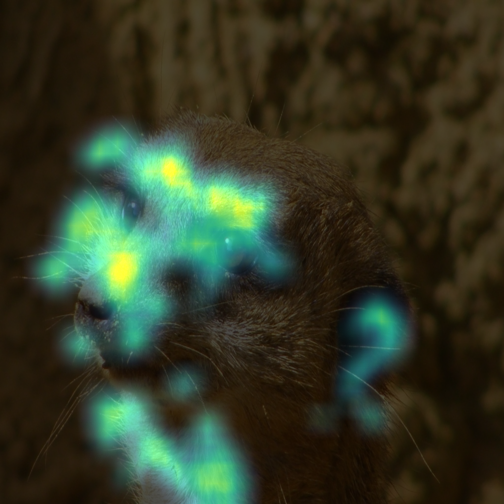} & 
    \includegraphics[width=0.08\linewidth, angle=0]{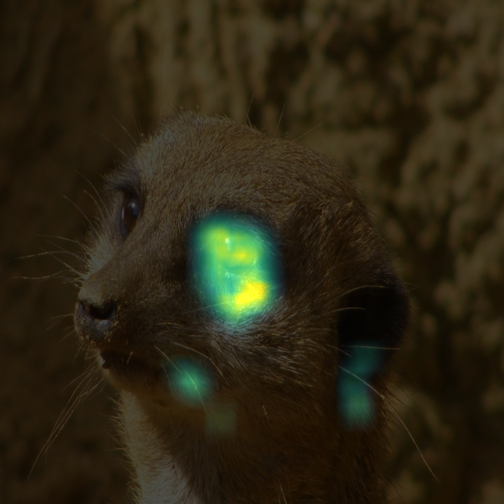} & 
    \includegraphics[width=0.08\linewidth, angle=0]{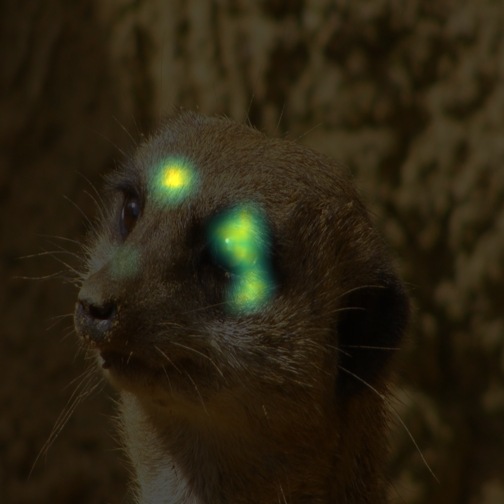} & 
    \includegraphics[width=0.08\linewidth, angle=0]{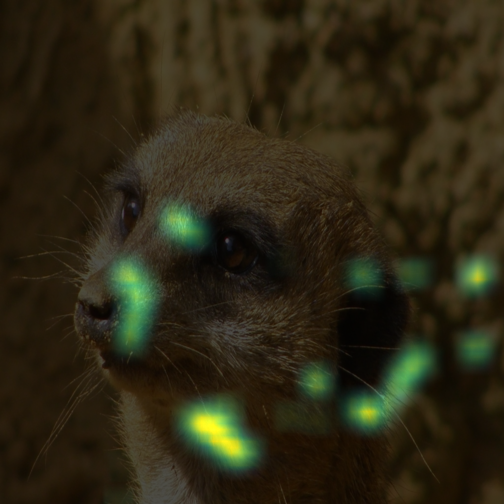} & 
    \includegraphics[width=0.08\linewidth, angle=0]{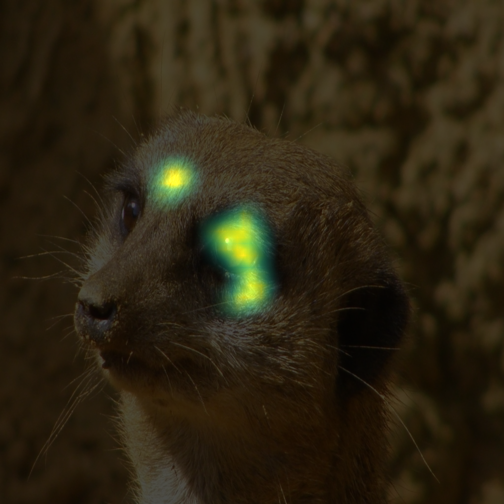} & 
    \\
 
    \includegraphics[width=0.08\linewidth, angle=0]{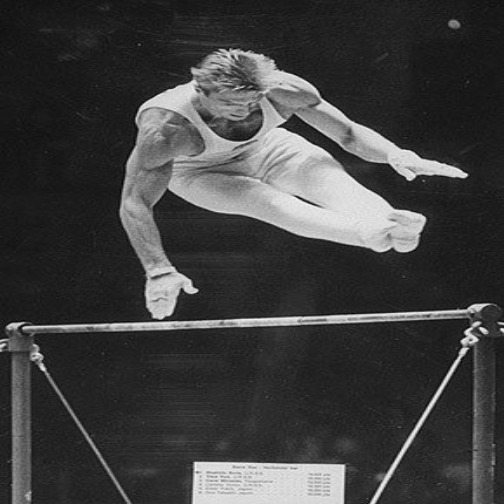} & 
    \includegraphics[width=0.08\linewidth, angle=0]{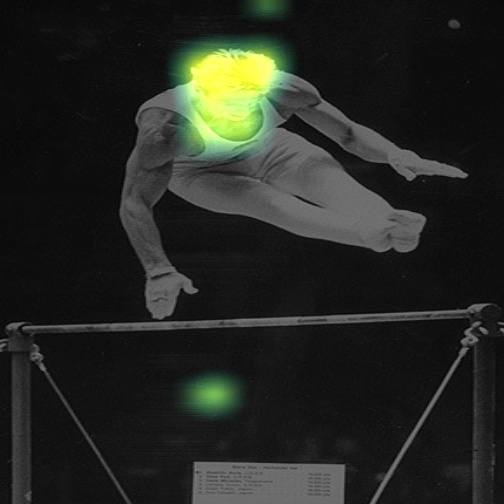} & 
    \includegraphics[width=0.08\linewidth, angle=0]{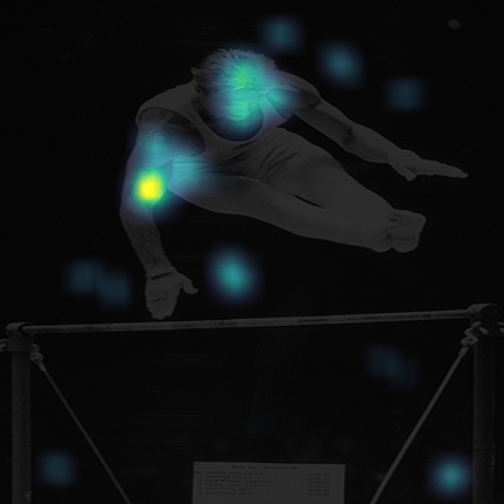} &
    \includegraphics[width=0.08\linewidth, angle=0]{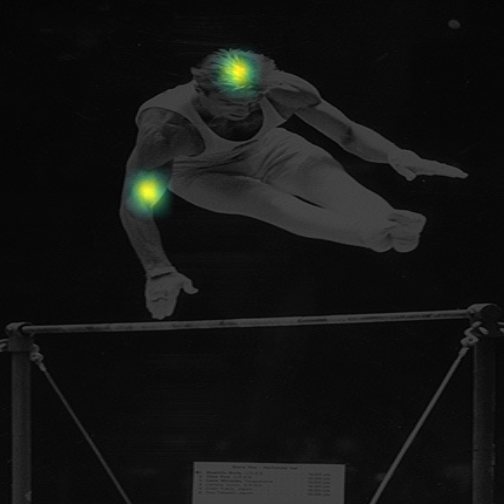} &  
    \includegraphics[width=0.08\linewidth, angle=0]{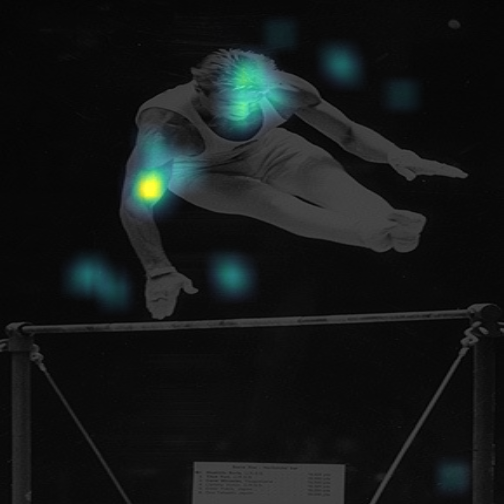} & 
    \includegraphics[width=0.08\linewidth, angle=0]{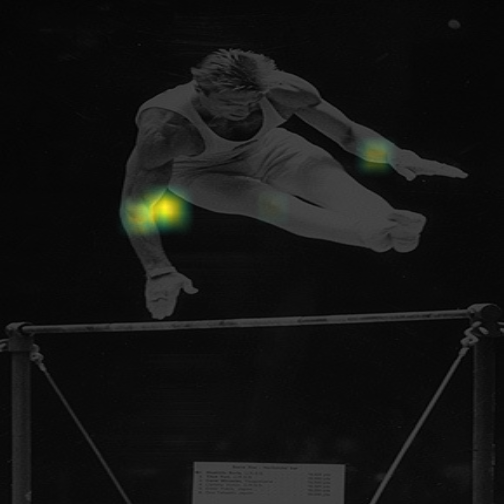} & 
    \includegraphics[width=0.08\linewidth, angle=0]{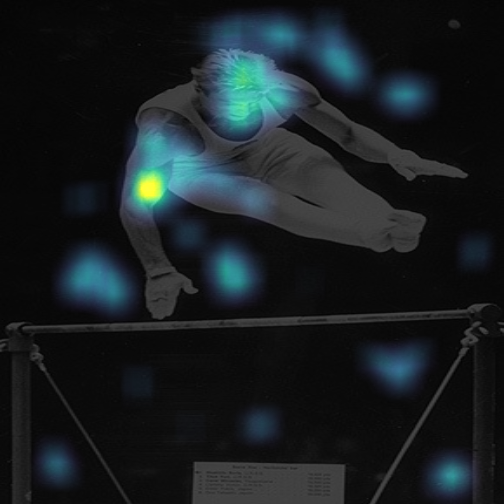} & 
    \includegraphics[width=0.08\linewidth, angle=0]{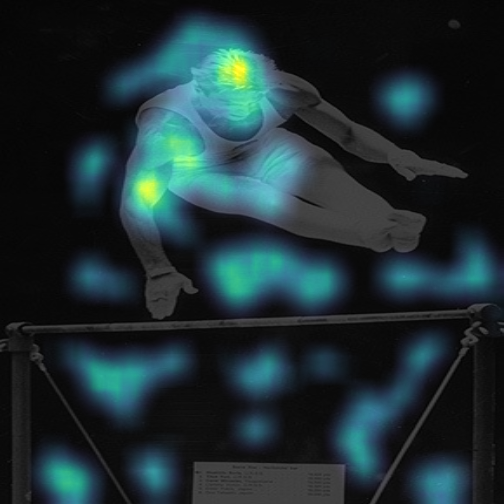} & 
    \includegraphics[width=0.08\linewidth, angle=0]{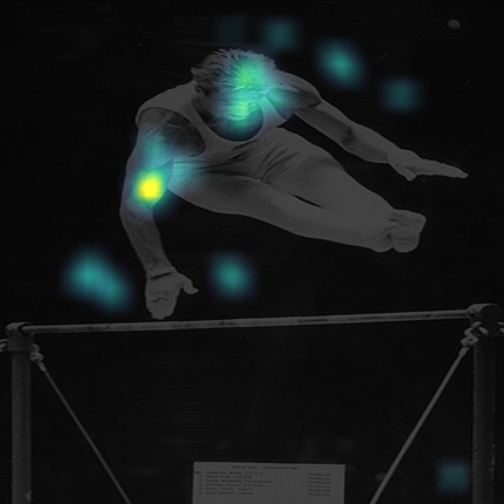} & 
    \includegraphics[width=0.08\linewidth, angle=0]{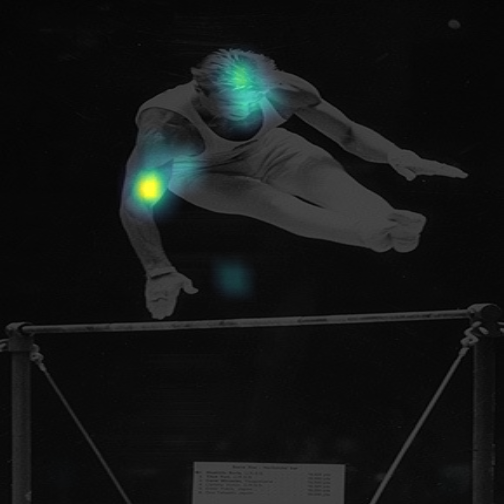} & 
    \includegraphics[width=0.08\linewidth, angle=0]{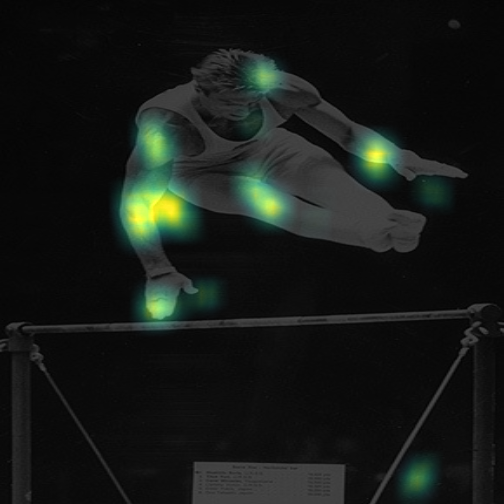} & 
    \includegraphics[width=0.08\linewidth, angle=0]{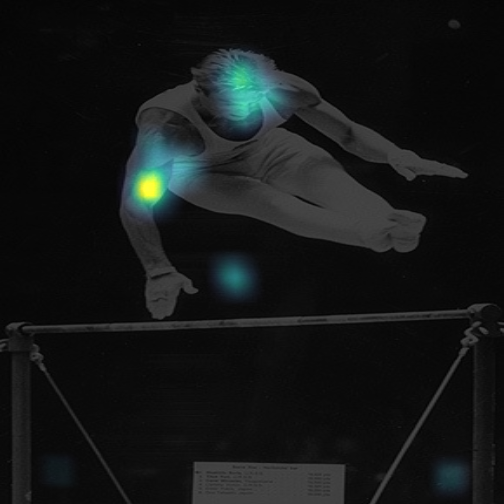} & 
    \\
    
     \includegraphics[width=0.08\linewidth, angle=0]{figures_and_plots/g_viz/valid_n0185567200000093/ori_.png} & 
    \includegraphics[width=0.08\linewidth, angle=0]{figures_and_plots/g_viz/valid_n0185567200000093/cross_entropy.jpg} & 
    \includegraphics[width=0.08\linewidth, angle=0]{figures_and_plots/g_viz/valid_n0185567200000093/g_0_.png} &
    \includegraphics[width=0.08\linewidth, angle=0]{figures_and_plots/g_viz/valid_n0185567200000093/g_1_.png} &  
    \includegraphics[width=0.08\linewidth, angle=0]{figures_and_plots/g_viz/valid_n0185567200000093/g_2_.png} & 
    \includegraphics[width=0.08\linewidth, angle=0]{figures_and_plots/g_viz/valid_n0185567200000093/g_3_.png} & 
    \includegraphics[width=0.08\linewidth, angle=0]{figures_and_plots/g_viz/valid_n0185567200000093/g_4_.png} & 
    \includegraphics[width=0.08\linewidth, angle=0]{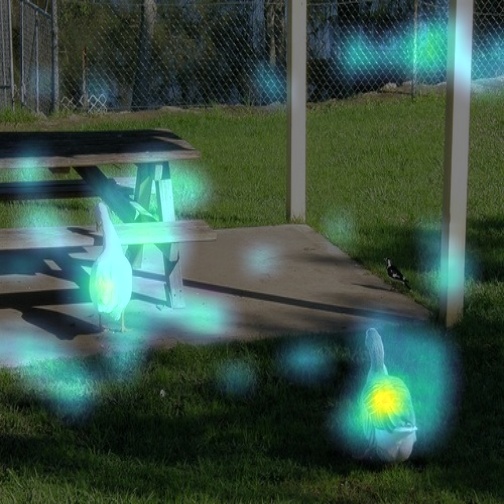} & 
    \includegraphics[width=0.08\linewidth, angle=0]{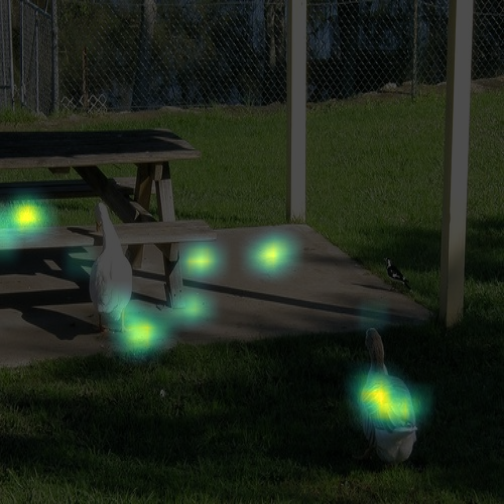} & 
    \includegraphics[width=0.08\linewidth, angle=0]{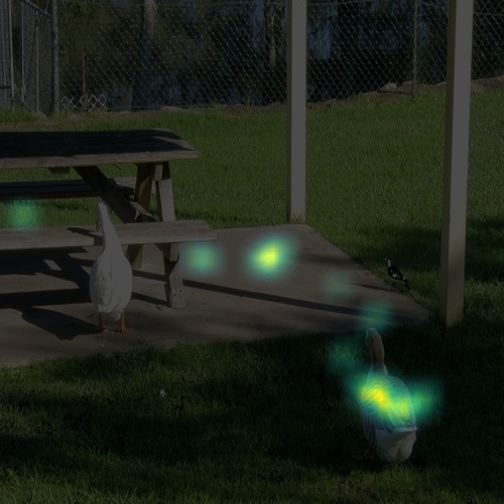} & 
    \includegraphics[width=0.08\linewidth, angle=0]{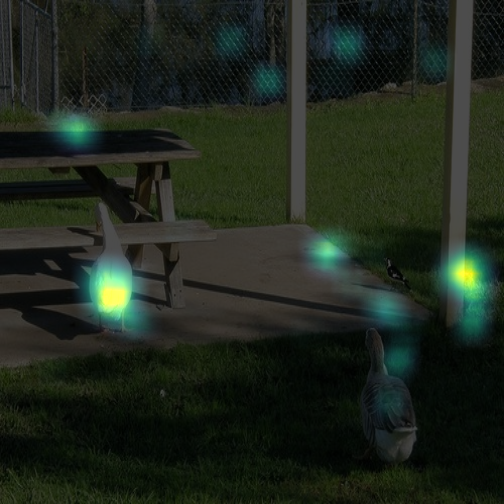} & 
    \includegraphics[width=0.08\linewidth, angle=0]{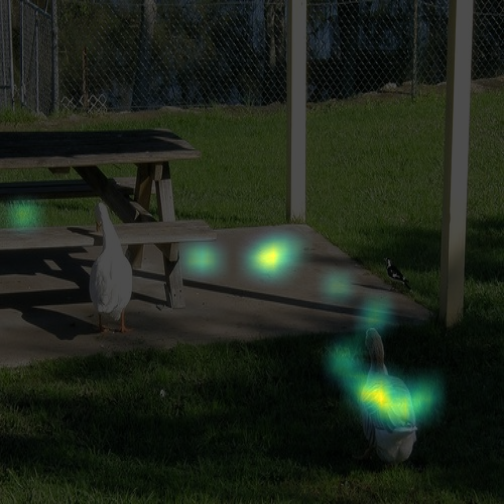} & 
    \\
    
     \includegraphics[width=0.08\linewidth, angle=0]{figures_and_plots/g_viz/valid_n0211454800000077/ori_.png} & 
    \includegraphics[width=0.08\linewidth, angle=0]{figures_and_plots/g_viz/valid_n0211454800000077/cross_entropy.jpg} & 
    \includegraphics[width=0.08\linewidth, angle=0]{figures_and_plots/g_viz/valid_n0211454800000077/g_0_.png} &
    \includegraphics[width=0.08\linewidth, angle=0]{figures_and_plots/g_viz/valid_n0211454800000077/g_1_.png} &  
    \includegraphics[width=0.08\linewidth, angle=0]{figures_and_plots/g_viz/valid_n0211454800000077/g_2_.png} & 
    \includegraphics[width=0.08\linewidth, angle=0]{figures_and_plots/g_viz/valid_n0211454800000077/g_3_.png} & 
    \includegraphics[width=0.08\linewidth, angle=0]{figures_and_plots/g_viz/valid_n0211454800000077/g_4_.png} & 
    \includegraphics[width=0.08\linewidth, angle=0]{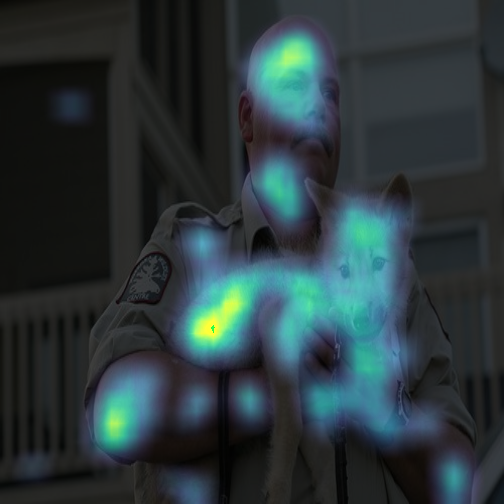} & 
    \includegraphics[width=0.08\linewidth, angle=0]{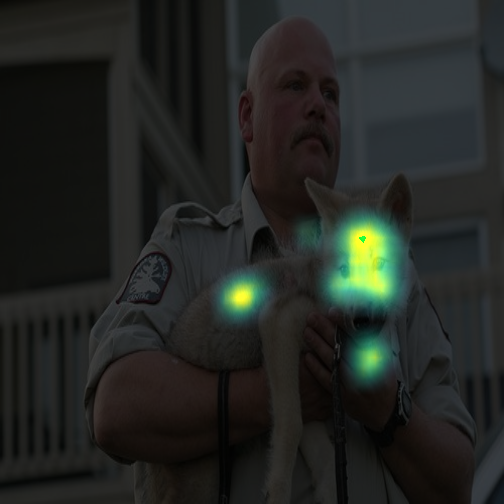} & 
    \includegraphics[width=0.08\linewidth, angle=0]{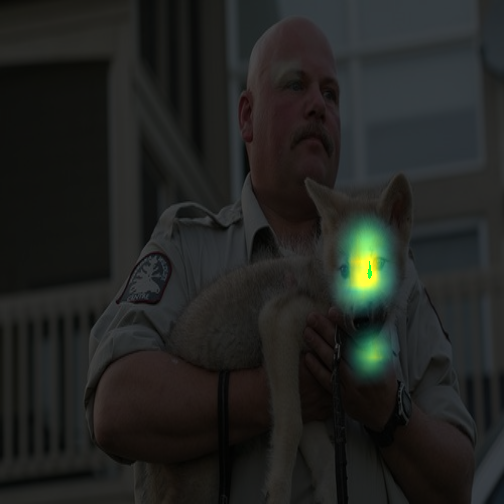} & 
    \includegraphics[width=0.08\linewidth, angle=0]{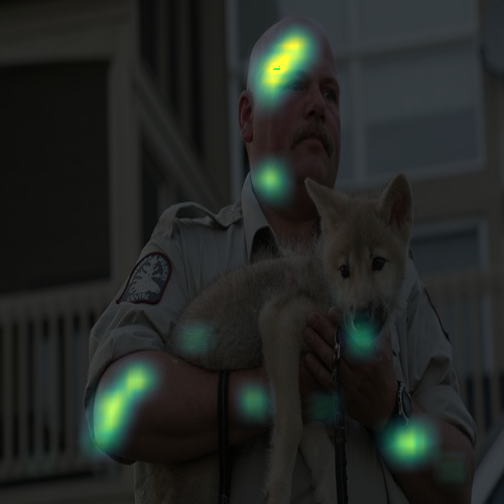} & 
    \includegraphics[width=0.08\linewidth, angle=0]{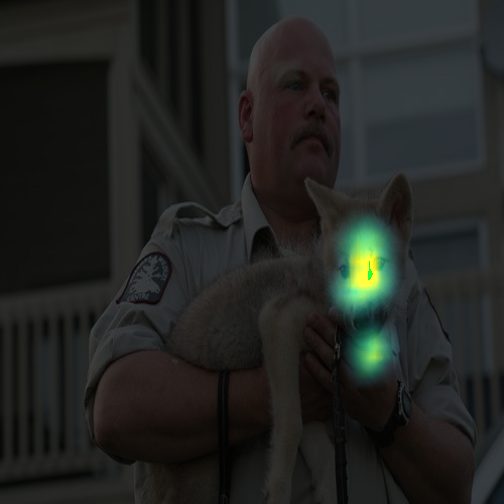} & 
    \\
    
    \includegraphics[width=0.08\linewidth, angle=0]{figures_and_plots/g_viz/valid_n0211454800000043/ori_.png} & 
    \includegraphics[width=0.08\linewidth, angle=0]{figures_and_plots/g_viz/valid_n0211454800000043/cross_entropy.jpg} & 
    \includegraphics[width=0.08\linewidth, angle=0]{figures_and_plots/g_viz/valid_n0211454800000043/g_0_.png} &
    \includegraphics[width=0.08\linewidth, angle=0]{figures_and_plots/g_viz/valid_n0211454800000043/g_1_.png} &  
    \includegraphics[width=0.08\linewidth, angle=0]{figures_and_plots/g_viz/valid_n0211454800000043/g_2_.png} & 
    \includegraphics[width=0.08\linewidth, angle=0]{figures_and_plots/g_viz/valid_n0211454800000043/g_3_.png} & 
    \includegraphics[width=0.08\linewidth, angle=0]{figures_and_plots/g_viz/valid_n0211454800000043/g_4_.png} & 
    \includegraphics[width=0.08\linewidth, angle=0]{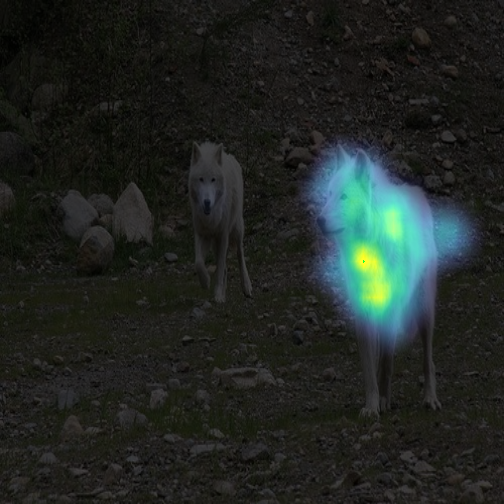} & 
    \includegraphics[width=0.08\linewidth, angle=0]{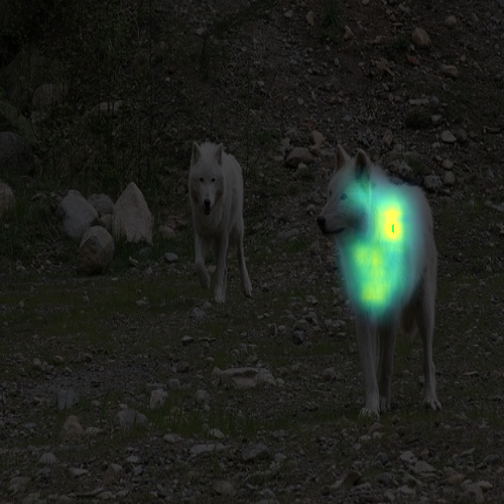} & 
    \includegraphics[width=0.08\linewidth, angle=0]{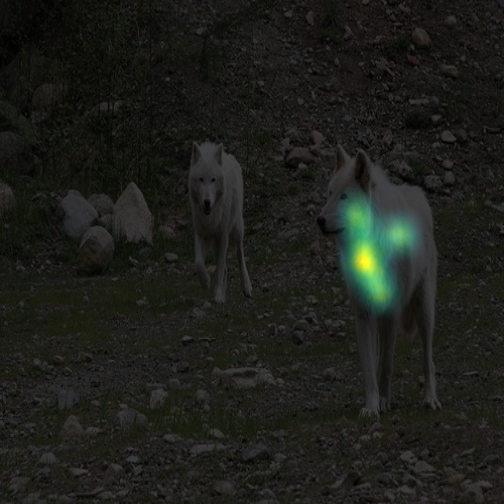} & 
    \includegraphics[width=0.08\linewidth, angle=0]{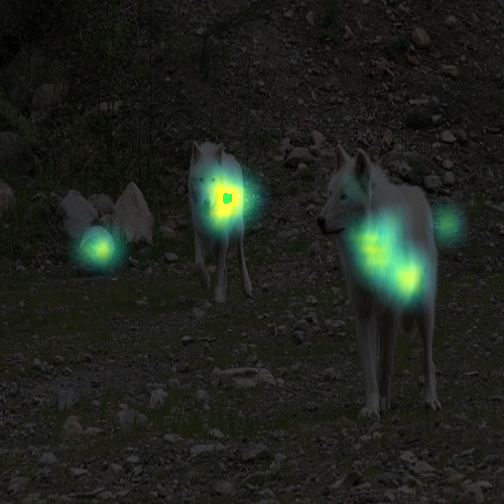} & 
    \includegraphics[width=0.08\linewidth, angle=0]{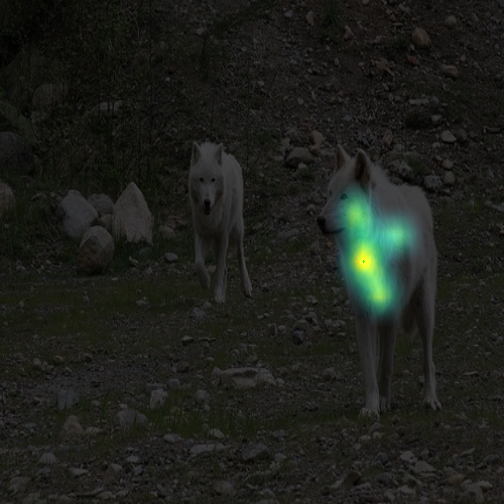} & 
    \\   
    input & 
    baseline~\cite{chen2019closer} &
    $g_1$ &
    $g_2$ &
    $g_3$ &
    $g_4$ &
    $g_5$ &
    $g_6$ &
    $g_7$ &
    $g_8$ &
    $g_9$ &
    $g_{10}$  \\

    \end{tabular}
    \vspace{0.5em}
    \caption{Gradient saliency maps after training SetFeat12 on miniImageNet. From left: input image, baseline~\cite{chen2019closer} trained with ResNet12, and 10 different mappers from our SetFeat12 ($g_i$ is the $i$-th mapper). The first five rows show examples from the training dataset, and the last five are from the validation set of miniImageNet.}
    \label{fig:gradientmap_SF12}
\end{figure*}

\Cref{fig:gradientmap_SF12,fig:gradientmap_SF4} compare the gradient saliency maps of SetFeat12 and SetFeat4-64 using our sum-min metric with ResNet12 and Conv4-64 using ``baseline'' from \cite{Chen_2019_CVPR}. Here SetFeat4-64 uses an FC-layer to compute mappers, while SetFeat12 uses a convolutional layer to do so. As shown in the figures, different mappers focus on different regions of the input image.

         
         
         

\newpage  
\section*{Class structure in cluster}
\label{sec:classStructure}  
Fig.~\ref{fig:tsne_mappers} shows that tSNE for each mapper independently exhibits the expected class structure for both validation (top row) and train (bottom row) sets. Since tSNE applied over all mappers jointly \textbf{on the validation set in fig.~4 of paper}, the largest variation (across mappers) is captured.

\begin{sidewaysfigure} 
    \centering
    \footnotesize
    \setlength{\tabcolsep}{1pt} 
    \begin{tabular}{ccccccccccc} 
    
    \rotatebox{90}{\hspace{2em} valid.}  
    \includegraphics[width=0.09\linewidth]{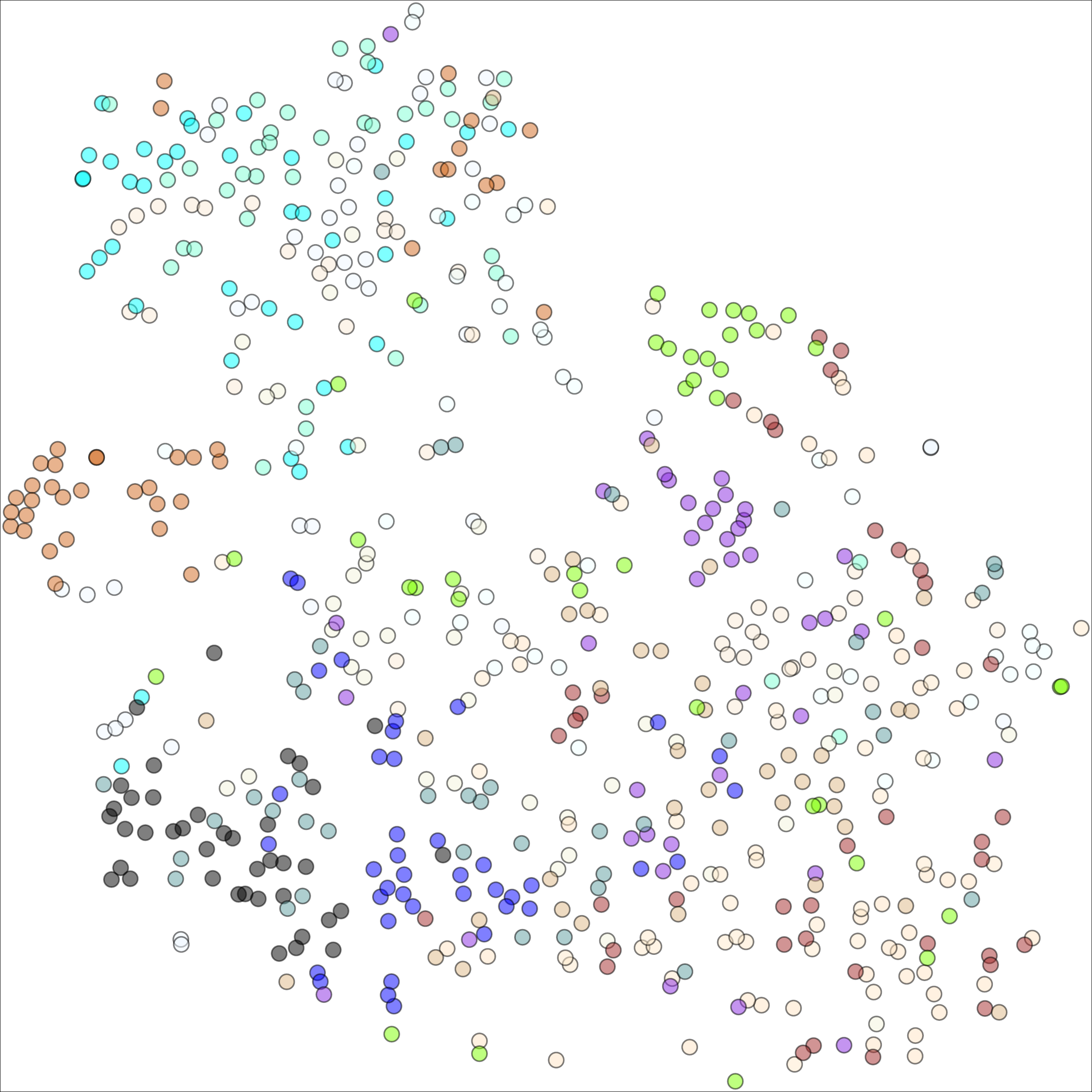} & 
    \includegraphics[width=0.09\linewidth]{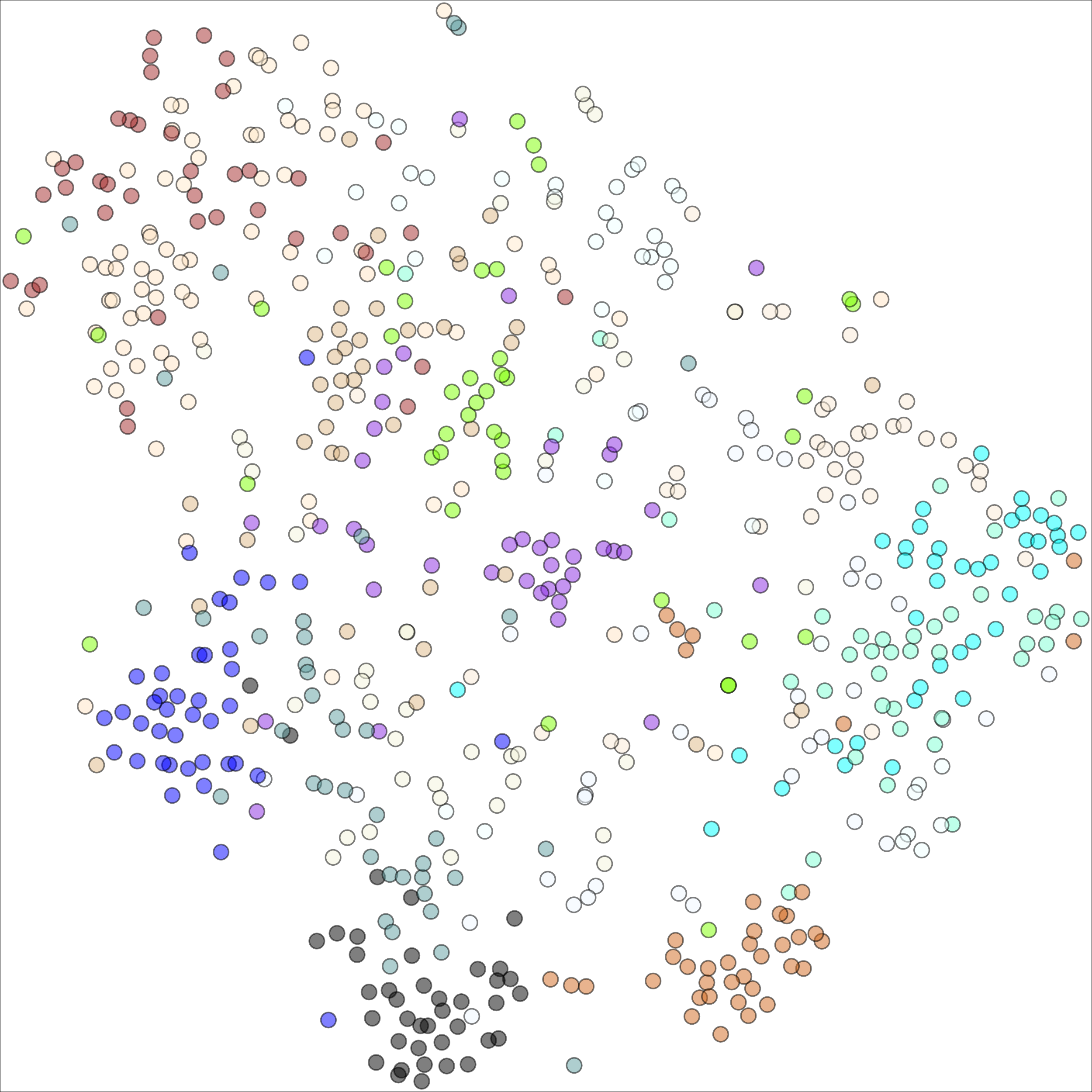} &   
    \includegraphics[width=0.09\linewidth]{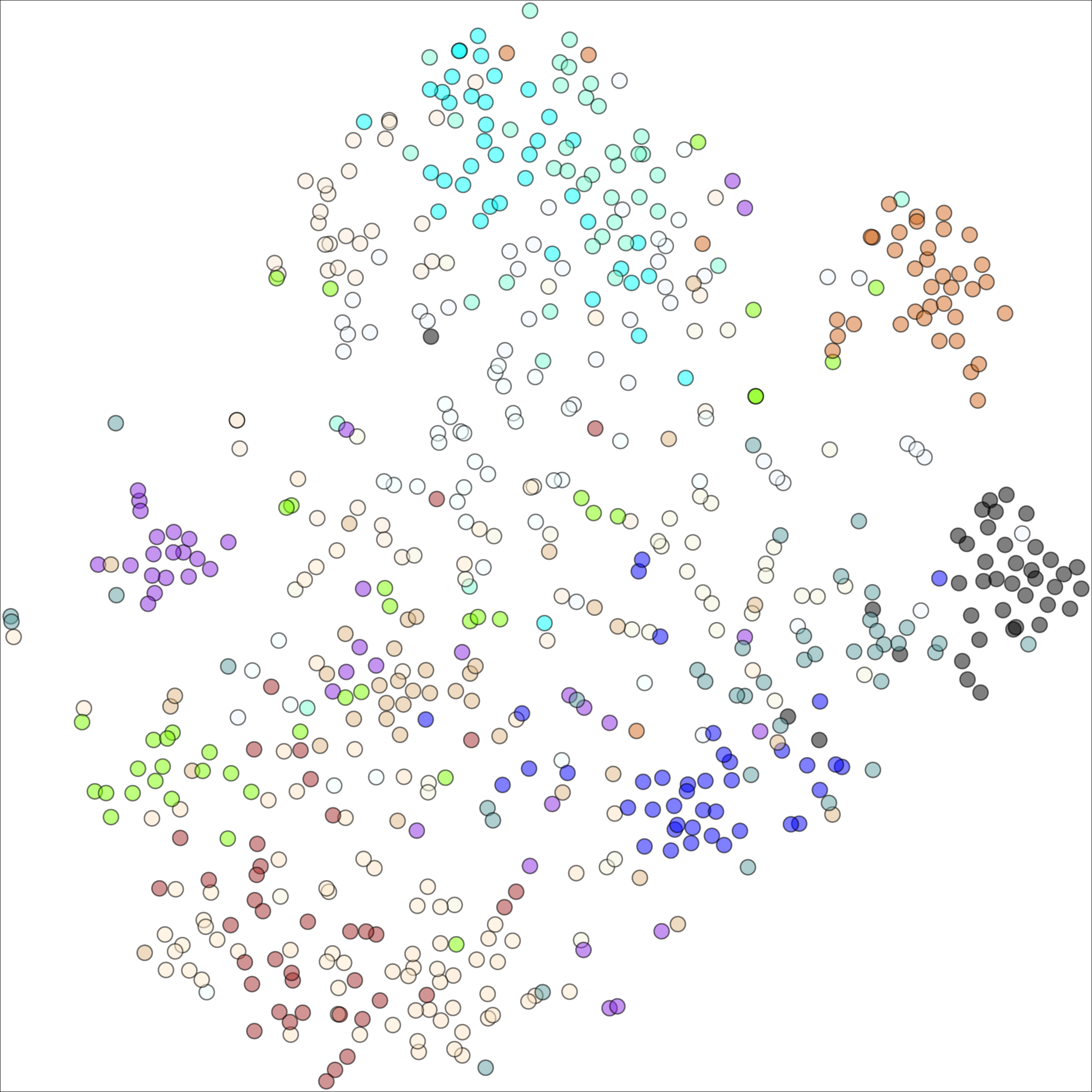} &   
    \includegraphics[width=0.09\linewidth]{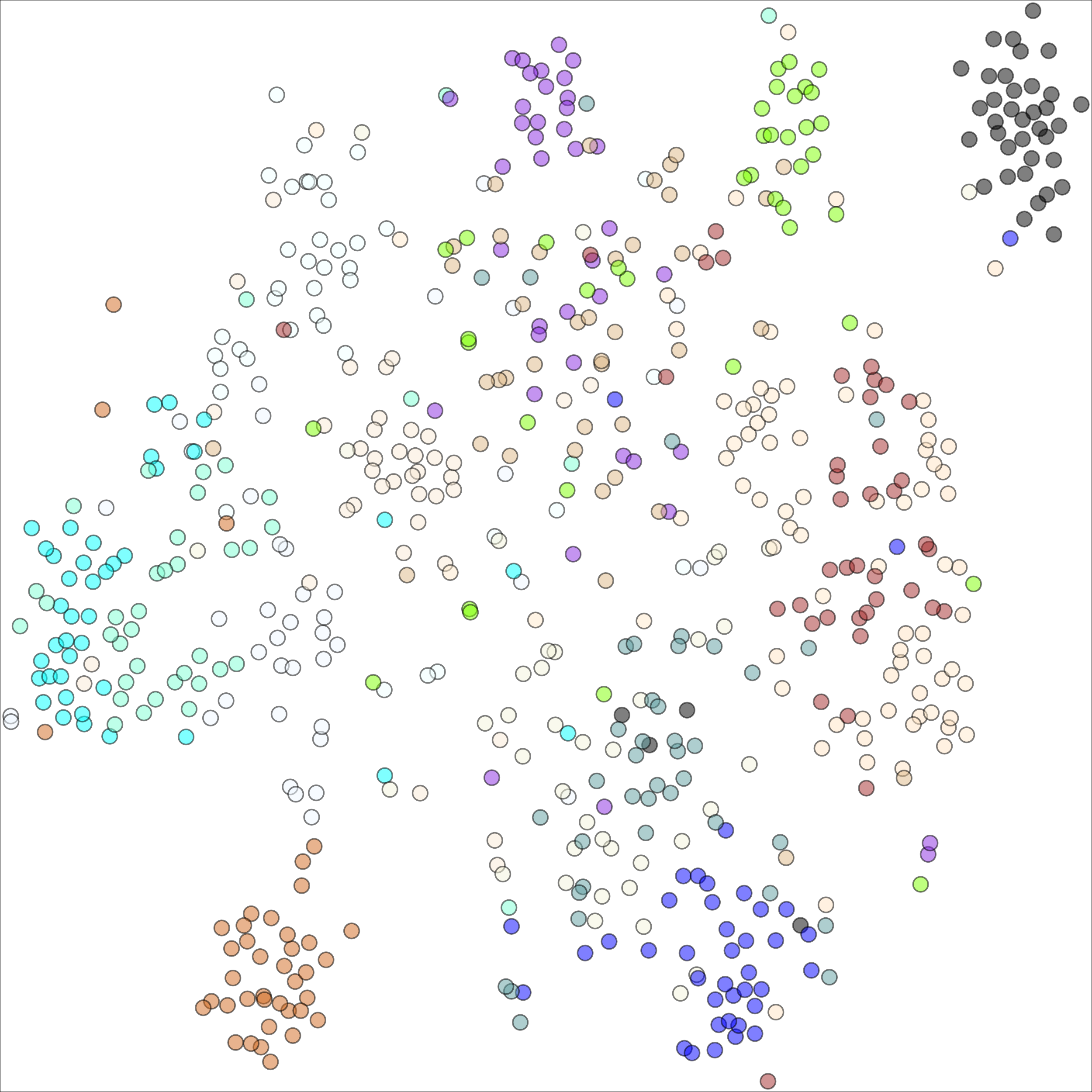} &   
    \includegraphics[width=0.09\linewidth]{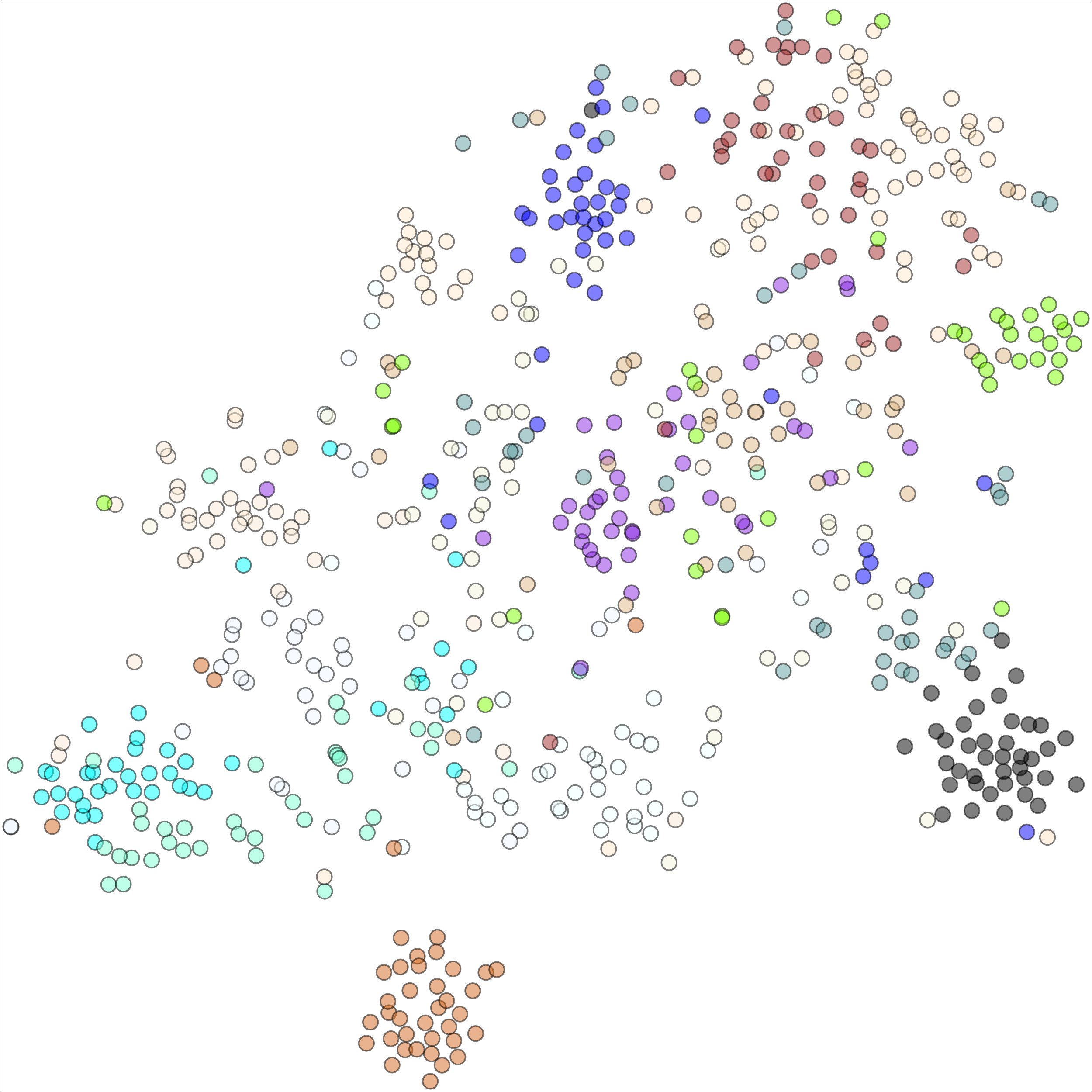} &  
    \includegraphics[width=0.09\linewidth]{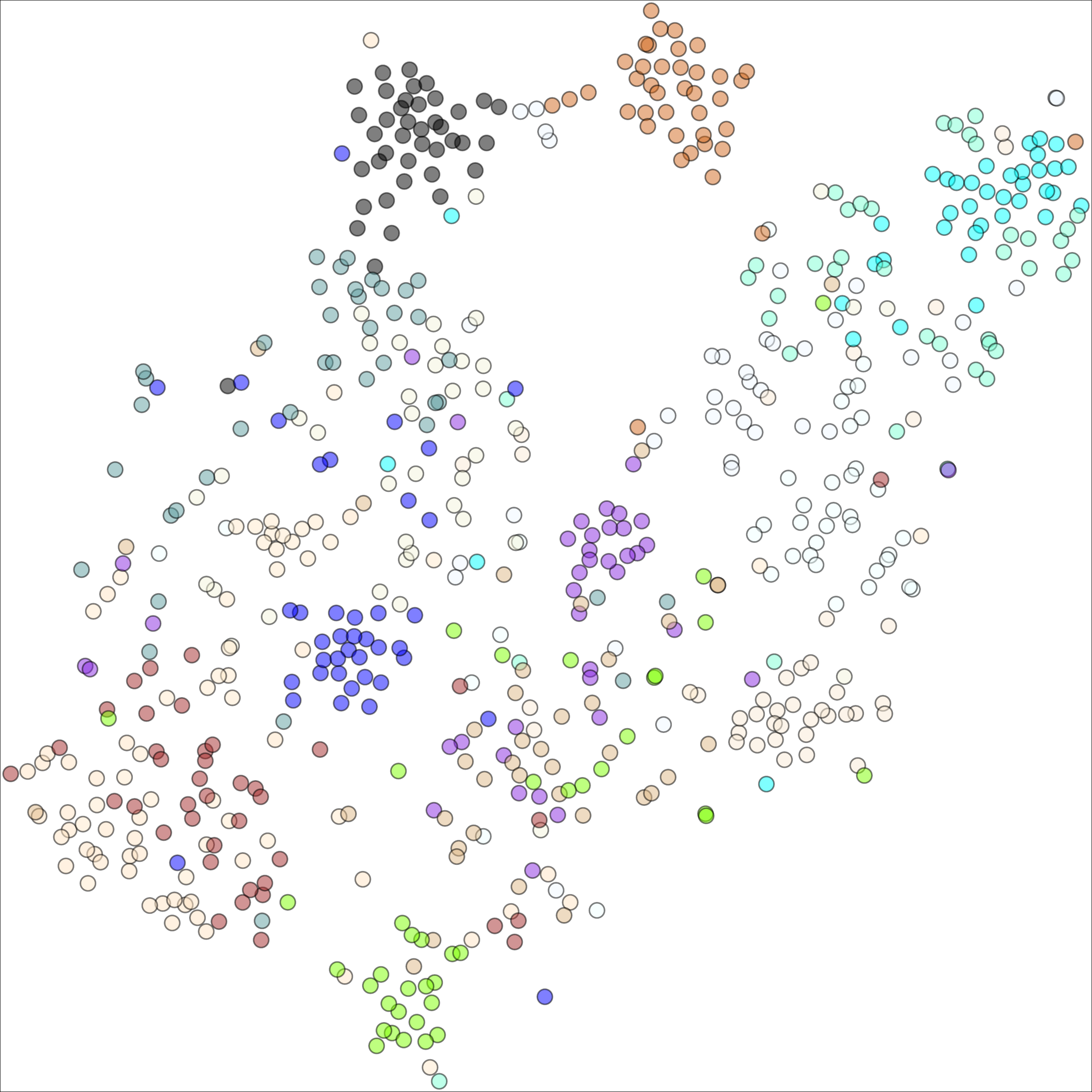} & 
    \includegraphics[width=0.09\linewidth]{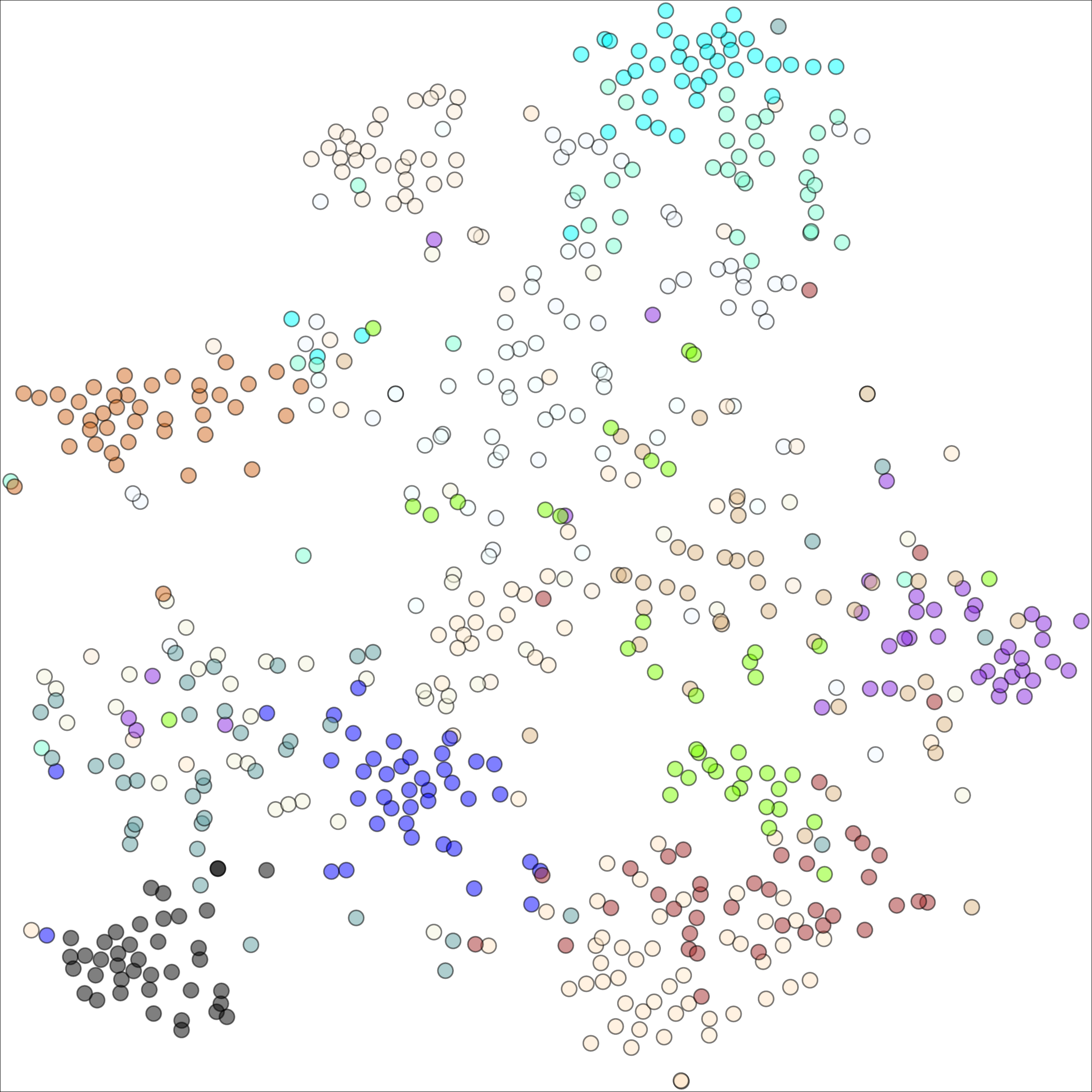} &   
    \includegraphics[width=0.09\linewidth]{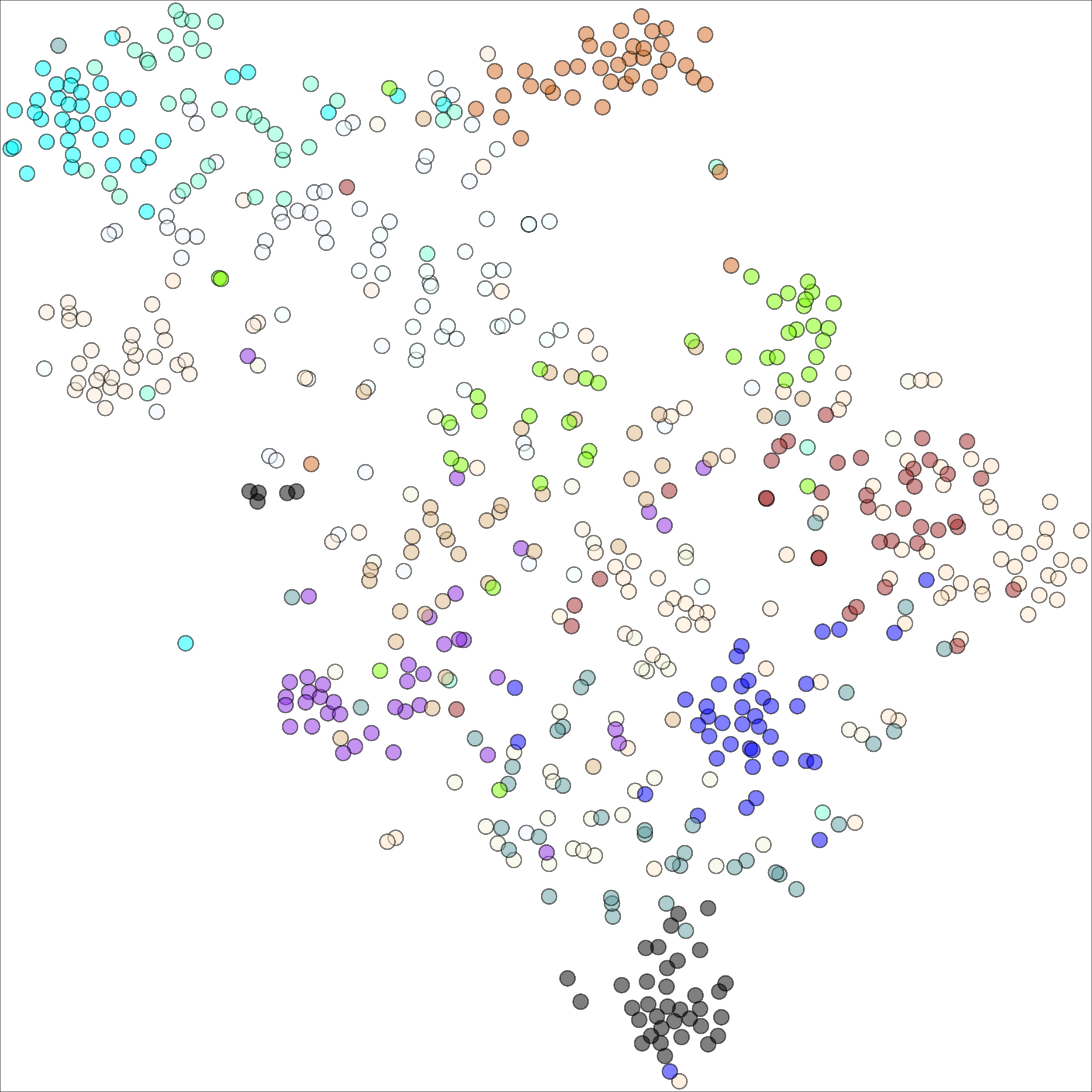} &   
    \includegraphics[width=0.09\linewidth]{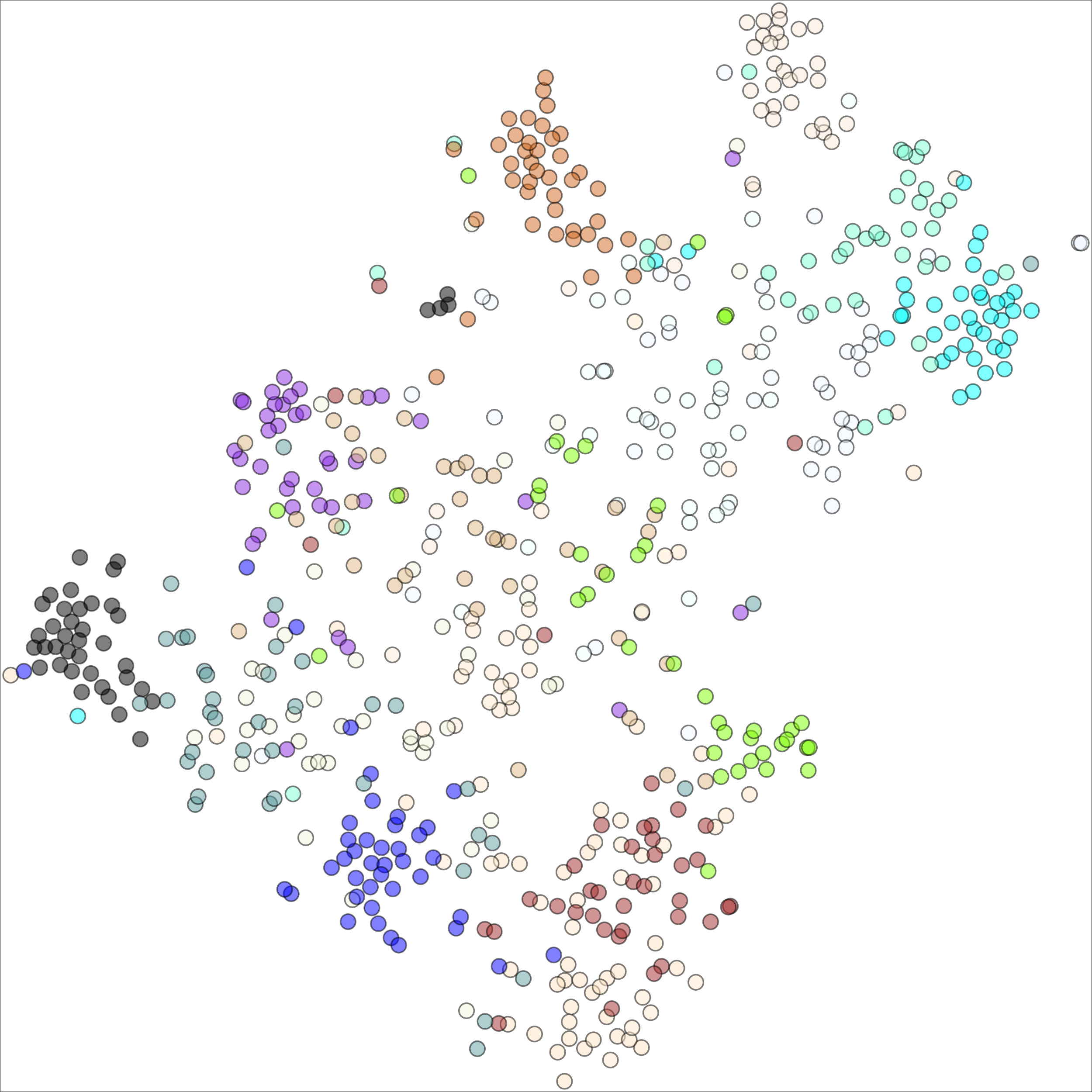} &   
    \includegraphics[width=0.09\linewidth]{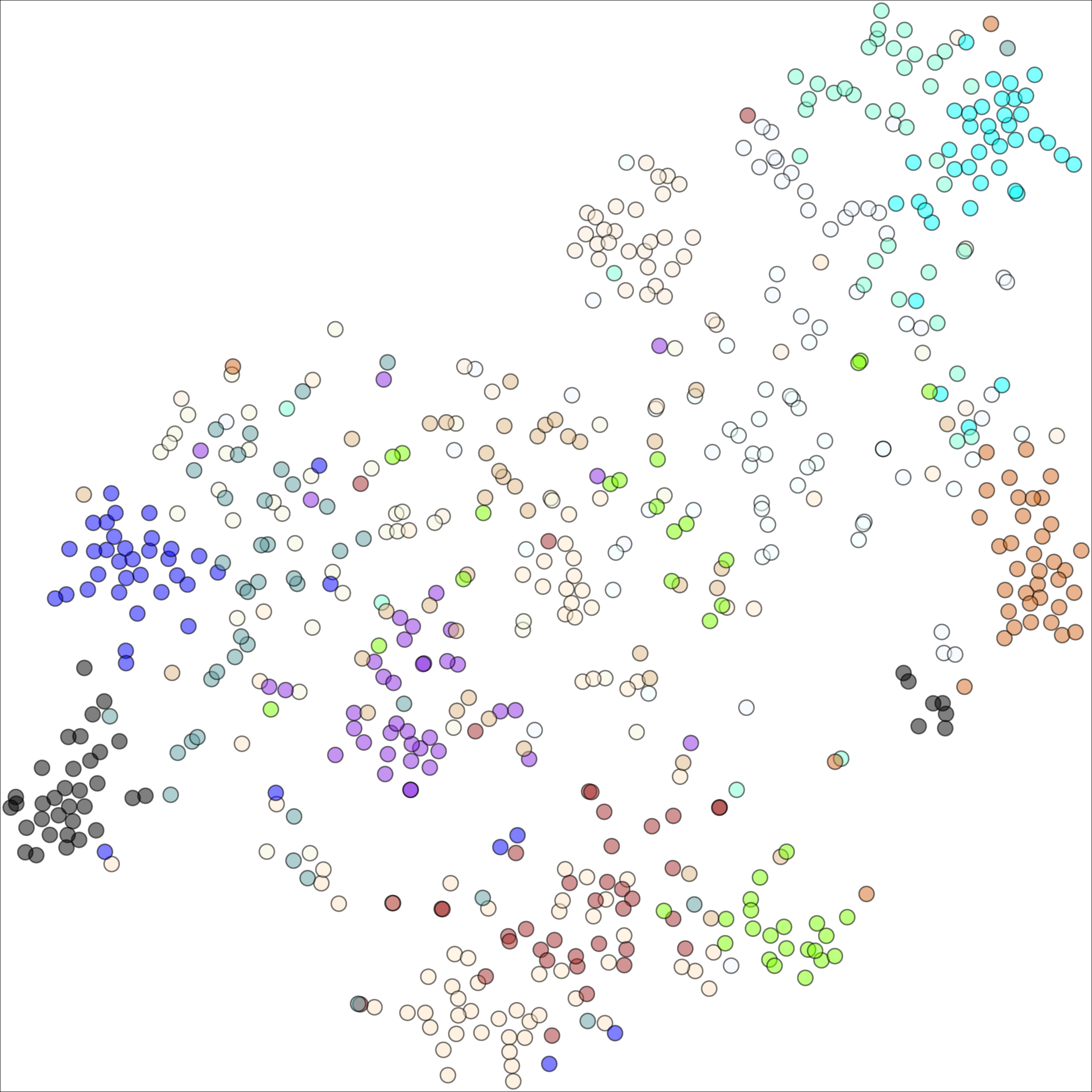} &  
    \\
    \rotatebox{90}{\hspace{2em} train}  
    \includegraphics[width=0.09\linewidth]{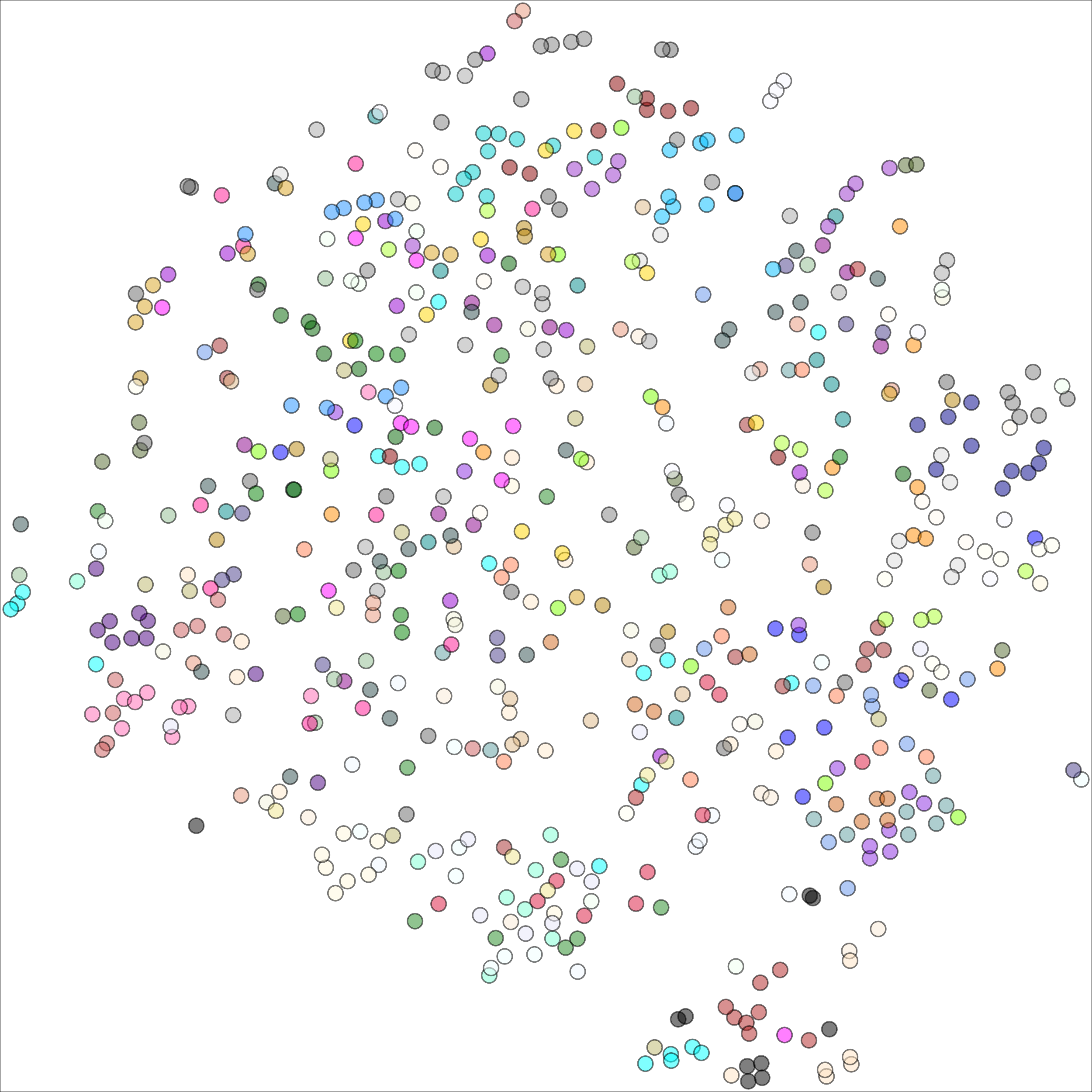} & 
    \includegraphics[width=0.09\linewidth]{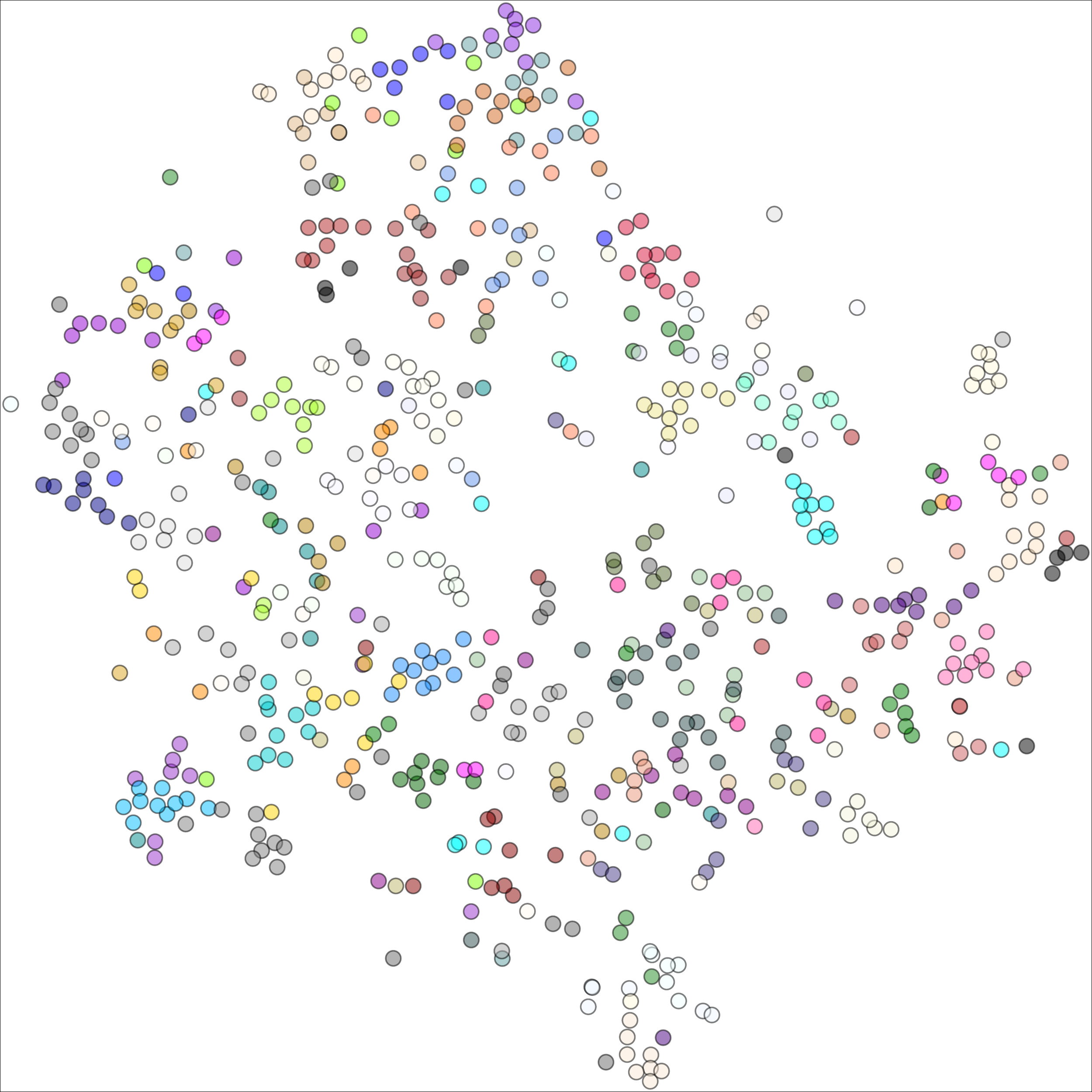} &   
    \includegraphics[width=0.09\linewidth]{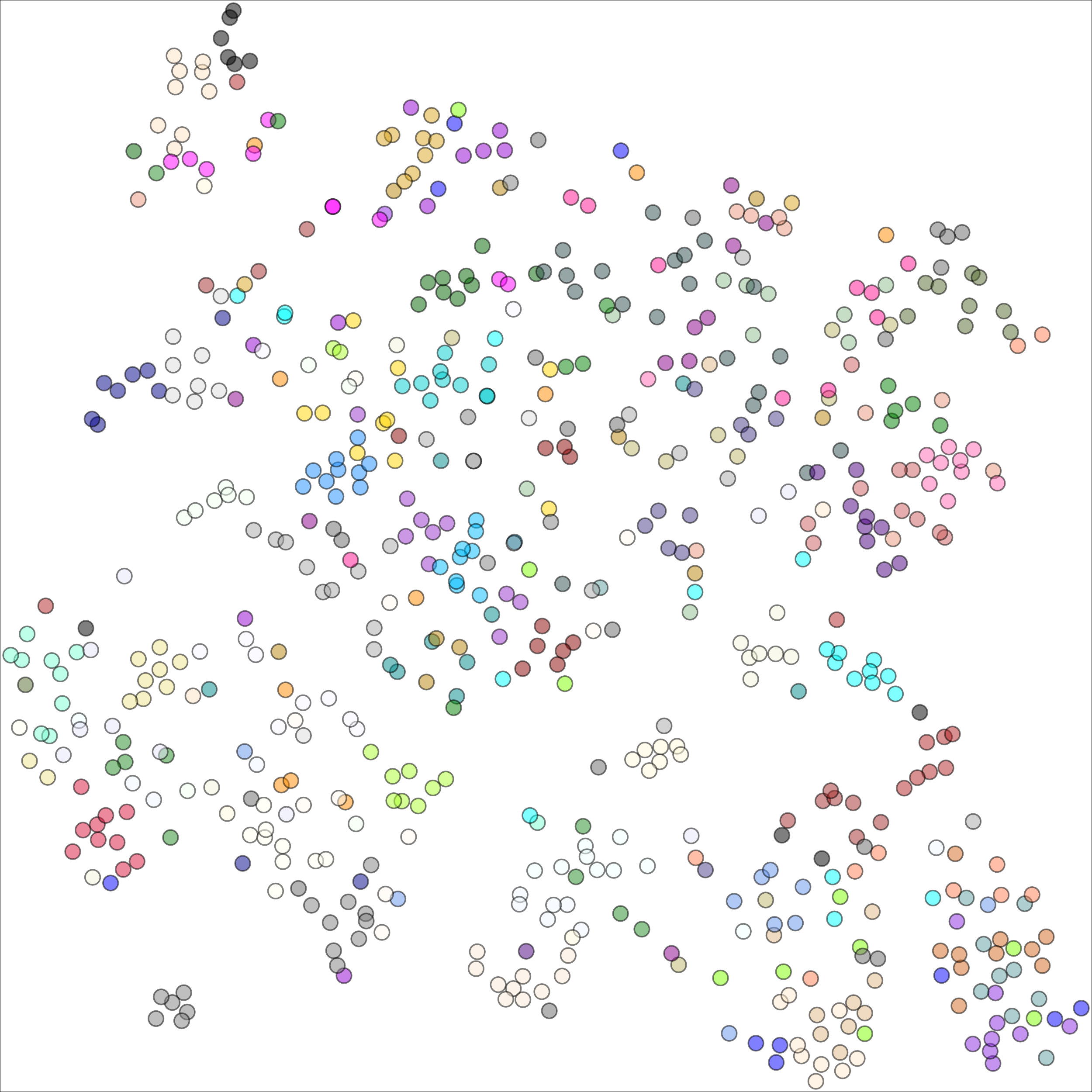} &   
    \includegraphics[width=0.09\linewidth]{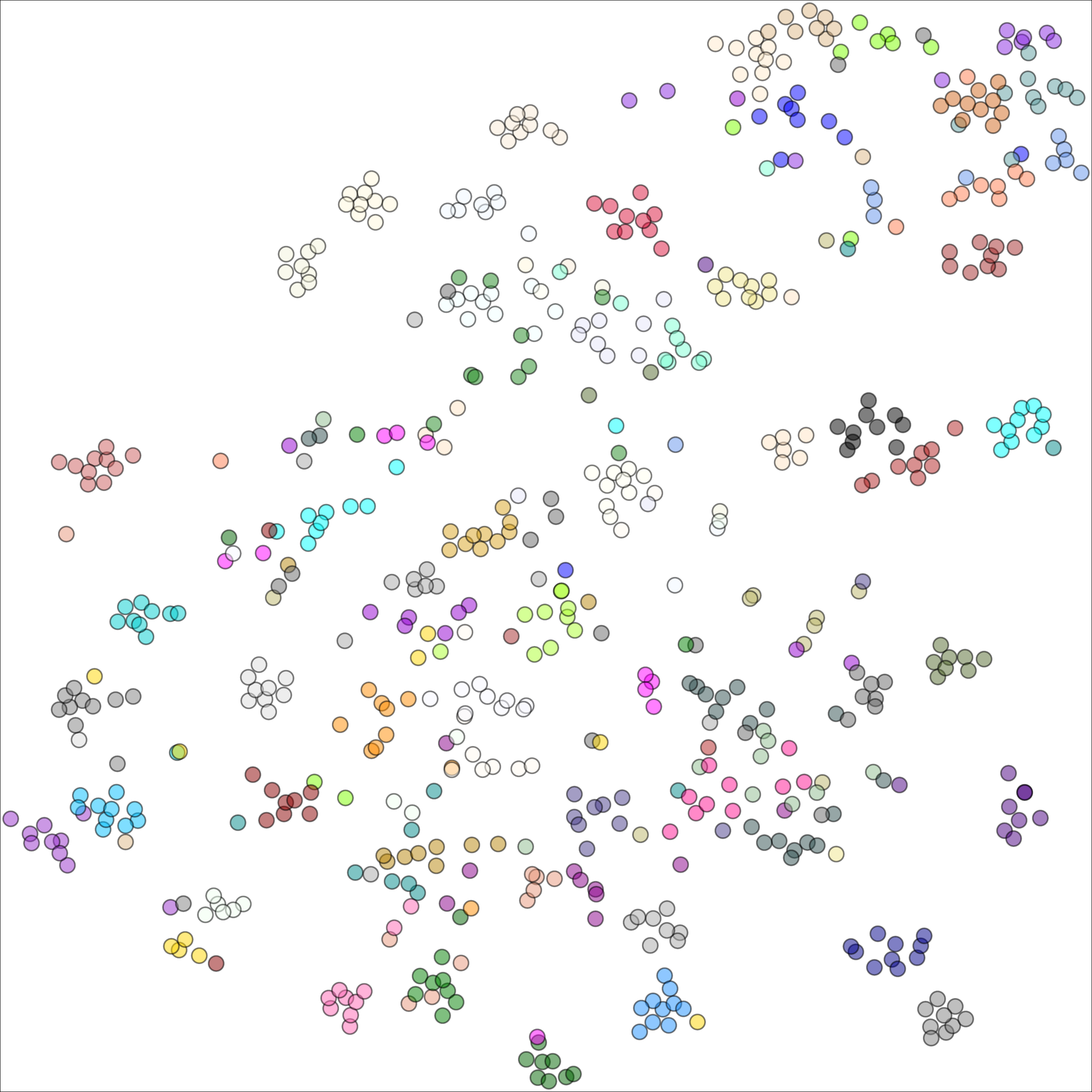} &   
    \includegraphics[width=0.09\linewidth]{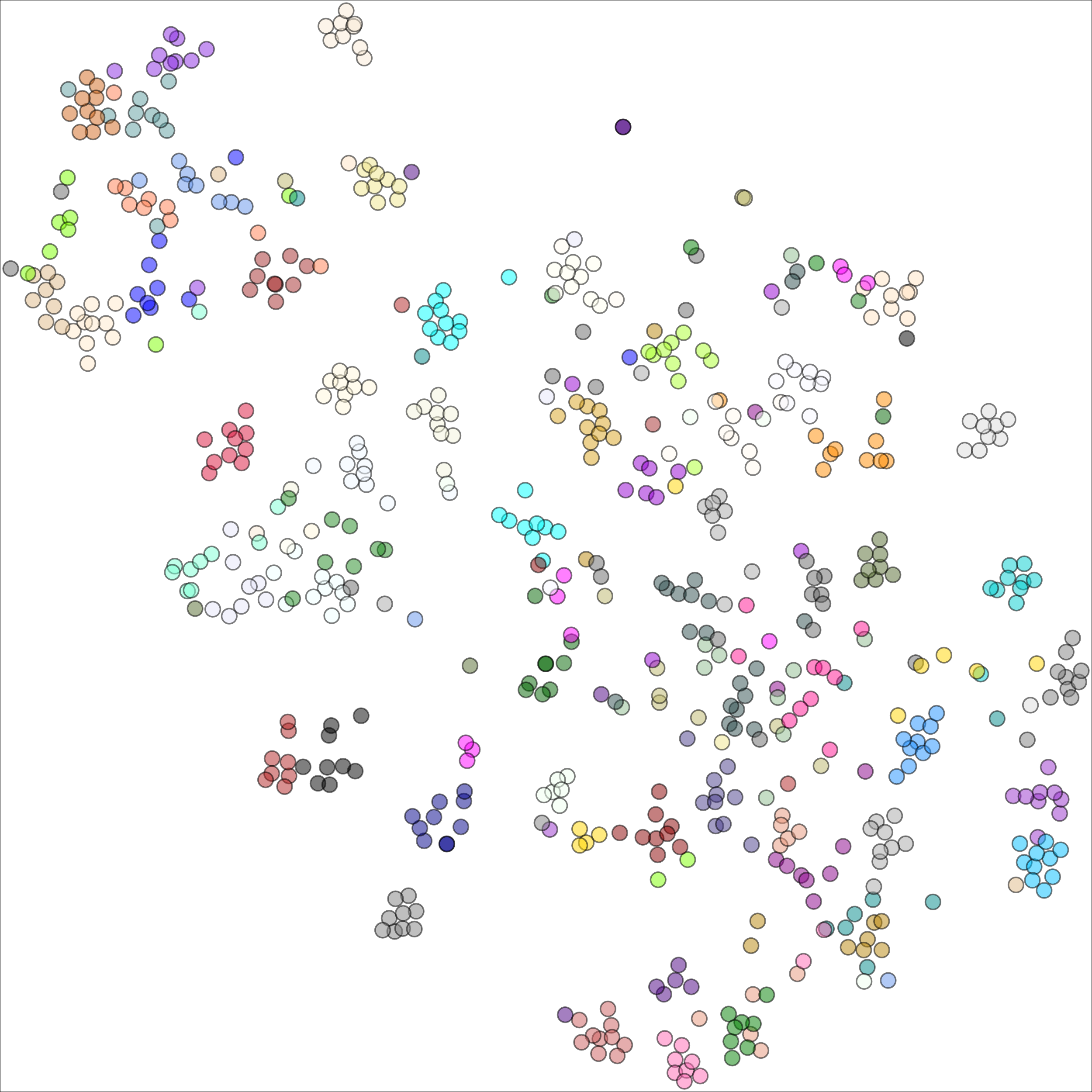} &  
    \includegraphics[width=0.09\linewidth]{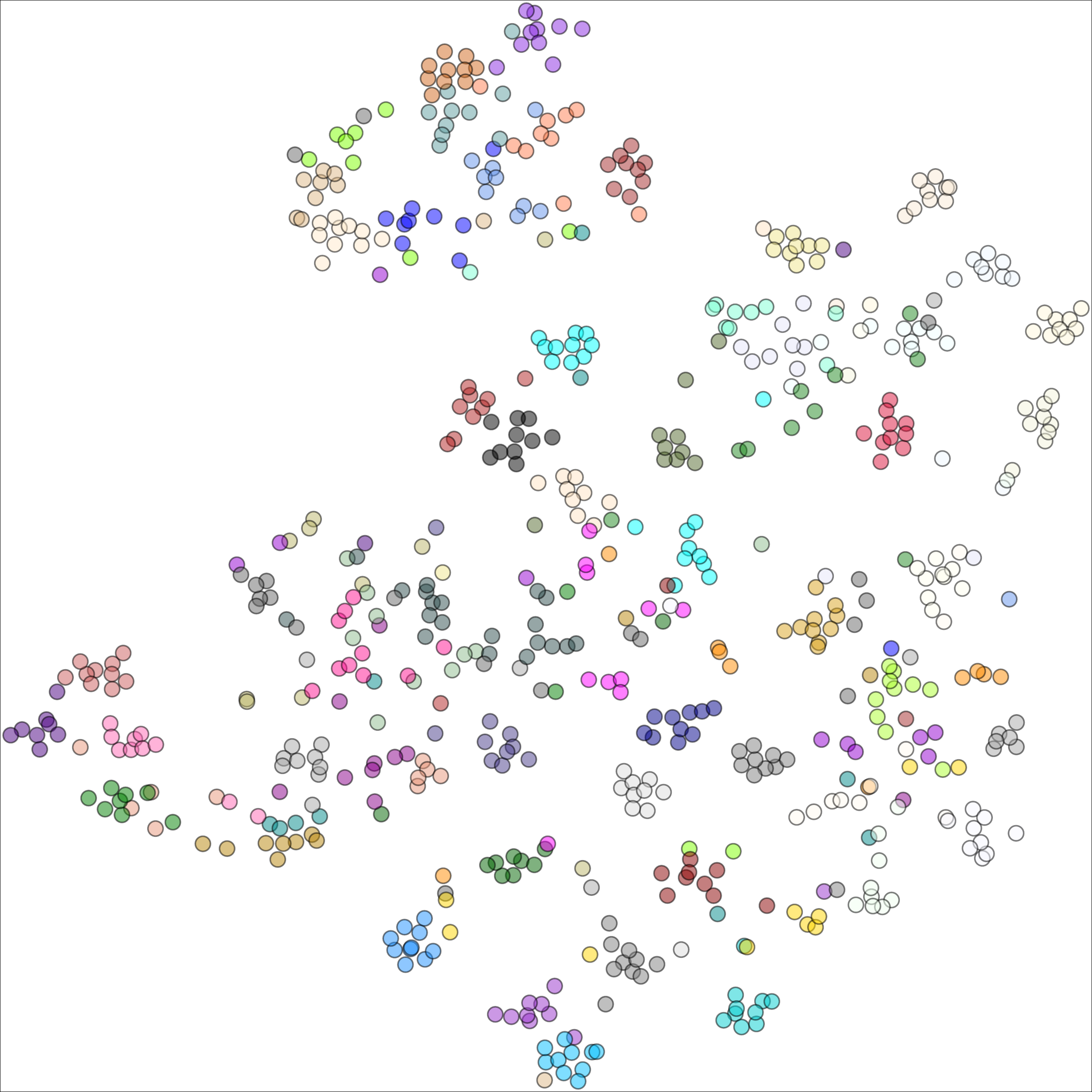} & 
    \includegraphics[width=0.09\linewidth]{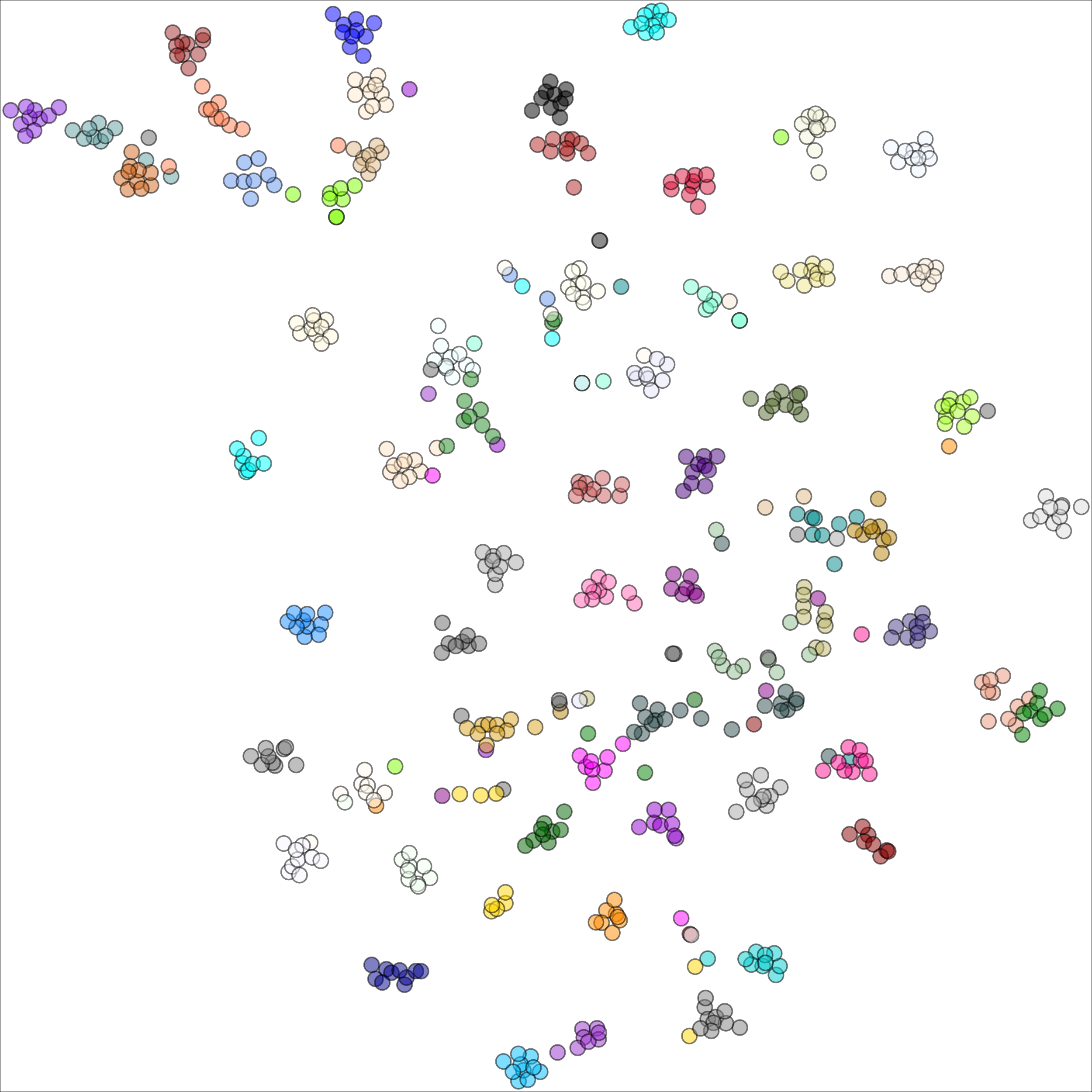} &   
    \includegraphics[width=0.09\linewidth]{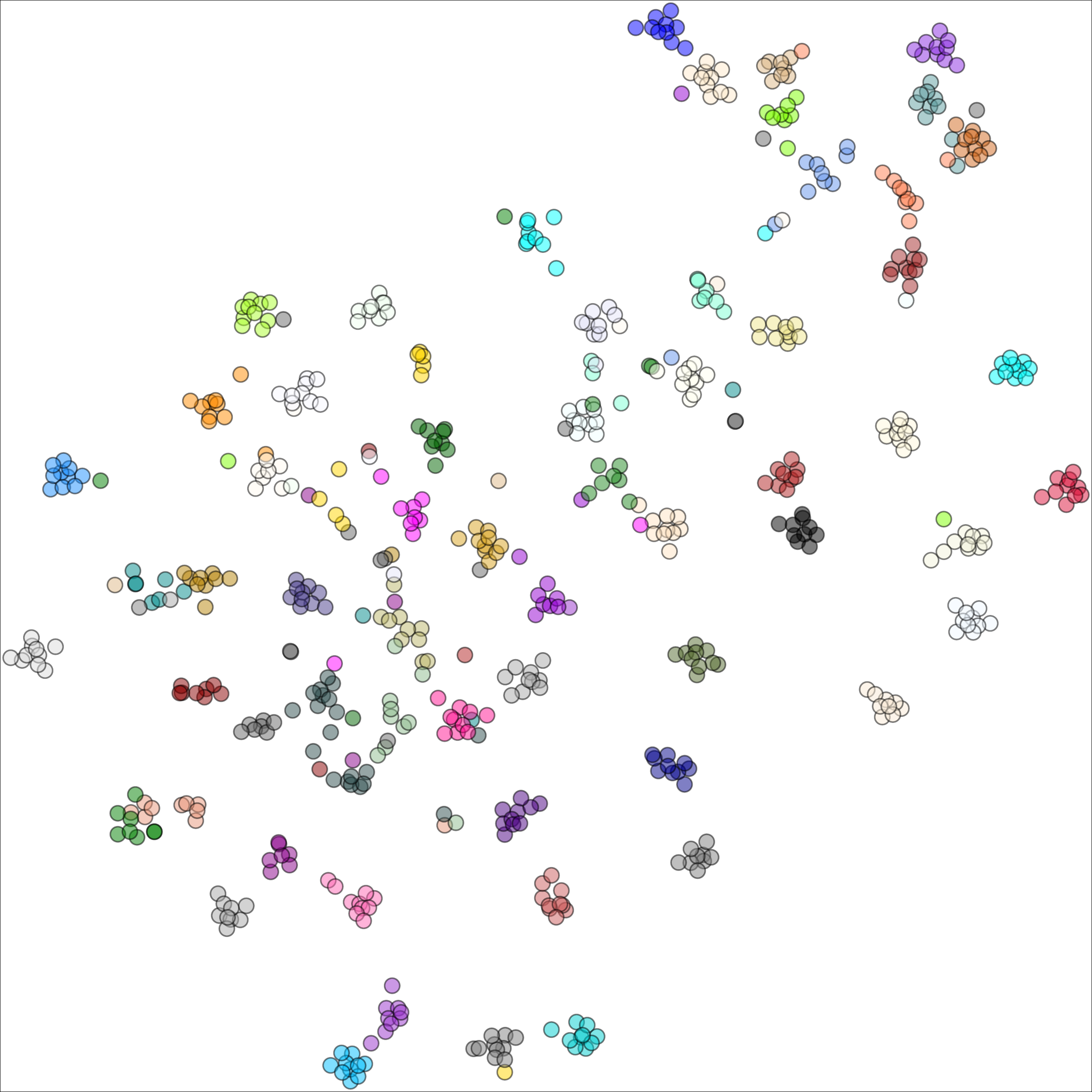} &   
    \includegraphics[width=0.09\linewidth]{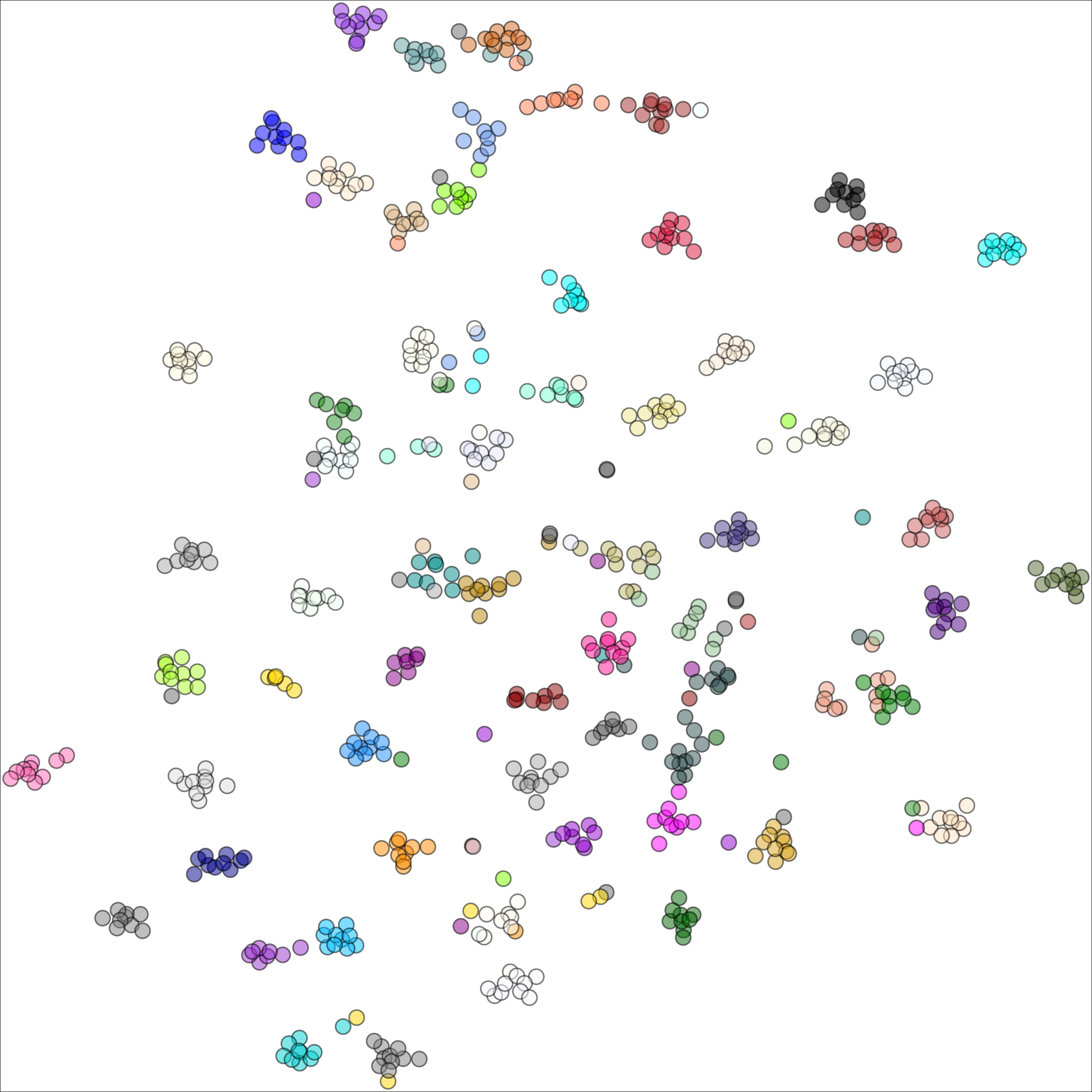} &   
    \includegraphics[width=0.09\linewidth]{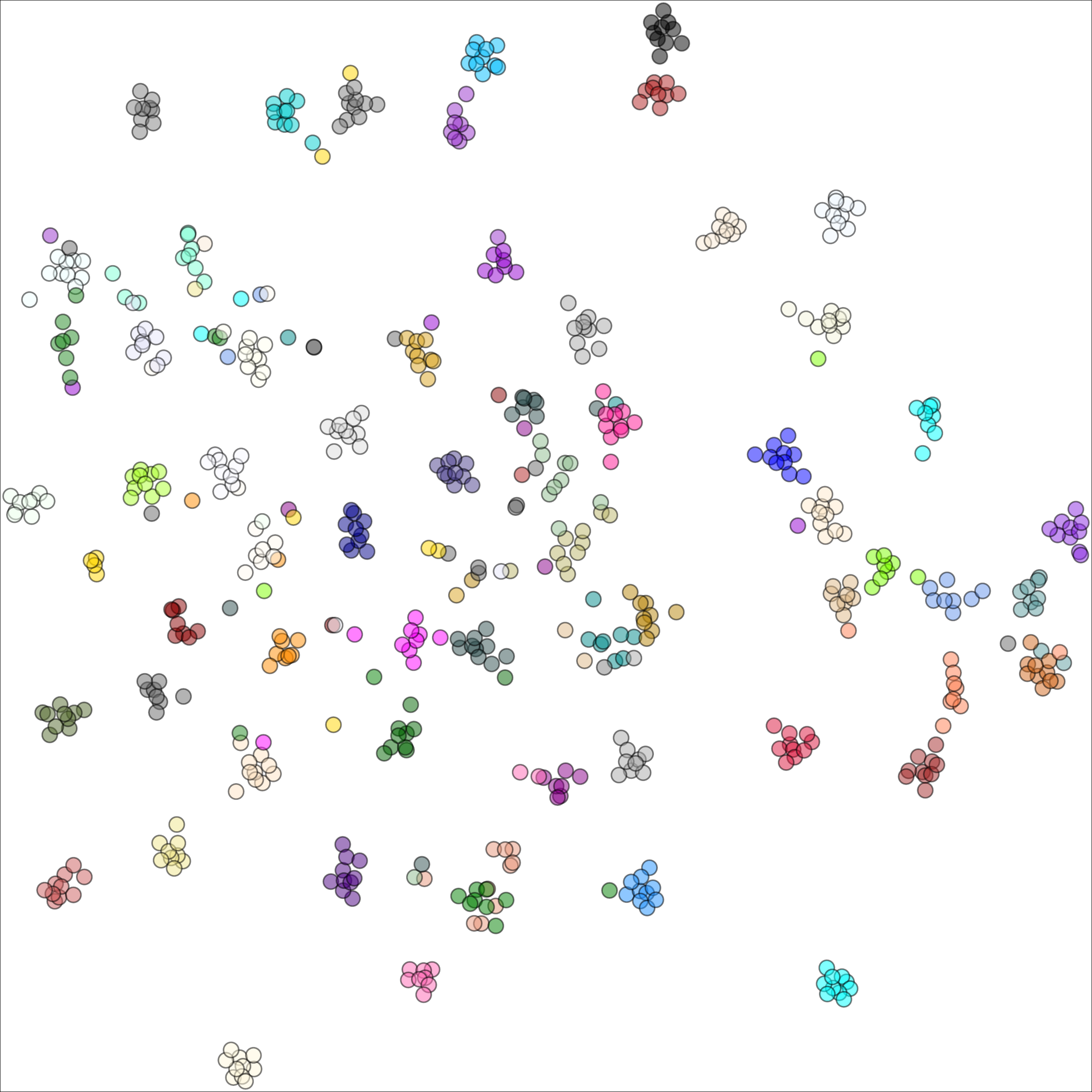} &  
    \ \\
   \footnotesize{map.1}  &  \footnotesize{map.2}  &  \footnotesize{map.3}  &  \footnotesize{map.4}  &  \footnotesize{map.5}  &  \footnotesize{map.6}  &  \footnotesize{map.7}  &  \footnotesize{map.8}  &  \footnotesize{map.9}  &  \footnotesize{map.10}  \\       \vspace{-2em}
    \end{tabular} 
    \caption{tSNE of miniIN's 640 samples of 64 train and 16 valid. classes (color-coded) by SF12 mappers separately (columns). 
    }  
     \vspace{-1.5em}
    \label{fig:tsne_mappers}
\end{sidewaysfigure}

\section*{Hausdorff distance ablation}
\label{sec:Hausdorff}  
Our matching feature set work can be extended to other set distances. Tab.~\ref{tab:Hausdorff} presents our method with Hausdorff (in blue) compared to our Sum-min for both miniIN and CUB.  


\begin{table*}[hbt!]
\centering
\label{sec:Hausdorff}  
 
\caption{ MiniIN (from Table 1) and CUB (from Table 3) by SF4-64 plus blue.} 
  \begin{tabular}{rlccc}  
        \toprule    
        & \textbf{config.}    &\textbf{1-shot}  &\textbf{5-shot}  \\ 
        \midrule    
         \multirow{2}{*}{\rotatebox{90}{miIN}} 
         & Sum-min                          & \textbf{57.18}        & \textbf{73.67} \\
         & {\color{blue}Hausdorff}          & 56.07        & 72.32      \\
        \midrule 
        \multirow{2}{*}{\rotatebox{90}{  CUB  }} 
        & Sum-min                          & \textbf{72.09}        & \textbf{87.05} \\
        & {\color{blue}Hausdorff}          & 70.20        & 84.85     \\
        \bottomrule       
\end{tabular}   \ \\ 
\label{tab:Hausdorff} 
\end{table*}


{\small
\bibliographystyle{ieee_fullname}
\bibliography{egbib.bib}
}

\end{document}